\documentclass[Afour,sageh,times]{sagej}

\usepackage{moreverb,url}

\usepackage[colorlinks,bookmarksopen,bookmarksnumbered,citecolor=red,urlcolor=red]{hyperref}

\usepackage{subfigure,tikz,tkz-graph}
\usepackage{mathtools}
\usepackage{algorithmic}
\usepackage{algorithm}
\usepackage{amsthm}
\usepackage{array}
\usepackage{multirow}
\usepackage{multicol}
\usepackage{makecell}
\usepackage{float}
\usepackage{indentfirst}

\newcommand\BibTeX{{\rmfamily B\kern-.05em \textsc{i\kern-.025em b}\kern-.08em
T\kern-.1667em\lower.7ex\hbox{E}\kern-.125emX}}

\def\volumeyear{2016}

\newcommand{\T}{^{\textrm{T}}}
\newcommand{\tetrahedron}{
	\mathchoice
	{\includegraphics[height=1.6ex]{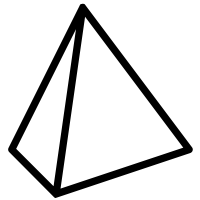}} 
	{\includegraphics[height=1.6ex]{f0.png}} 
	{\includegraphics[height=1.3ex]{f0.png}} 
	{\includegraphics[height=0.9ex]{f0.png}} 
}
\newtheorem{remark}{Remark}
\newtheorem{definition}{Definition}
\newtheorem{theorem}{Theorem}
\newtheorem{assumption}{Assumption}
\newtheorem{lemma}{Lemma}
\newtheorem{proposition}{Proposition}

\begin{document}

\runninghead{Liang et al.}

\title{3-D Relative Localization for Multi-Robot Systems with Angle and Self-Displacement Measurements}

\author{Chenyang Liang\affilnum{1}$^,$ \affilnum{2}, Liangming Chen \affilnum{2}, Baoyi Cui \affilnum{1} and Jie Mei \affilnum{1}$^,$ \affilnum{3}}

\affiliation{\affilnum{1}School of Intelligence Science and Engineering, Harbin Institute of Technology, Shenzhen, China.\\
\affilnum{2}School of Automation and Intelligent Manufacturing, and Guangdong Provincial Key Laboratory of Fully Actuated System Control Theory and Technology, Southern University of Science and Technology, China. \\
\affilnum{3}Guangdong Key Laboratory of Intelligent Morphing Mechanisms and Adaptive Robotics, Harbin Institute of Technology, Shenzhen, China. \\
Chenyang Liang and Liangming Chen contribute equally to this work.
}

\corrauth{
	Jie Mei, HIT Campus of University Town of Shenzhen, Shenzhen 518055, China. Email: jmei@hit.edu.cn.\\Liangming Chen, 1088 Xueyuan Avenue, Shenzhen 518055, China. Email: chenlm6@sustech.edu.cn.}

\begin{abstract}
	Inter-robot relative localization (including positions and orientations) is necessary for multi-robot systems to execute collaborative tasks.
	Realizing relative localization by leveraging inter-robot local measurements is a challenging problem, especially in the presence of measurement noise.
	Motivated by this challenge, in this paper we propose a novel and systematic 3-D relative localization framework based on inter-robot interior angle and self-displacement measurements, which are accessible from existing sensors.
	Initially, we propose a linear relative localization theory comprising a distributed linear relative localization algorithm and sufficient conditions for localizability. 
	According to this theory, robots can determine their neighbors' relative positions and orientations in a purely linear manner, relying solely on angle and self-displacement measurements.
	Subsequently, in order to deal with measurement noise, we present an advanced Maximum a Posterior (MAP) estimator by addressing three primary challenges existing in the MAP estimator. 
	Firstly, it is common to formulate the MAP problem as an optimization problem, whose inherent non-convexity can result in local optima when finding the optimal solution. 
	To address this issue, we reformulate the linear computation process of the linear relative localization algorithm as a Weighted Total Least Squares (WTLS) optimization problem on manifolds.
	The optimal solution of the WTLS problem is more accurate and closer to the true values, which can then be used as initial values when solving the optimization problem associated with the MAP problem, thereby reducing the risk of falling into local optima.
	The second challenge is the lack of knowledge of the prior probability density of the robots' relative positions and orientations at the initial time, which is required as an input for the MAP estimator.
	To deal with it, we combine the WTLS with a Neural Density Estimator (NDE).
	Thirdly, to prevent the increasing size of the relative positions and orientations to be estimated as the robots continuously move when solving the MAP problem, a marginalization mechanism is designed, which ensures that the computational cost remains constant.
	Indoor and outdoor experiments of multiple drones' relative localization are performed to verify the effectiveness of the proposed framework.
\end{abstract}


\maketitle

\section{Introduction}

Inter-robot relative localization, including relative positions and orientations of neighbor robots, is fundamental for multi-robot systems to cooperatively execute diverse tasks, such as information gathering \cite{atanasov2014information} and exploration \cite{bhattacharya2014multi}.  
Localization of a single robot has been widely researched and one of the most popular algorithms is the Simultaneous Localization and Mapping (SLAM) \cite{cadena2016past}.
Most SLAM algorithms are based on laser radars or cameras, and extending SLAM algorithms from single-robot systems to multi-robot systems to achieve relative localization remains a significant challenge.
Alternatively, algorithms based on inter-robot measurements have been developed to achieve relative localization, such as ultra-wideband (UWB) for distances \cite{zhang2008high}, cameras or passive sonars for bearings \cite{montesano2005cooperative, hahn1975optimum}, and cameras or antenna arrays (such as Bluetooth 5.1 modules) for angles \cite{sambu2022experimental}.

Existing relative localization algorithms based on inter-robot measurements can be mainly categorized into three classes, namely learning-based algorithms, algebraic-based algorithms and probability-based algorithms.
Learning-based algorithms, particularly multi-agent deep reinforcement learning (MARL) algorithms, employ trained neural networks to determine optimal estimations while bypassing complex analytical computations.
In \cite{peng2019decentralized}, the multi-agent localization problem is formulated as a Markov decision process and a distributed MARL algorithm is proposed to find the optimal estimation.
Similarly, \cite{mishra2024multi} presents another MARL algorithm designed for persistent localization tasks.
However, these learning-based algorithms are difficult to generalize across different environments and operational conditions \cite{lyle2022learning, packer2018assessing}.
Algebraic-based algorithms involve using the inter-robot measurements and then conducting algebraic equations based on related physical laws.
For instance, in \cite{mao2013relative} and \cite{pugh2006relative}, by measuring offset angles among several on-board receivers, algebraic equations are formulated based on the angle relationship between receivers, which are then solved to determine the inter-robot relative positions.
In \cite{zhou2012determining} and \cite{trawny2010interrobot}, 3-D relative localization is achieved by directly using distance or bearing measurements between two robots, in which ambiguous solutions exist.
Recently, a new type of localization algorithms, known as the linear localization algorithms, has been proposed, which guarantees a unique solution.
For example, in \cite{han2017barycentric, lin2016distributed, chen20223d}, linear localization algorithms are developed based on different types of measurements to address the sensor network localization problem, where the unique positions of sensors can be determined in a purely linear algebraic manner.
Additionally, the algorithm proposed in \cite{nguyen2019persistently} achieves relative localization by using robots’ self-displacements and inter-robot distances, which guarantees global convergence regardless of the initial estimation.
However, it assumes that all robots’ odometry coordinate frames are aligned, which may be difficult to ensure without explicit calibration.

While algebraic-based algorithms offer simplicity in implementation, their localization accuracy is limited in the presence of measurement noise.
To obtain more accurate localization results in general scenarios, probability-based localization algorithms are often preferred.
Existing probability-based 3-D relative localization algorithms can be further categorized into two classes: MLE-based (maximum likelihood estimation) and BI-based (Bayesian inference).
MLE-based algorithms operate by formulating the MLE as an optimization problem and finding its optimal solution, which corresponds to the estimation with the highest likelihood given all measurements.
In \cite{jiang20193}, by utilizing inter-agent distance measurements, the MLE problem is first transformed into a semidefinite programming (SDP) problem. 
The solution obtained from this SDP is then used as the initial value when finding the optimal solution of the original optimization problem related to the MLE problem.
Differently, in \cite{li20223} and \cite{nguyen2023relative}, the MLE problem is formulated as a quadratically constrained quadratic programming (QCQP) problem, which is subsequently converted into an SDP.
The main advantage of MLE-based algorithms is that they only depend on the measurements and the corresponding measurement model.
Compared to MLE-based algorithms, BI-based algorithms aim to estimate the relative localization with the highest posterior probability density by integrating both the likelihood and prior information.
The extended Kalman filter (EKF) is a typical BI algorithm, frequently applied in localization problems, as demonstrated in \cite{kia2018server}, \cite{luft2018recursive} and \cite{roumeliotis2000bayesian}. 
Moreover, the Maximum a Posterior (MAP) algorithm, which is another type of BI-based algorithms, is also extensively utilized in various tasks, such as SLAM \cite{cadena2016past}.
For example, authors in \cite{cieslewski2018data} propose a relative localization algorithm based on shared visual features between robots.
To address the high communication bandwidth requirements, they employ an efficient place representation algorithm \cite{cieslewski2017efficient} that significantly reduces the bandwidth consumption while maintaining system scalability.
A key advantage of this algorithm is that it does not require any extra sensors to achieve relative localization.
In \cite{nguyen2015bayesian} and \cite{cossette2021relative}, different relative localization algorithms based on MAP have been developed.
The difference between the EKF and the MAP is that the EKF is a real-time estimation algorithm, while the MAP is a batch estimation algorithm that processes all available measurements simultaneously to estimate the states during a period.
Compared to other probability-based algorithms, MAP algorithms commonly offer estimation with better accuracy since they utilize both the likelihood and the prior information \cite{beerli2006comparison, smith1987comparison}.

\begin{figure}[tbhp]
  \centering
\subfigure[Localization errors of different prior densities]
  {
      \includegraphics[width=0.4\textwidth]{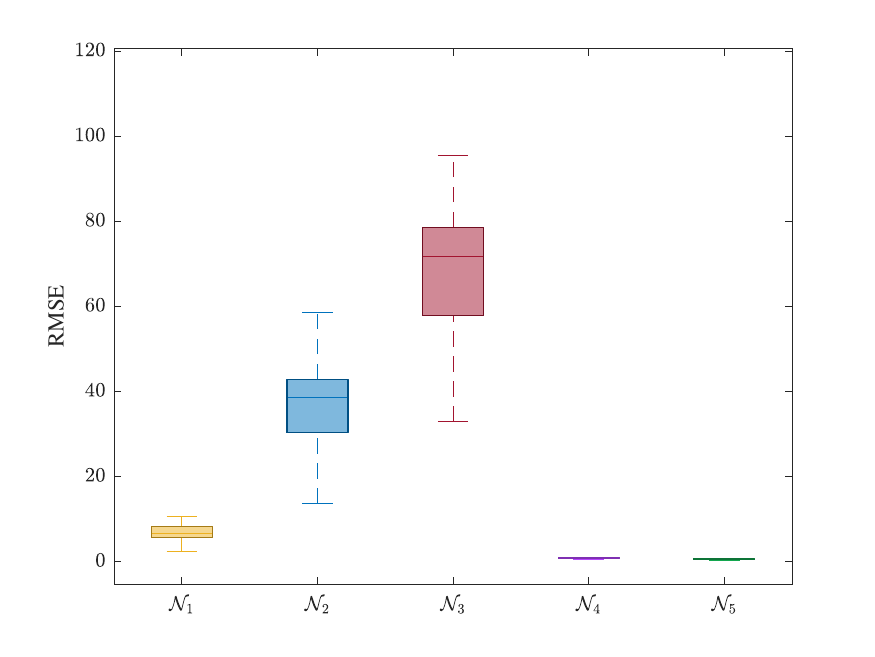}
  }
  \subfigure[$\mathcal{N}(0, 500)$]
  {
      \includegraphics[width=0.13\textwidth]{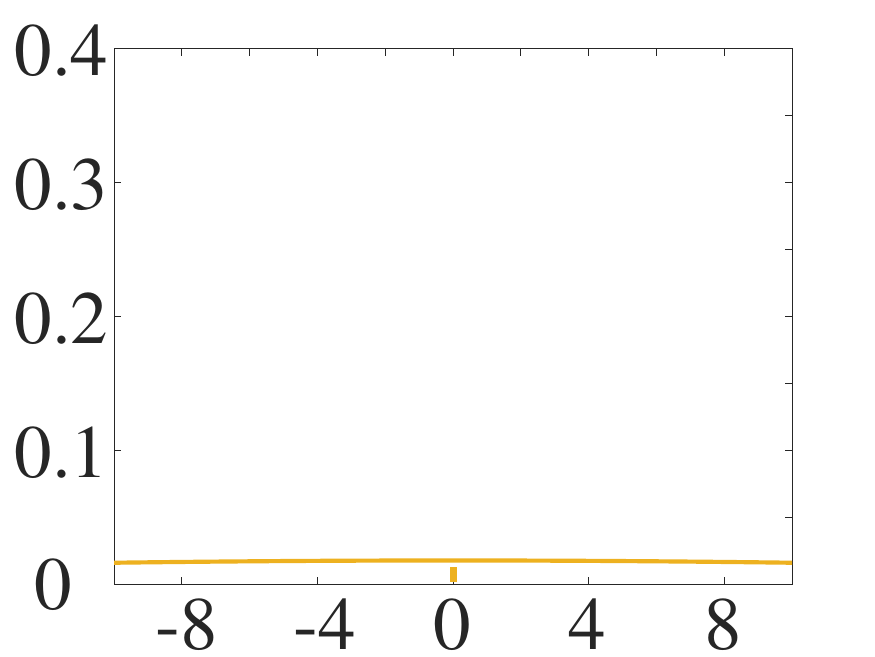}
  }
  \subfigure[$\mathcal{N}(0.2, 25)$]
  {
      \includegraphics[width=0.13\textwidth]{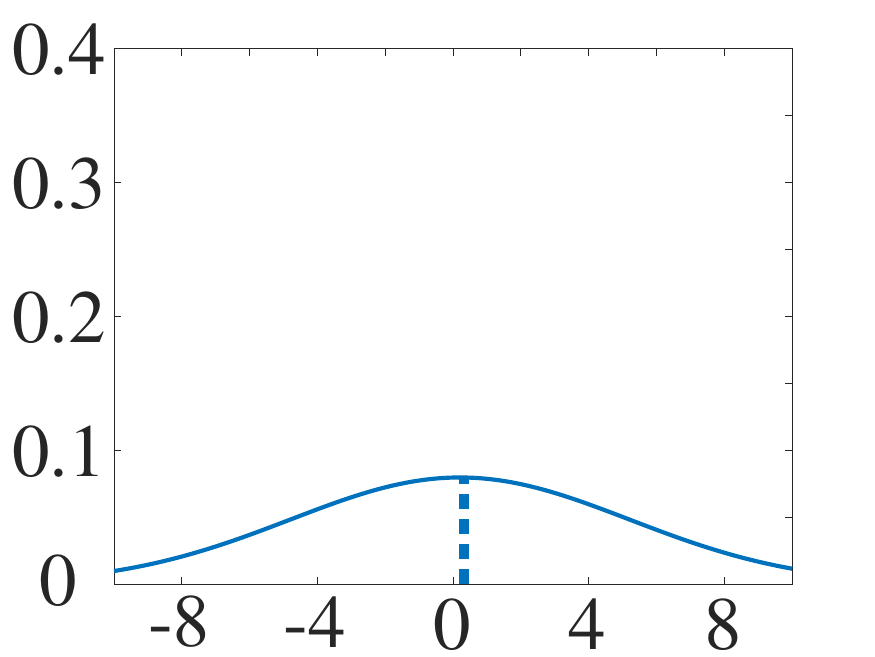}
  }
\subfigure[$\mathcal{N}(0.4, 25)$]
  {
      \includegraphics[width=0.13\textwidth]{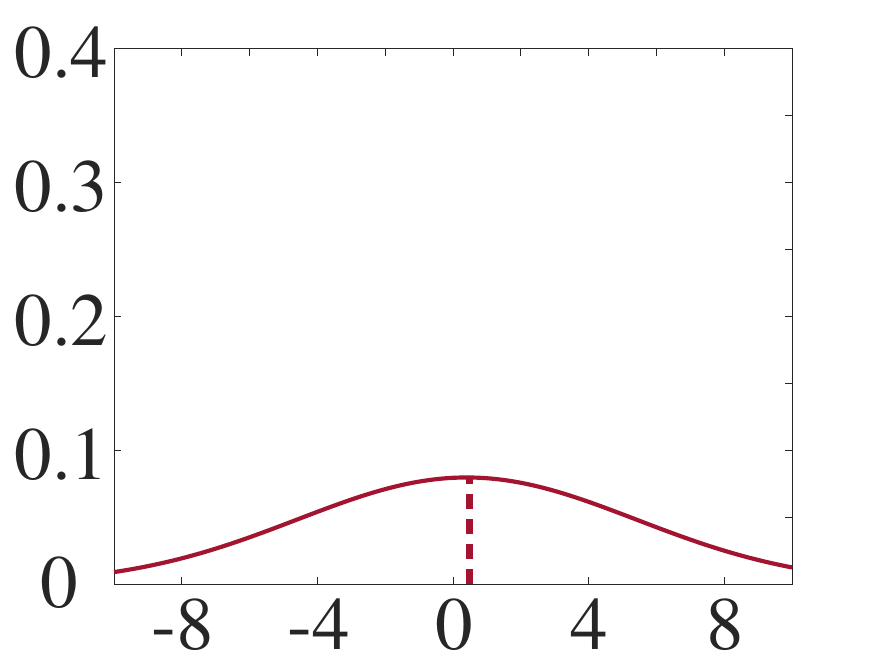}
  }
\subfigure[$\mathcal{N}(0, 5)$]
  {
      \includegraphics[width=0.13\textwidth]{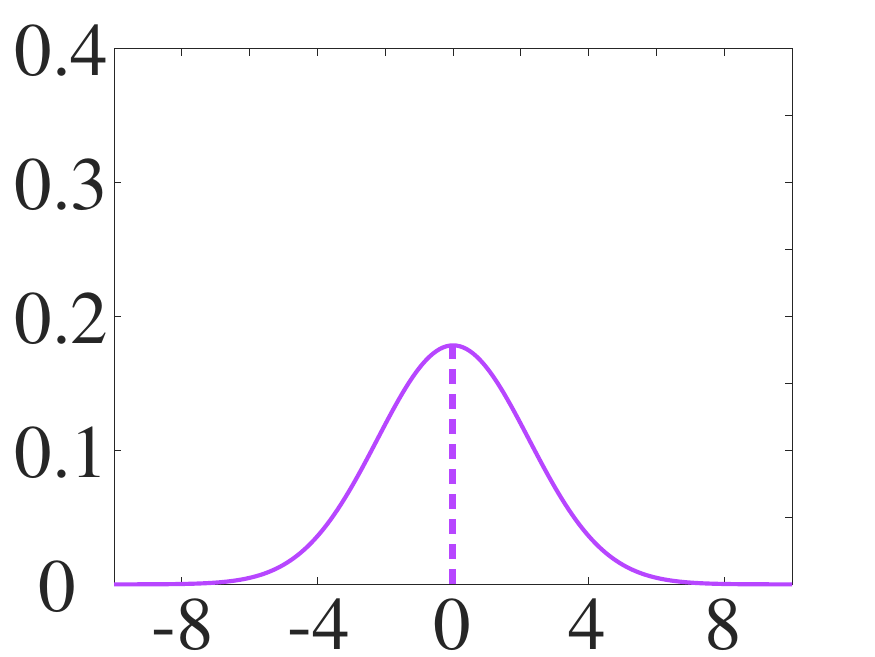}
  }
\subfigure[$\mathcal{N}(0, 0.2)$]
  {
      \includegraphics[width=0.13\textwidth]{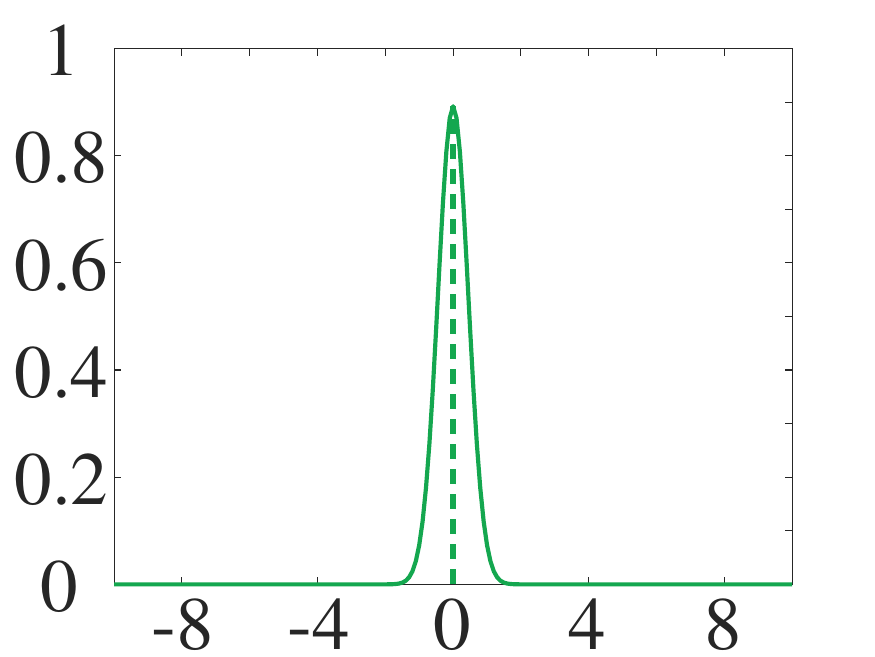}
  }
  \caption{
	Impacts of prior densities on the relative localization accuracy.
	In this example, the truth of the initial value $\mu$ is set to zero and all prior densities are modeled as Gaussian distributions.
	An ideal prior density has a mean close to the truth and a small covariance.
	Specifically, we define $\mathcal{N}_1 = \mathcal{N}(\mu_0, 500 \cdot I_n)$, $\mathcal{N}_2 = \mathcal{N}(\mu_{0.2}, 25 \cdot I_n)$, $\mathcal{N}_3 = \mathcal{N}(\mu_{0.4}, 25 \cdot I_n)$, $\mathcal{N}_4 = \mathcal{N}(\mu_0, 5 \cdot I_n)$, $\mathcal{N}_5 = \mathcal{N}(\mu_0, 0.2 \cdot I_n)$, where $\mu_{z}$ satisfies $\frac{\left\| \mu_{z} - \mu \right\|}{\left\| \mu \right\|} = z$, and $z$ denotes the offset of the mean $\mu_{z}$ with respect to $\mu$.
	Figures (b)-(f) illustrate one-dimensional examples corresponding to $\mathcal{N}_1$ to $\mathcal{N}_5$.
}
\label{fig prior effect}
\end{figure}

To obtain highly accurate relative localization results, the MAP algorithm is preferred.
However, some challenges exist with MAP algorithms.
The first challenge arises when solving the optimization problem related to the MAP, i.e., finding the optimal solution with the maximum posterior.
Due to the non-convexity of the MAP optimization problem, it's crucial to avoid convergence to local optima.
A common way to address this problem is transforming the non-convex optimization problem into an SDP problem \cite{li20223, nguyen2023relative}.
Another effective way is to initialize the optimization problem with a value close to the global optimum, which has been proven to be more effective comparing to SDP \cite{jiang20193}, since SDP may fail to yield a correct solution when strict constraints are enforced, as also noted in \cite{lesieutre2011examining}.
The second challenge is determining the appropriate prior information.
When the MAP is applied to multi-robot relative localization, the prior information pertains specifically to the prior probability density of the robots' relative positions and orientations at the initial time instant \cite[Section 4.3]{barfoot2024state}, which has a significant impact on the estimation accuracy \cite{van2021bayesian}.
We conduct a simulation to demonstrate the effects of the prior density on localization accuracy, as shown in Figure \ref{fig prior effect}.
In order to address this challenge, some works such as \cite{vezzani2017memory} and \cite{cossette2021relative} assume that the prior density is a Gaussian distribution with known mean and covariance matrix. 
However, this assumption may lack theoretical support, and even if it holds, obtaining the mean and covariance matrix might not be feasible.
Authors in \cite{nguyen2015bayesian} propose a form of approximate Bayesian computation (ABC) algorithm to avoid estimating intractable distribution by approximating the posterior density based on a set of samples.
Nonetheless, ABC requires a vast number of samples, making it impractical when sampling is computationally expensive \cite{papamakarios2019sequential, alsing2019fast}.
Lastly, since MAP is a batch estimator, it determines a set of relative positions and orientations over a time window simultaneously rather than sequentially. 
Although MAP generally results in estimation with superior accuracy, the number of states requiring estimation increases as robots move, i.e., the time window expands.
The increase of the problem's dimensionality leads to higher computational costs, which may become impractical in real-world scenarios.

\begin{figure*}[hbtp]
    \centering
    \includegraphics[width=0.98\textwidth]{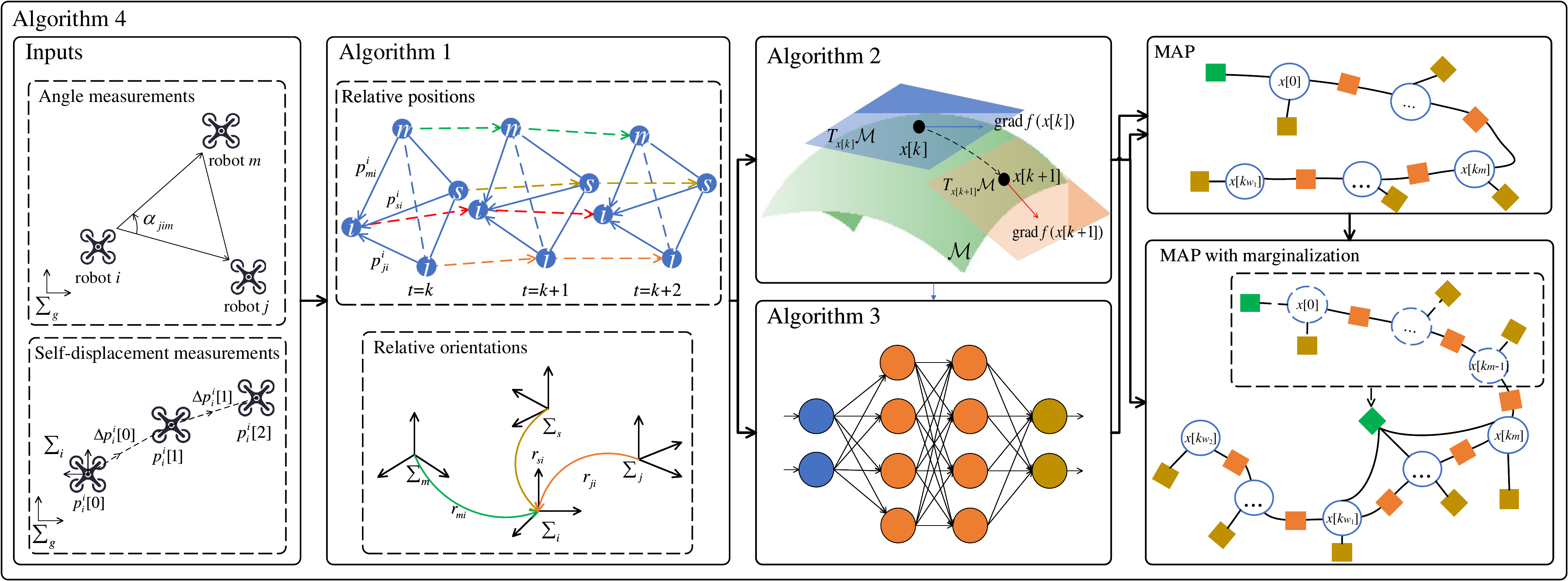}
    \caption{
      Overview of the proposed relative localization framework.
      Algorithm \ref{al general} formulates the relative localization problem at a single time instant as a linear equation based on angle and self-displacement measurements.
      Algorithm \ref{al wtls} enhances robustness by reformulating this linear equation as a WTLS optimization problem on manifolds.
      For improved accuracy over multiple time instants, a MAP estimator is employed in Algorithm \ref{al total}, where Algorithm \ref{al nde} approximates the required prior probability density and the output of Algorithm \ref{al wtls} serves as the initial values for the MAP optimization problem.}
    \label{fig str}
  \end{figure*}

Motivated by the above discussion, in this paper we aim to propose a novel relative localization framework based on interior angle and self-displacement measurements.
The overall structure of the proposed relative localization framework is shown in Figure \ref{fig str}.
We choose angle measurements for two reasons.
First, the results based on angle measurements can be extended to the cases with bearing or distance measurements. 
Second, there exist sensors capable of measuring angles, which has a potential application in industry \cite{buehrer2018collaborative}.
The main contributions are summarized as follows.

Firstly, we propose a linear relative localization theory, including a distributed linear relative localization algorithm (Algorithm \ref{al general}), and a sufficient condition for relative localizability (Theorem \ref{th loc cond}).
The linear relative localization algorithm allows robots to determine their relative positions and orientations by purely solving linear equations.

Secondly, we transform the linear computation process of the linear relative localization algorithm into a WTLS optimization problem on manifolds (Algorithm \ref{al wtls}), yielding more accurate estimation.
These WTLS estimation results are used as initial values, thereby reducing the risk of the convergence to local optima when finding the optimal solution of the MAP optimization problem.

Thirdly, to obtain the prior density of initial relative positions and orientations required by the MAP estimator, we develop a NDE (Algorithm \ref{al nde}) by leveraging the independence on prior information and low computational cost of the linear relative localization algorithm.

Lastly, we propose a marginalization mechanism to prevent the increasing the number of states when solving the MAP optimization problem, ensuring that the computational cost remains constant as the robots continue to move.
By utilizing the NDE estimator, the WTLS estimator and the marginalization mechanism, three key challenges associated with the MAP estimator are addressed (Algorithm \ref{al total}).

The rest of this paper is organized as follows. 
Preliminaries and problem formulations are introduced in Section \ref{sec 2}.
Section \ref{3dsection} presents the linear relative localization theory.
Section \ref{sec map} presents the novel relative localization algorithm.
Section \ref{sec ext} shows extensions of the proposed relative localization framework.
Simulations and real-world experiments are shown in Section \ref{sec sim}.

\section{Preliminaries and problem formulation}
\label{sec 2}

\subsection{Notations}

Table \ref{tab nomenclature} provides a nomenclature of key symbols used throughout this paper.

\begin{table}[tbhp]
    \centering
    \caption{Nomenclature of symbols used in this paper}
    \label{tab nomenclature}
    \begin{tabular}{m{0.15\textwidth}<{\centering}m{0.28\textwidth}<{\centering}}
      \toprule
      Symbols & Meaning \\
      \midrule
      $\tetrahedron ijms$ & Tetrahedron formed by robots $i, j, m, s$ \\[1ex]
      $\triangle ijm$ & Triangle formed by robots $i, j, m$ \\[1ex]
      $\sum_g$, $\sum_i$ & Global coordinate frame and local coordinate frame of robot $i$ \\[1ex]
      $p_{ji}^i \in \mathbb{R}^3$ & Relative positions of $j$ with respect to $i$ in $\sum_i$ \\[1ex]
      $r_{ji} \in \mathcal{S}^1$ & Relative orientation from $\sum_j$ to $\sum_i$ \\[1ex]
      $\alpha_{(\cdot)}$ & Angle measurement \\[1ex]
      $\Delta p_i^i \in \mathbb{R}^3$ & Self-displacement measurement of robot $i$ in $\sum_i$ \\[1ex]
      $x$ & General representation of relative positions and orientations of all robots \\[1ex]
      $\check{x}$, $\hat{x}$ & Estimation of $x$ obtained by Algorithm \ref{al wtls} and Algorithm \ref{al total}, respectively \\[1ex]
      A letter with an upper \textasciitilde, such as $\tilde{A}$ & True value plus noise \\[1ex]
      $\mathcal{G}$, $\mathcal{E}$, $\mathcal{V}$, $\mathcal{N}_{i}$ & Graph, sets of edges, vertices, and neighbors of vertex $i$, respectively \\[1ex]
      $\mathcal{A}$, $\mathcal{T}$ & Sets of angle measurements and tetrahedra, respectively \\[1ex]
	  $\mathcal{S}^1$ & Stiefel manifold \\[1ex]
	  $I_n$ & The $n$-by-$n$ identity matrix \\[1ex]
	  $\small{R(\theta) = \begin{bsmallmatrix} \cos\theta & -\sin\theta \\ \sin\theta & \cos\theta \end{bsmallmatrix}}$ & 2-D rotation matrix with rotating angle $\theta$ \\[1ex]
      \bottomrule
    \end{tabular}
  \end{table} 

\subsection{Interior angle and self-displacement measurements}

\begin{figure}[tbhp]
    \centering
    \subfigure[Interior angle $\alpha_{jim}$]
    {
        \includegraphics[width=0.2\textwidth]{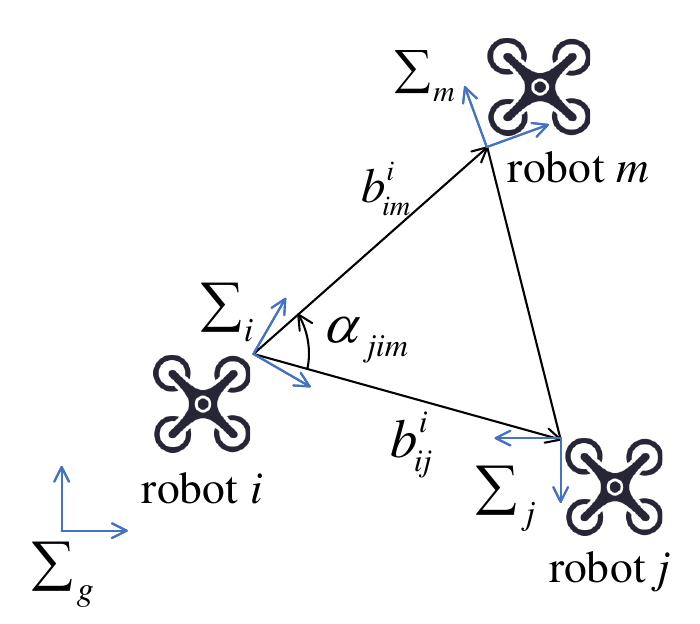}
    }
    \subfigure[Self-displacement $\Delta p_i^i$]
    {
        \includegraphics[width=0.2\textwidth]{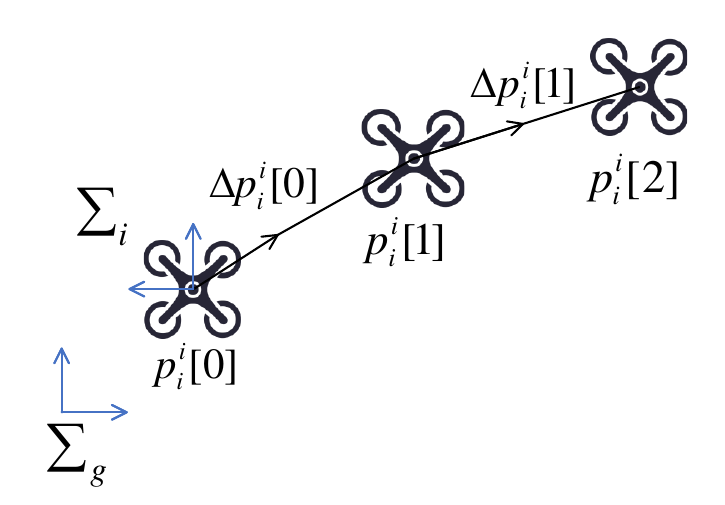}
    }
    \caption{Illustration of measurements}
	\label{fig mea def}
\end{figure}

As shown in Figure \ref{fig mea def}(a), when three robots are coplanar, e.g., positioned on the $XOY$ plane, we assume that each robot $i$ can measure the signed angle $\alpha_{jim} \in [-\pi, \pi)$ with respect to its neighbor robots $m, j \in \mathcal{V}$ under the counterclockwise direction, using on-board sensors with unique identifiers, such as antenna arrays \cite{sambu2022experimental}.
Unlike typical angles constrained to $[0,\pi]$, the angle defined here is calculated under a specific direction, which we refer to as a signed angle and can be calculated as follows
\begin{equation}
	\label{anglecalcu}
	\alpha_{jim} = 
	\begin{cases}
		\arccos(b_{im}\T b_{ij}), ~\text{if}~ b_{im}\T R_z(\frac{\pi}{2})b_{ij}> 0,\\
		-\arccos(b_{im}\T b_{ij}), ~\text{otherwise},
	\end{cases}
\end{equation}
where $b_{ij} = \frac{p_j-p_i}{\|p_j-p_i\|}$ is the bearing direction vector from robot $i$ to robot $j$ which is a unit vector, 
$j, m \in \mathcal{N}_i$, 
$R_z(\frac{\pi}{2}) = \begin{bsmallmatrix}
	R(\frac{\pi}{2}) & 0_{2\times 1} \\
	0_{1\times 2} & 1
\end{bsmallmatrix}$,
and $R_z(\frac{\pi}{2})b_{ij}$ represents the unit vector obtained by rotating $b_{ij}$ counterclockwise by $\frac{\pi}{2}$ on the $XOY$ plane. 
Note that since $b_{im}\T b_{ij}=(b_{im}^{i})\T b^i_{ij}$, $\alpha_{jim}$ is independent of specific coordinate frames. 
We also assume that the angles measured by robot $i$ are associated with the unique identifiers of $i$'s neighbor robots.

The other measurement of each robot $i$ is the self-displacement $\Delta p_i^i[k] = p_i^i[k+1] - p_i^i[k]$, which is defined as the displacement from the robot's current position $p^i_i[k]$ to its next position $p^i_i[k+1]$ in its own local coordinate frame $\sum_i$, as shown in Figure \ref{fig mea def}(b). 
With the recent development in sensing technology, robots' self-displacement measurements can be obtained by several ways, such as the visual–inertial odometry \cite{qin2018vins} and optical flow \cite{nguyen2019persistently}.
Each robot's odometry coordinate frame is created and fixed at initialization \cite{taketomi2017visual} and serves as the robot's local frame.
Since the transformation between the body frame and local frame can be measured by the odometry \cite{nguyen2023relative} or inertial measurement unit (IMU) \cite{ahmad2013reviews}, we assume that robot $i$'s body frame is equivalent to its local frame.
Besides, we assume that each robot's local frame has a common $Z$-axis direction pointing to the reverse direction of gravity \cite{qin2018vins, xu2022fast}, but can have different $X$-axis and $Y$-axis directions.
In our implementation, the self-displacements are obtained using the visual-inertial system (VINS) \cite{qin2018vins}.

\subsection{Multi-robot system definition}
\label{sec msd}

Considering a multi-robot system with $n \in \mathbb{N}^{+}$ robots, we define a vertex set of robots $\mathcal{V}=\{1,\cdots,n\}$ and an angle set $\mathcal{A} \subset \mathcal{V} \times \mathcal{V} \times \mathcal{V} = \{(i,j,m) |i,j,m \in \mathcal{V}, i \neq j\neq m \}$.
For an angle set $\mathcal{A}_{\triangle ijm} = \{(i,j,m)\}$, if robots $i$, $j$ and $m$ can form a triangle and measure the interior angles within the triangle $\triangle ijm$, we say that $\mathcal{A}_{\triangle ijm}$ is a triangular angle set \cite[Section 2.3]{chen2022triangular}.
Now we give the definition of the tetrahedrally angle rigid set.
\begin{definition}
	\label{def angle set}
	\cite{chen20223d}
	For four robots $i$, $j$, $m$ and $s$, a set $\mathcal{T}_{\tetrahedron ijms} = \left\{\mathcal{A}_{\triangle ijm}, \mathcal{A}_{\triangle ijs}, \mathcal{A}_{\triangle ims}, \mathcal{A}_{\triangle jms} \right\}$ is a tetrahedrally angle rigid set if each element $\mathcal{A}_{(\cdot)} \in \mathcal{T}_{\tetrahedron ijms}$ is a triangular angle set, and $i, j, m, s \in \mathcal{V}$ are non-planar.
\end{definition}

For the overall multi-robot system, the tetrahedrally angle rigid set is defined as $\mathcal{T} = \mathcal{T}_{\tetrahedron i_1 j_1 m_1 s_1} \cup \cdots \cup \mathcal{T}_{\tetrahedron i_{n_t} j_{n_t} m_{n_t} s_{n_t}}$, where $n_t \in \mathbb{R}$ denotes the number of tetrahedral angle sets as defined in Definition \ref{def angle set} and if $\mathcal{T}_{\tetrahedron i_1 j_1 m_1 s_1} \subset \mathcal{T}$, there always exists at least one $\mathcal{T}_{\tetrahedron i_2 j_2 m_2 s_2} \subset \mathcal{T}$ such that $\mathcal{T}_{\tetrahedron i_1 j_1 m_1 s_1}$ and $\mathcal{T}_{\tetrahedron i_2 j_2 m_2 s_2}$ share at least two vertices.
The combination of $\mathcal{V}$ and $\mathcal{T}$ is called the multi-robot system's topology, which we denote by $(\mathcal{V}, \mathcal{T})$.
To formalize the communication relationship, we introduce an undirected graph denoted as $\mathcal{G}(\mathcal{V}, \mathcal{E})$ to represent the communication graph, where $(i, j) \in \mathcal{E}$ if $i$, $j$ belong to a common triangular angle set.

\subsection{Problem formulation}

\begin{figure}[tbhp]
	\centering
	\includegraphics[width=0.35\textwidth]{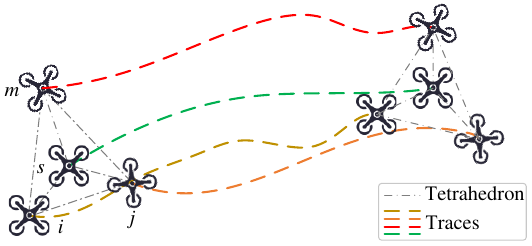}
	\caption{The scenario with four robots}
	\label{fig prob form}
	\centering
\end{figure}

Assume that each robot of the multi-robot system, denoted by $i$, can measure its self-displacement $\Delta p^i_i$ in its local frame $\sum_i$, the angles $\alpha_{j i m}$ with respect to its neighbors $j, m \in \mathcal{N}_i$,  and the angles formed by the inter-robot bearing and robots' $Z$-axis.
A group of four robots can form a tetrahedron, and the set $\mathcal{T}_{\tetrahedron ijms}$ is a tetrahedrally angle rigid set.
As shown in Figure \ref{fig prob form}, the objective of this paper is to estimate the relative positions and orientations of the robots using the measured angles and self-displacements, which can be noisy, as robots move.

\section{Linear relative localization theory}
\label{3dsection}

In this section, we introduce the linear relative localization theory, including a sufficient condition for relative localizability and the linear relative localization algorithm.
In the following, we first present the necessary assumptions and definition.

\begin{assumption}
\label{ass not collinear}
	For all time instants $k\in\mathbb{N}$, three conditions hold, namely,
	a) $i$, $j$, $m$, $s$ should not be coplanar, and
	b) $i$, $j'$, $s'$ should not be collinear, and
	c) $i$, $m'$, $s'$ should not be collinear.
\end{assumption}

In Assumption \ref{ass not collinear}, $j'$, $m'$ and $s'$ denote the orthogonal projection of robot $j$, $m$ and $s$ onto the $X O_i Y$ plane, respectively.
The necessity of this assumption arises from the fundamental topological relationship between robots' positions along the $Z$-axis, which is utilized by Algorithm \ref{al general}, as thoroughly discussed in \cite{chen2022maneuvering}.
To detect violations of these assumptions in real-world scenarios, one can check whether $\max(|\alpha_{miZ}[k] - \alpha_{jiZ}[k]|, |\alpha_{miZ}[k] - \alpha_{siZ}[k]|, |\alpha_{jiZ}[k] - \alpha_{siZ}[k]|) \geq \epsilon_1$ holds, where $\epsilon_1 \in \mathbb{R}$ denotes a predefined threshold.
If this condition fails, Algorithm \ref{al general} cannot be directly applied. 
In such cases, the relative position can be approximated using displacement information, as shown in \eqref{eq aligned rela} and \eqref{eq k plus plus}.
Even if coplanar configurations in Assumption \ref{ass not collinear} occur repeatedly over a time duration (like for drones forming a flat formation), our proposed framework is still efficient and is automatically reduced to relative localization in 2D \cite{ourtro}, which is sufficiently applicable for real-world scenarios.

\begin{assumption}
	\label{ass com}
	If four robots forming a tetrahedron can measure interior angles with respect to their neighbors, then they can communicate with each other.
\end{assumption}


\begin{definition}
	\label{def sim}
	Two tetrahedra $\tetrahedron ijms [k_1]$ and $\tetrahedron ijms [k_2]$ formed by robots $i$, $j$, $m$ and $s$ at time instants $t=k_1$ and $t=k_2$ respectively are said to be strongly similar if all the corresponding interior angles and the sign of the signed volume of the two tetrahedra are the same, which we denote by $\tetrahedron ijms [k_1] \simeq \tetrahedron ijms [k_2]$. 
	The case where they are not strongly similar is denoted by $\tetrahedron ijms [k_1] \nsim \tetrahedron ijms [k_2]$.
\end{definition}

\begin{assumption}
	\label{ass sim}
	For a given sequence of consecutive time instants, the tetrahedron $\tetrahedron ijms$ formed by the four robots $i$, $j$, $m$ and $s$ is not strongly similar to one another at any two distinct time instants.
\end{assumption}

Assumption \ref{ass sim} is required to guarantee that linear equations \eqref{basiccalculationr3D}, \eqref{eq th2} in Theorems \ref{theorem8} and \ref{thm  un lin sol} have a unique solution without singularity.
This issue is commonly referred to as the persistent excitation problem \cite{chen2022simultaneous, lin2016distributed}.
To detect violations of this assumption in real-world scenarios, one can check $\| \alpha[k] - \alpha[k+1] \| \geq \epsilon_2$, where $\epsilon_2 \in \mathbb{R}$ denotes a predefined threshold.
If the condition is not satisfied, the strong similarity is detected and Algorithm \ref{al general} cannot be directly applied. 
In such cases, the relative position can be approximated by \eqref{eq aligned rela} or \eqref{eq k plus plus}.

\subsection{Angle-induced linear equation}

\begin{figure}[tbhp]
	\centering
	\includegraphics[width=0.3\textwidth]{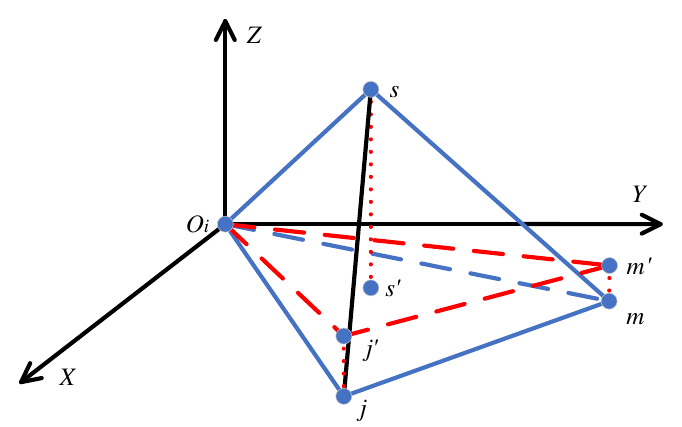}
	\caption{ $\tetrahedron ijms$ and geometric relation between $i,j,m,s$ and  $j',m',s'$}
	\label{AR/fig-angle30}
	\centering
\end{figure}

According to \cite[Section IV.A]{chen2022maneuvering} and \cite{chen20223d}, an angle-induced six-dimensional linear equation among four robots $i$, $j$, $m$, $s$ forming $\tetrahedron ijms$ is given by
\begin{equation}
	\label{linearequation3d}
	A_j^{\tetrahedron ijms}(\alpha)p^i_{ji}+A_m^{\tetrahedron ijms}(\alpha)p^i_{mi}+A_s^{\tetrahedron ijms}(\alpha)p^i_{si} = 0,
\end{equation}
where
$
A_j^{\tetrahedron ijms}(\alpha) = 
\begin{bsmallmatrix}
	\sin\alpha_{s'j'i}I_2 & 0_{2\times 1} \\
	0_{1\times 2} & \cos\alpha_{miZ} \sin \alpha_{mji} \\
	0_{2\times 2} & 0_{2\times 1} \\
	0_{1\times 2} & \cos\alpha_{siZ}\sin\alpha_{sji}
\end{bsmallmatrix} \in\mathbb{R}^{6\times 3}
$,
$
A_m^{\tetrahedron ijms}(\alpha) = 
\begin{bsmallmatrix}
	0_{2\times 2} & 0_{2\times 1} \\
	0_{1\times 2}& -\cos\alpha_{jiZ}\sin\alpha_{jmi}\\
	\sin\alpha_{s'm'i}I_2 & 0_{2\times 1}\\
	0_{1\times 2}& 0_{1\times 1}\\
\end{bsmallmatrix} \in\mathbb{R}^{6\times 3}
$,
$
A_s^{\tetrahedron ijms}(\alpha) = 
\begin{bsmallmatrix}
	-\sin\alpha_{j's'i}R(\alpha_{s'ij'}) & 0_{2\times 1}\\
	0_{1\times 2} & 0_{1\times 1} \\
	-\sin\alpha_{m's'i}R(\alpha_{s'im'}) & 0_{2\times 1}\\
	0_{1\times 2}& -\cos\alpha_{jiZ}\sin\alpha_{jsi}\\
\end{bsmallmatrix} \in\mathbb{R}^{6\times 3}
$.
Here, $j'$, $m'$, $s'$ are the orthogonal projection of $j,m,s$ onto the $X O_i Y$ plane, respectively, as shown in Figure \ref{AR/fig-angle30}, and $\alpha$ denotes the vector of interior angles in $\tetrahedron ijms$.
In the following, we denote $A_j^{\tetrahedron ijms}(\alpha)$, $A_m^{\tetrahedron ijms}(\alpha)$ and $A_s^{\tetrahedron ijms}(\alpha)$ as $A_j$, $A_m$, $A_s$ respectively for brevity.

Indeed, $A_j$, $A_m$, $A_s$ contain interior angles such as $\alpha_{mji},\alpha_{sji},\alpha_{jmi},\alpha_{jsi}$ within four surface triangles of $\tetrahedron ijms$, $\alpha_{s'j'i},\alpha_{s'm'i},\alpha_{j's'i},\alpha_{m's'i},\alpha_{s'im'},\alpha_{s'ij'}$ within $\triangle ij's'$, $\triangle im's'$, and $\alpha_{miZ},\alpha_{siZ},\alpha_{jiZ}$ between $\overrightarrow{iZ}$ and $\overrightarrow{ij},\overrightarrow{im},\overrightarrow{is}$, where $\alpha_{s' i j'}$, $\alpha_{s' i m'}$ are signed angles.
Since the $Z$-axis direction can be easily obtained via VINS, these angles can be determined by the four robots' sensor measurements according to \cite[Section IV.A]{chen2022maneuvering}, which indicates that $A_j$, $A_m$, $A_s$ are known.
In  \cite[Lemma 2]{chen2022maneuvering}, we proved that the rank of $[A_j,A_m,A_s]$ is six. 
We now show that the rank of the concatenation of any two of  $A_j,A_m,A_s$ is also six.
\begin{lemma}
	\label{lemma1}
	If Assumption \ref{ass not collinear} holds, then $\mathrm{rank}([A_j,A_m])=\mathrm{rank}([A_j,A_s])=\mathrm{rank}([A_m,A_s])=6$ and moreover, 
	\begin{align}
		&\begin{bmatrix}
			p^i_{ji}\\
			p^i_{mi}
		\end{bmatrix}=-\left[A_j, A_m\right]^{-1}A_s p^i_{si}. \label{inverse3d}
	\end{align}
\end{lemma}

The proof of Lemma \ref{lemma1} is provided in Appendix \hyperref[sec pro lemma1]{I}.
Lemma \ref{lemma1} indicates that when the coefficient matrices $A_j$, $A_m$, $A_s$ are known, one only needs to determine one unknown relative position in each $\tetrahedron ijms$ and then the remaining relative positions are all known.

\subsection{Relative localization among four robots}
\label{sub sec four r}

In this section, we discuss relative localization among the minimum number of robots, i.e., four robots.
We first consider a simple case where the robots' local coordinate frames are aligned, meaning that their coordinate frames' three axes have the same direction.
Subsequently, we address the more general case where the robots' frames are not aligned.

\subsubsection{The case with aligned coordinate frames:}

\begin{figure}[tbhp]
	\centering\includegraphics[width=0.4\textwidth]{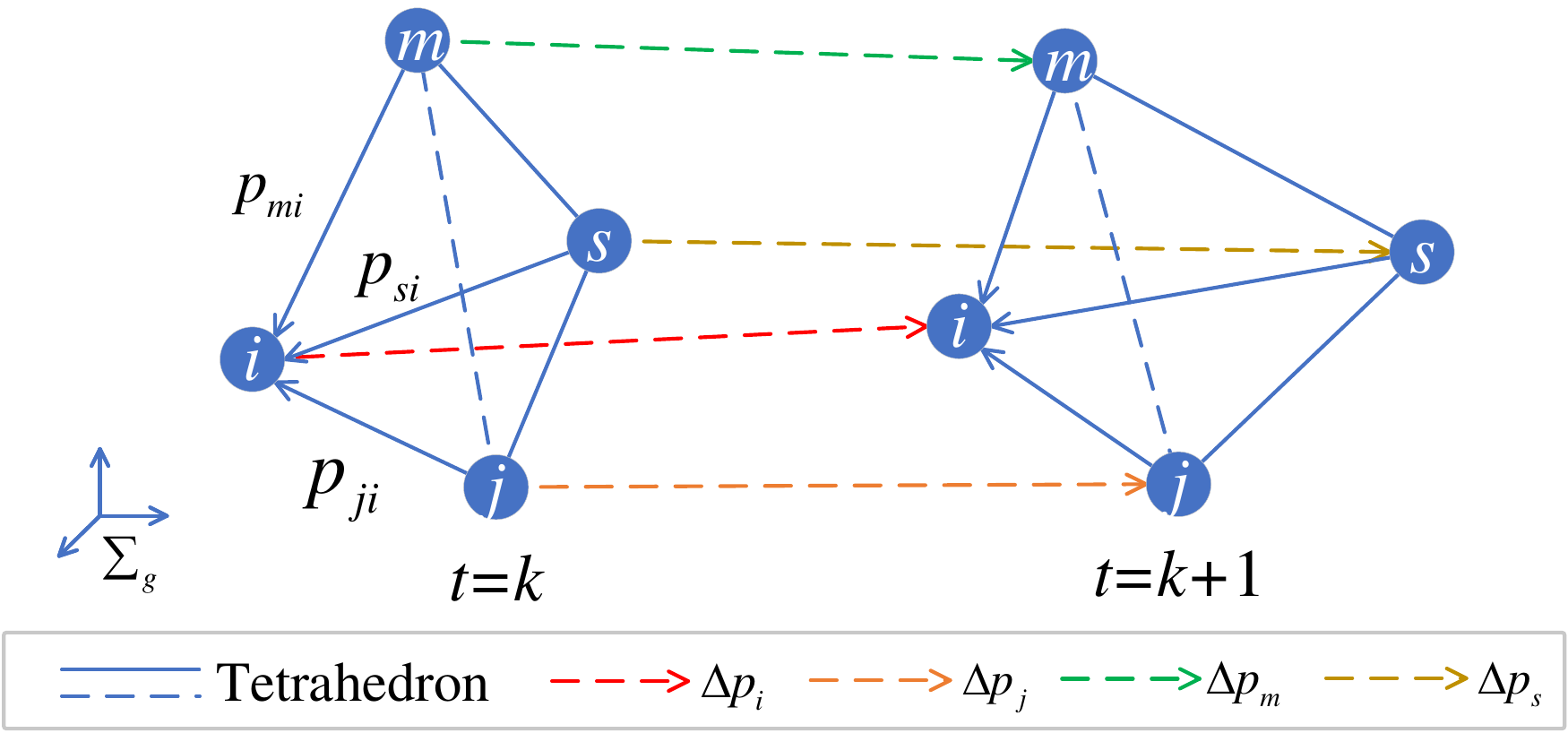}
	\caption{Relative positions among robots $i$, $j$, $m$ and $s$ with aligned coordinate frames}
	\label{3dobjective1}
\end{figure}

As shown in Figure \ref{3dobjective1}, according to the angle-induced equation $A_j[k]p_{ji}[k]+A_m[k]p_{mi}[k]+A_s[k]p_{si}[k]=0$ at $t=k$ and Lemma \ref{lemma1}, one has
\begin{equation}
	\label{eq aligned pji psi}
	\begin{aligned}
		& p_{ji}[k] = -[I_3,0_3] \left[ A_j[k], A_m[k] \right]^{-1} A_s[k] p_{si}[k], \\
		& p_{mi}[k] = -[0_3,I_3] \left[ A_j[k], A_m[k] \right]^{-1} A_s[k]p_{si}[k],
	\end{aligned}
\end{equation}
where we drop the superscript $i$ since robots share an aligned frame.
Similarly, at time instant $k+1$ the angle-induced equation is given by
\begin{equation}
	\label{eq linear eq kp}
	\begin{aligned}
	A_j[k+1] p_{ji}[k+1] + A_m[k+1]p_{mi}[k+1] \\
	+ A_s[k+1]p_{si}[k+1] = 0.
	\end{aligned}
\end{equation}
The relationship between $p_{ji}[k]$ and $p_{ji}[k+1]$ is given by
\begin{equation}
	\label{eq aligned rela}
	p_{ji}[k+1] = p_{ji}[k] + \Delta p_{ji}[k],
\end{equation}
where $\Delta p_{ji}[k] = \Delta p_{i}[k] - \Delta p_{j}[k]$ can be calculated by self-displacement measurements $\Delta p_i[k]$ and $\Delta p_j[k]$.
The relationships between $p_{mi}[k]$ and $p_{mi}[k+1]$, and between $p_{si}[k]$ and $p_{si}[k+1]$, are similar to \eqref{eq aligned rela}.
By combining \eqref{eq aligned pji psi}, \eqref{eq linear eq kp} and \eqref{eq aligned rela} one can obtain
\begin{equation}
	\label{eq aligned Q1}
	Q_1(k,k+1) p_{si}[k] = q_1(k,k+1),
\end{equation}
where 
$Q_1(k, k+1) 
	= A_s[k+1] - \big[\big.A_j[k+1], A_m[k+1]\big.\big] \big[\big.A_j[k], A_m[k] \big.\big]^{-1} A_s[k] \in \mathbb{R}^{6 \times 3}$,
$q_1(k, k+1) = 
- A_j[k+1] \Delta p_{ji}[k] 
- A_m[k+1] \Delta p_{mi}[k]
- A_s[k+1] \Delta p_{si}[k]
\in \mathbb{R}^{6}$.
Let $\Delta \alpha_{ijm}[k] = \alpha_{ijm}[k+1] - \alpha_{ijm}[k]$ denote the change in $\alpha_{ijm}$ from time instants $t=k$ to $t=k+1$.
Now, we present the following result.

\begin{theorem}
	\label{theorem8}
	Under Assumptions \ref{ass not collinear}-\ref{ass sim}, if one of the two conditions holds: \\
	a) one of $\Delta \alpha_{j's'i}[k]$, $\Delta \alpha_{s'j'i}[k]$ and $\Delta \alpha_{s'ij'}[k]$ is nonzero, and $\frac{\sin \alpha_{jsi}[k] \cos\alpha_{jiz}[k]}{\sin \alpha_{jsi}[k+1] \cos\alpha_{jiz}[k+1]} - \frac{\sin\alpha_{sji}[k] \cos \alpha_{siz}[k]}{\sin\alpha_{sji}[k+1] \cos \alpha_{siz}[k+1]} \not = 0$ are satisfied, and \\
	b) one of $\Delta \alpha_{m's'i}[k]$, $\Delta \alpha_{s'm'i}[k]$ and $\Delta \alpha_{s'im'}[k]$ is nonzero, and $\frac{\sin \alpha_{jmi}[k] \cos\alpha_{jiz}[k]}{\sin \alpha_{jmi}[k+1] \cos\alpha_{jiz}[k+1]} - \frac{\sin\alpha_{mji}[k] \cos \alpha_{miz}[k]}{\sin\alpha_{mji}[k+1] \cos \alpha_{miz}[k+1]} \not = 0$
	are satisfied, then the relative position $p_{si}[k]$ can be uniquely determined by
	\begin{equation}
		\begin{aligned}
		\label{basiccalculationr3D}
		p_{si}[k] 
		=& 
		\left( 
			Q\T_1(k,k+1) Q_1(k,k+1)
		\right)^{-1} \\
		& \cdot Q\T_1(k,k+1) q_1(k,k+1),	
	\end{aligned}
	\end{equation} 
	and similarly, $p_{ji}[k],p_{mi}[k]$, $p_{mi}[k+1]$, $p_{si}[k+1]$, $p_{ji}[k+1]$ are uniquely determined by using \eqref{eq aligned pji psi} and \eqref{eq aligned rela}. 
\end{theorem}

The proof of this theorem can be found in Appendix \hyperref[pro theorem8]{II}.

\begin{figure}[tbhp]
    \centering
    \subfigure[Relative positions]
    {
        \includegraphics[width=0.25\textwidth]{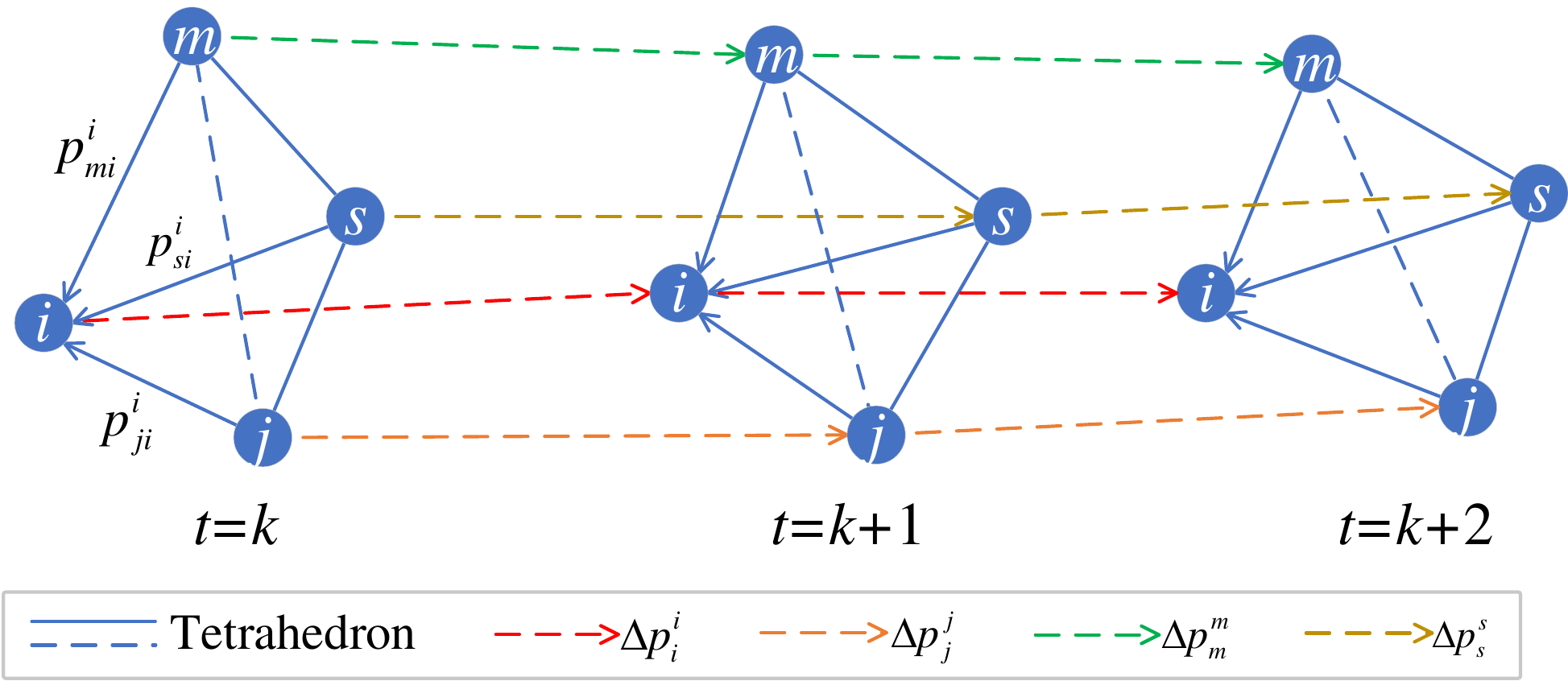}
    }
    \subfigure[Relative orientations]
    {
        \includegraphics[width=0.2\textwidth]{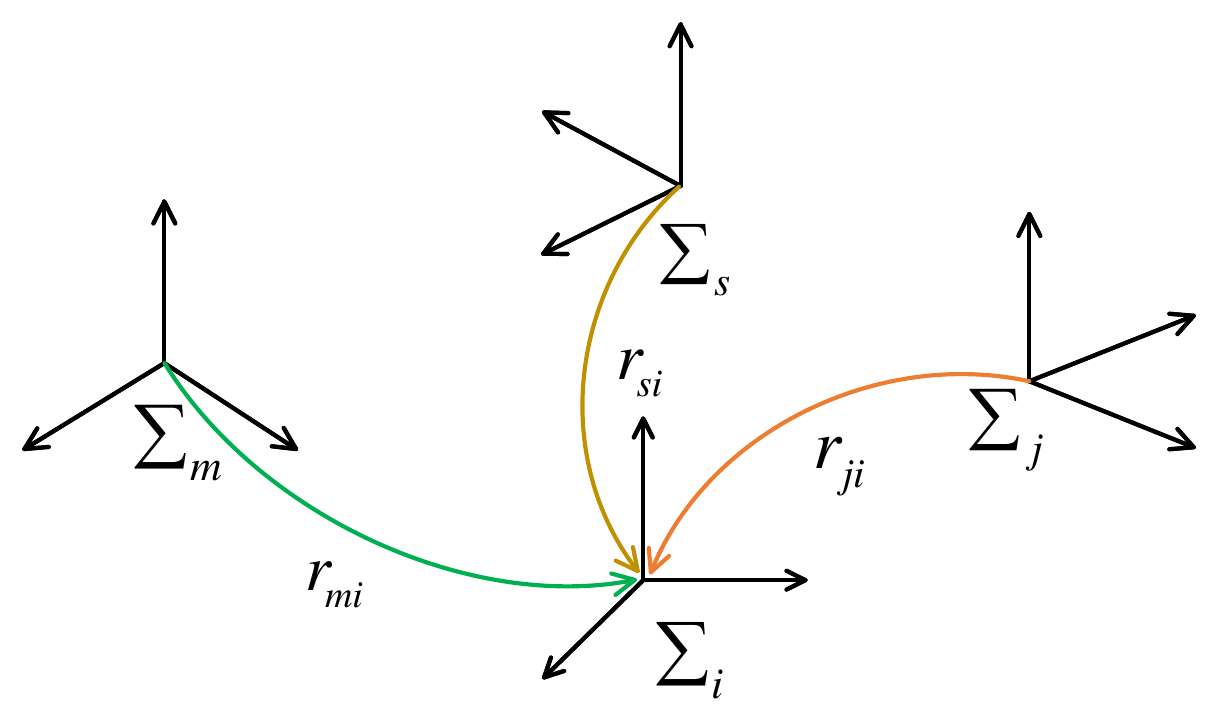}
    }
    \caption{Relative positions and orientations among robots $i$, $j$, $m$ and $s$ with unaligned coordinate frames}
    \label{fig th2}
\end{figure}

\subsubsection{The case with unaligned coordinate frames:}
\label{fourfollowerscase2}
In this case where the local frames of robots are unaligned, both relative positions and relative orientations, as illustrated in Figure \ref{fig th2}, need to be estimated.
Note that \eqref{eq aligned Q1} still holds if the positions and self-displacements are described in the same local frame.
For example, in robot $i$'s local coordinate frame, \eqref{eq aligned Q1} becomes
\begin{equation}
	\label{eq unaligned Q1}
	\begin{aligned}
		Q_{1}(k,k+1) p_{si}^i[k] = 
		- A_j[k+1] \Delta p^i_{ji}[k] \\
		- A_m[k+1] \Delta p^i_{mi}[k] 
		- A_s[k+1] \Delta p^i_{si}[k],
	\end{aligned}
\end{equation}
where $\Delta p^i_{ji}[k] = \Delta p^i_i[k] - \Delta p^i_j[k]$.
Here, $\Delta p^i_j[k]$ denotes robot $j$'s self-displacement at time instant $t=k$ described in robot $i$'s local frame, which is unknown.
Note that $\Delta p_j^j[k]$ is only available to robot $j$. 
Inspired by \cite{ourtro}, we then introduce a linear transformation to represent $\Delta p^i_j[k]$, i.e.,
\begin{equation}
	\label{eq una times}
	\begin{aligned}
	\Delta p^i_j[k] 
	=& R_z(\theta_j^i) \Delta p_j^j[k]
	= \begin{bsmallmatrix}
		(\Delta p_j^j[k])^{\times} & 0_{2 \times 1} \\
		0_{1 \times 2} & \Delta p_j^j[k](3)
	\end{bsmallmatrix}
	\begin{bsmallmatrix}
		r_{ji} \\
		1
	\end{bsmallmatrix},
	\end{aligned}
\end{equation}
where	
$
	(\Delta p_j^j[k])^{\times} = 
	\begin{bsmallmatrix}
		\Delta p_j^j[k](1) & -\Delta p_j^j[k](2) \\
		\Delta p_j^j[k](2) & \Delta p_j^j[k](1).
	\end{bsmallmatrix}
	\in \mathbb{R}^{2 \times 2}
$,
$
	r_{ji} = \left[\cos \theta_j^i, \sin \theta_j^i\right]\T \in \mathbb{R}^2
$,
$R_z(\theta_j^i) = \begin{bsmallmatrix}
	R(\theta_j^i) & 0_{2\times 1} \\
	0_{1\times 2} & 1
\end{bsmallmatrix} \in SO(3)$ is the rotation matrix around the $Z$-axis,
and $\theta_j^i$ denotes the rotation angle from $\sum_{j}$ to $\sum_{i}$.
Substituting \eqref{eq una times} into \eqref{eq unaligned Q1} yields
\begin{equation}
	\label{eq una Q1}
	Q_{21}(k,  k+1) 
	\begin{bmatrix}
		p^i_{si}[k] \\
		r_{ji} \\
		r_{mi} \\
		r_{si}
	\end{bmatrix}
	= q_{21}(k,  k+1),
\end{equation}
where 
$
	Q_{21}(k,  k+1) =
	\big[\big. 
		Q_{1}(k,k+1) 
		, -A_{j, 1:2}[k+1](\Delta p_j^j[k])^{\times} 
		, -A_{m, 1:2}[k+1](\Delta p_m^m[k])^{\times} 
		, -A_{s, 1:2}[k+1](\Delta p_s^s[k])^{\times}
	\big.\big]
	\in \mathbb{R}^{6 \times 9}
$,
$
	q_{21}(k,  k+1) = 
	- \big(\big. A_j[k+1] + A_m[k+1] + A_s[k+1] \big.\big) \Delta p^i_{i}[k] 
	+ A_{j, 3}[k+1] \Delta p_j^j[k](3)
	+ A_{m, 3}[k+1] \Delta p_m^m[k](3)
	+ A_{s, 3}[k+1] \Delta p_s^s[k](3)
	\in \mathbb{R}^{6}
$.
Here, $A_{j, 1:2}[k+1] \in \mathbb{R}^{6 \times 2}$ and $A_{j, 3}[k+1] \in \mathbb{R}^{6 \times 1}$ satisfy $\big[\big. A_{j, 1:2}[k+1], A_{j, 3}[k+1] \big.\big] = A_{j}[k+1]$.

There are four unknown variables in \eqref{eq una Q1}, namely, $p^i_{si}[k]\in\mathbb{R}^3$, $r_{ji}\in\mathbb{R}^2$, $r_{mi}\in\mathbb{R}^2$, $r_{si}\in\mathbb{R}^2$, totally nine dimensions.
Since there are only six independent linear equations in \eqref{eq una Q1}, at least three additional linear equations are required to determine these unknown variables.
As shown in Figure \ref{fig th2}, we extend the angle-induced linear equation to one time instant forward, i.e., $t=k+2$, under which one has
\begin{equation}
	\label{eq una angle kpp}
	\begin{aligned}
		& A_j[k+2] p^i_{ji}[k+2] + A_m[k+2]p^i_{mi}[k+2] \\
		&+ A_s[k+2]p^i_{si}[k+2] = 0.
	\end{aligned}
\end{equation}
Based on the self-displacement measurements, one has
\begin{equation}
	\label{eq k plus plus}
	\begin{aligned}
		& p^i_{ji}[k+2] = p^i_{ji}[k] + \Delta p_{ji}^i[k] + \Delta p_{ji}^i[k+1], \\
		& \Delta p_{ji}^i[k+1] = \Delta p_{i}^i[k+1] - (\Delta p_{j}^j[k+1])^{\times} \cdot r_{ji},
	\end{aligned}
\end{equation}
which also holds for $p^i_{mi}[k+2]$, $\Delta p_{mi}^i[k+1]$, and $p^i_{si}[k+2]$, $\Delta p_{si}^i[k+1]$.
Note that in robot $i$'s local frame $\sum_i$, \eqref{eq aligned pji psi} remains valid if all vectors are labeled with a superscript $i$.
Then substituting \eqref{eq aligned pji psi} and \eqref{eq k plus plus} into \eqref{eq una angle kpp} yields
\begin{equation}
	\label{eq una Q22}
	Q_{22}(k:k+2)
	\begin{bmatrix}
		p^i_{si}[k] \\
		r_{ji} \\
		r_{mi} \\
		r_{si}
	\end{bmatrix}
	= q_{22}(k:k+2),
\end{equation}
where
$
	Q_{22}(k : k+2) = 
	\big[\big.
		Q_{1}(k,k+2) 
		, -A_{j, 1:2}[k+2] \big(\big. \Delta p_j^j[k] + \Delta p_j^j[k+1] \big.\big)^{\times} 
		, -A_{m, 1:2}[k+2] \big(\big. \Delta p_m^m[k]+\Delta p_m^m[k+1] \big.\big)^{\times} 
		, -A_{s, 1:2}[k+2] \big(\big. \Delta p_s^s[k]+\Delta p_s^s[k+1] \big.\big)^{\times}
		\big.\big]
	\in \mathbb{R}^{6 \times 9}
$,
$
	q_{22}(k:k+2) = 
	- \big(\big. A_j[k+2] + A_m[k+2] + A_s[k+2] \big.\big) \big(\big. \Delta p^i_{i}[k] + \Delta p^i_{i}[k+1] \big.\big)
	+ A_{j, 3}[k+2] \big(\big. \Delta p_j^j[k](3) + \Delta p_j^j[k+1](3) \big.\big)
	+ A_{m, 3}[k+2] \big(\big.\Delta p_m^m[k](3) + \Delta p_m^m[k+1](3) \big.\big)
	+ A_{s, 3}[k+2] \big(\big.\Delta p_s^s[k](3) + \Delta p_s^s[k+1](3) \big.\big)
	\in \mathbb{R}^{6}
$.
Combining \eqref{eq una Q1} and \eqref{eq una Q22} yields
\begin{equation}
	\label{eq Q2}
	Q_2(k:k+2) 
	\begin{bmatrix}
		p^i_{si}[k] \\
		r_{ji} \\
		r_{mi} \\
		r_{si}
	\end{bmatrix}
	= q_{2}(k:k+2),
\end{equation}
where $Q_2(k:k+2) = \big[\big. Q\T_{21}(k, k+1), Q\T_{22}(k:k+2) \big.\big]\T \in \mathbb{R}^{12 \times 9}$ and $q_2(k:k+2) = \big[\big. q\T_{21}(k, k+1), q\T_{22}(k:k+2) \big.\big]\T \in \mathbb{R}^{12}$.

\begin{theorem}
	\label{thm  un lin sol}
	Under Assumptions \ref{ass not collinear}-\ref{ass sim}, if $Q\T_2(k:k+2) Q_2(k:k+2)$ is nonsingular, the relative position $p^i_{si}[k]$ and relative orientations $r_{ji}$, $r_{mi}$ and $r_{si}$ can be uniquely determined by 
	\begin{equation}
		\label{eq th2}
		\begin{aligned}
			\begin{bmatrix}
				p^i_{si}[k] \\
				r_{ji} \\
				r_{mi} \\
				r_{si}
			\end{bmatrix}
			= \big(\big. Q\T_2(k:k+2) Q_2(k:k+2) \big.\big)^{-1} \\
			\cdot Q\T_2(k:k+2) q_2(k:k+2),
		\end{aligned}
	\end{equation}
	and similarly, other relative positions $p_{ji}^i$, $p_{mi}^i$, $p_{si}^i$ at time instants $t=k$, $t=k+1$, $t=k+2$ are also uniquely determined by using \eqref{eq aligned pji psi} and \eqref{eq k plus plus}.
\end{theorem}

The proof of Theorem \ref{thm  un lin sol} can be found in Appendix \hyperref[sec pro th2]{III}.
The algebraic condition that $Q\T_2(k:k+2) Q_2(k:k+2)$ is nonsingular serves as a broad sufficient criterion for the solvability of \eqref{eq Q2}.
Despite our best efforts, it is challenging to devise a more explicit condition due to the large dimensions and intricate forms of the elements in $Q_2(k:k+2)$.
Nevertheless, from the perspective of persistent excitation, this nonsingularity generally holds given that the four robots have different movements.

\begin{remark}
	It is worth noting that $r_{ji}$, $r_{mi}$ and $r_{si}$ in Theorem \ref{thm  un lin sol} refer to the rotation from robots' local frames $\sum_j$, $\sum_m$, $\sum_s$ to robot $i$'s local frame $\sum_i$, respectively, which differ from the relative attitudes between robots.
	A robot's attitude represents its heading at a given time instant, while its local frame is established at initialization, whose orientation cannot be measured directly.
	However, if the relative orientation is determined, by using the measurements obtained by IMU (if equipped), the relative attitudes can also be determined.
\end{remark}

\subsection{Relative localizability and localization for general multi-robot systems}

\label{sub sec localizability}

In this subsection, we discuss the sufficient condition for relative localizability \cite{ourtro} of general multi-robot systems, extending beyond systems composed of only four robots $i$, $j$, $m$, $s$.
We now propose a sufficient topological condition for the relative localizability.

\begin{theorem}
	\label{th loc cond}
	Under Assumptions \ref{ass not collinear}-\ref{ass sim}, if $\mathcal{T}$ is a tetrahedrally angle rigid set, then $(\mathcal{V},\mathcal{T})$ is relatively localizable.
\end{theorem}

\begin{figure}[tbhp]
    \color{blue}
    \centering\includegraphics[width=0.3\textwidth]{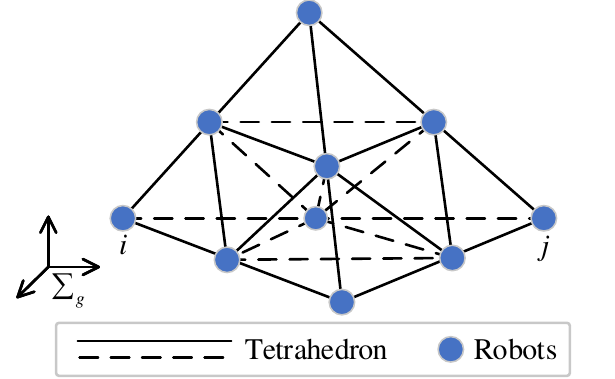}
    \caption{Geometric representation of a large-scale multi-robot system}
    \label{fig theorem3}
\end{figure}

The proof of this theorem can be found in Appendix \hyperref[sec pro loc cond]{IV}.
Theorem \ref{th loc cond} introduces the concept of relative localizability to characterize whether the relative positions and orientations between any pair of robots in a large-scale multi-robot system can be uniquely determined.
The theorem provides a sufficient condition for relative localizability, i.e., if the multi-robot system is tetrahedrally angle rigid, then the system is relatively localizable.
For example, as shown in Figure \ref{fig theorem3}, the multi-robot system is tetrahedrally angle rigid, which can be decomposed into several tetrahedra.
Within each tetrahedron, relative localization of robots is achieved using Algorithm \ref{al general}.
For relative localization between robots in different tetrahedra, such as robots $i$ and $j$ in Figure \ref{fig theorem3}, although they are not within the same tetrahedron, a simple coordinate transformation can still be employed to obtain the relative localization through intermediate neighbor robots.
The transformation of the relative orientation is similar.

The design and selection of the tetrahedra must be done carefully to ensure that the multi-robot system is relatively localizable.
As described in Theorem \ref{th loc cond}, if the system formed by the set of tetrahedra is tetrahedrally angle rigid, then the relative positions and orientations between any pair of robots in a multi-robot system can be uniquely determined.
To be brief, such a tetrahedrally angle rigid set can be constructed through adding repeatedly new vertices, namely neighbor robots, to the original small angle rigid set with carefully chosen angle constraints.
We refer readers to \cite[Section 3]{chen2023angle} for more details.

From Section \ref{sub sec four r} one can see that the relative localizability of a multi-robot system depends on not only the network topology, but also the number of time instants used to establish the linear equations. 
Thus, we define a $d$-step sensor measurement vector $\omega_d = [\alpha_d\T, \upsilon_d\T]\T$, where the column vectors $\alpha_d$ and $\upsilon_d$ consist of all the interior angles and self-displacements measured by the robots within $d$ consecutive instants from $t=k$ to $t=k+d-1$, respectively.
The combination of $\mathcal{V}$, $\mathcal{T}$, and $\omega_d$ is called a multi-robot system with $d$-step measurements, which we denote by $(\mathcal{V}, \mathcal{T}, \omega_d)$.
Now we give the definition of relative localizability.

\begin{definition}
	A multi-robot system $(\mathcal{V},\mathcal{T}, \omega_d)$ is said to be $d$-step relatively localizable if all the relative positions and orientations among the robots in $\mathcal{V}$ are uniquely determined in the robots' local coordinate frames based on the measurement $\omega_d$. 
\end{definition}

We then provide a detailed condition for relative localizability, taking into account the number of time instants.

\begin{proposition}
	Assume the conclusion in Theorem \ref{th loc cond} holds.
	a) If the robots in $\mathcal{V}$ have aligned coordinate frames, then $(\mathcal{V},\mathcal{T}, \omega_d)$ is 2-step relatively localizable.
	b) If the robots in $\mathcal{V}$ have unaligned coordinate frames, then $(\mathcal{V},\mathcal{T}, \omega_d)$ is 3-step relatively localizable.
\end{proposition}

We assume that the multi-robot system $(\mathcal{V}, \mathcal{T})$ is localizable, and then the general linear relative localization algorithm is presented in Algorithm \ref{al general}. 

\begin{algorithm}
    \caption{Robot $i$'s linear relative localization algorithm at time instant $t=k$}
    \label{al general}
    \begin{algorithmic}
        \REQUIRE The measured interior angles $\alpha$ and self-displacements $\upsilon$ from time instants $t=k$ to $t=k+2$

		\FOR{Each neighbor $j$ in $\mathcal{N}_{i}$}

			\IF{The neighbor $j$ has not yet formed a tetrahedron with $i$}
				\STATE Find a tetrahedron with $j$ and other two neighbors $m$ and $s$ according to Theorem \ref{th loc cond}.
			\ENDIF

			\STATE Exchange measurements through communication with its neighbors in $\mathcal{N}_{i}$.

			\STATE Use Theorem \ref{theorem8} or Theorem \ref{thm  un lin sol} to compute the relative positions and orientations of the robots.

		\ENDFOR

        \RETURN relative positions and orientations of $i$'s neighbors at time instant $t=k$.
    \end{algorithmic}
\end{algorithm}

\begin{remark}
	One of the main novelties of the linear relative localization proposed in this paper is its linear determination of the relative positions and orientations, distinguishing it from existing nonlinear algebraic-based algorithms such as \cite{zhou2012determining} and \cite{mao2013relative}.
	The proposed linear relative localization requires very low computation cost since it only involves solving a linear equation.
	Moreover, since it does not rely on prior information, the proposed algorithm eliminates the need for initial steps such as frame calibration, thereby facilitating practical deployment in multi-robot systems.
\end{remark}

\section{Relative localization with measurement noise}
\label{sec map}

In this section, we propose a robust relative localization algorithm by combining Algorithm \ref{al general} and the MAP estimator, particularly when measurement noise significantly affects localization accuracy. 
We use four robots as the case to conduct the analysis.

\subsection{Maximum a posterior estimation}
\label{sec map 1}

The state transition models of $p^i_{ji}$ and $r_{ji}$ are given by
\begin{equation}
    \label{eq state trans mod}
    \begin{aligned}
        &p_{ji}^i[k+1] = p_{ji}^i[k] + \Delta p_i^i[k] - 
		\begin{bsmallmatrix}
			(\Delta p_j^j[k])^{\times} & 0_{2 \times 1}, \\
			0_{1 \times 2} & \Delta p_j^j[k](3)
		\end{bsmallmatrix}
		\begin{bsmallmatrix}
			r_{ji}[k] \\
			1
		\end{bsmallmatrix}, \\
        &r_{ji}[k+1] = r_{ji}[k],
    \end{aligned}
\end{equation}
where $p_{ji}^i$, $r_{ji}$ are defined as in \eqref{eq unaligned Q1} and \eqref{eq una times}, respectively.
The transition models of $p^i_{mi}$, $r_{mi}$ and $p^i_{si}$, $r_{si}$ have the similar form to \eqref{eq state trans mod}.
The measurement model of $\alpha_{ijm}$ is
\begin{equation}
	\label{eq mea mod}
    \begin{aligned}
        \alpha_{ijm} (p^i_{mi}, p^i_{ji})
        =& \textrm{sgn}(p^i_{ji}, p^i_{ji} - p^i_{mi}) \\
		& \cdot \arccos
        \Big(\Big.
            \frac{p^i_{ji}}{\left\| p^i_{ji} \right\|} \cdot \frac{p^i_{ji} - p^i_{mi}}{\left\| p^i_{ji} - p^i_{mi} \right\|}
        \Big.\Big),
    \end{aligned}
\end{equation}
where $\textrm{sgn}(p_1, p_2) = 1$ if $\frac{p_{1}^{\textrm{T}}}{\left\| p_{1} \right\|} R_z(\frac{\pi}{2}) \frac{p_{2}}{\left\| p_{2} \right\|} > 0$, otherwise $\textrm{sgn}(p_{1}, p_{2}) = -1$.
The measurement models for other angles are similar to \eqref{eq mea mod}.

For notation simplicity, let $\upsilon$, $\alpha$ denote the stack vectors of self-displacements and interior angles, respectively.
Abbreviate \eqref{eq state trans mod} and \eqref{eq mea mod} as $x[k+1] = f(x[k], \upsilon[k])$ and $\alpha[k] = y(x[k])$, respectively. 
Here, $x = [(p^i)\T, r\T] \in \mathbb{R}^{15}$, where $p^i = [(p_{ji}^i)\T, (p_{mi}^i)\T, (p_{si}^i)\T]\T \in \mathbb{R}^{9}$, and $r\T = [r_{ji}\T, r_{mi}\T, r_{si}\T]\T \in \mathbb{R}^{6}$.
We introduce the notations $\Delta \alpha$, $\Delta \upsilon$ to represent the noise associated with the measurements of angles and self-displacements, respectively. 
Then we define $\tilde{\alpha} = \alpha + \Delta \alpha$, $\tilde{\upsilon} = \upsilon + \Delta \upsilon$ to denote the noisy measurements.
Based on this definition, we make the following assumption.

\begin{assumption}
	\label{ass noise}
	The measurement noise follows Gaussian distribution with zero mean and isotropic covariance, namely, $\Delta \alpha \sim \mathcal{N}(0, P)$, $\Delta \upsilon \sim \mathcal{N}(0, Q)$.
\end{assumption}

When considering noise, by using the Taylor first-order expansion, models $f(\cdot)$ and $y(\cdot)$ can be approximated by 
\begin{equation}
	\label{eq approx}
	\begin{aligned}
		&x[k+1] 
		= f(x[k], \tilde{\upsilon}[k]) \approx f(x[k], \upsilon[k]) 
		+ J_{\upsilon}[k] \Delta \upsilon[k], \\
		&\tilde{\alpha}[k] = y(x[k]) + \Delta \alpha[k],
	\end{aligned}
\end{equation}
where $J_{\upsilon}[k] = \frac{\partial f}{\partial \upsilon} \Big|_{x[k], \tilde{\upsilon}[k]} \in \mathbb{R}^{15 \times 12}$ represents the Jacobian matrix of $f(\cdot)$ with respect to $\upsilon$ at time instant $t=k$.
According to Bayesian inference \cite[Section 4.3]{barfoot2024state}, the MAP estimation is given by
\begin{equation}
	\label{eq map}
	\begin{aligned}
		\hat{x} 
		= \arg\max_{x} ~ \frac{p(x | \tilde{\alpha}, \tilde{\upsilon}) p(x | \tilde{\upsilon})}{p(\tilde{\alpha} | \tilde{\upsilon})}
		= \arg\max_{x} ~ p(\tilde{\alpha} | x) p(x | \tilde{\upsilon}),
	\end{aligned}
\end{equation}
where we drop the denominator since it does not depend on $x$. Additionally, we omit $\tilde{\alpha}$ in $p(x | \tilde{\alpha}, \tilde{\upsilon})$ since it does not affect $x$.
Based on the facts $p(\tilde{\alpha} | x) = \prod_{k=0}^{k_{w_1}} p(\tilde{\alpha}[k] | x[k])$ and $p(x | \tilde{\upsilon}) = \prod_{k=0}^{k_{w_1}-1} p(x[k+1] | x[k], \tilde{\upsilon}[k])$, Assumption \ref{ass noise} and \eqref{eq approx}, taking $\log(\cdot)$ on both sides of \eqref{eq map} yields
\begin{equation}
	\label{eq logmap}
	\begin{aligned}
		\hat{x} =& \arg\min_x ~ 
		\log p(x[0] | \check{x}[0]) 
		+ \frac{1}{2} \sum_{k=0}^{k_{w_1}} \left\| \tilde{\alpha}[k] - y(x[k]) \right\|^2_{Q} \\
		&+ \frac{1}{2} \sum_{k=0}^{k_{w_1}-1} \left\| x[k+1] - f(x[k], \tilde{\upsilon}[k]) \right\|^2_{P'[k]},
	\end{aligned}
\end{equation}
where $\check{x}[0] \in \mathbb{R}^{15}$ denotes the prior estimation on $x[0]$, $P'[k] = J_{\upsilon}[k]\T |_{\upsilon = \tilde{\upsilon}} \cdot P \cdot J_{\upsilon}[k] |_{\upsilon = \tilde{\upsilon}} \in \mathbb{R}^{15 \times 15}$,
and $k_{w_1} \in \mathbb{N}$ denotes the time instant at which MAP estimation is performed.
The optimization problem \eqref{eq logmap} is known as the MAP estimation.
By solving it, one can find the optimal solution $\hat{x}$ with the highest posterior probability density.

One of the main challenges is the non-convexity of \eqref{eq logmap}, which leads to the risk of the solution converging to a local optima. 

Another challenge lies in determining the term $\log p(x[0] | \check{x}[0])$ in \eqref{eq logmap}.
Most existing works such as \cite{huang2015bank}, \cite{vezzani2017memory}, and \cite{cossette2021relative} would replace $\log p(x[0] | \check{x}[0])$ in \eqref{eq logmap} with $\left\| x[0] - \check{x}[0] \right\|^2_{R}$ under the additional assumption $x[0] \sim \mathcal{N}(\check{x}[0], R)$.
However, this assumption lacks theoretical justification and may lead to significant localization errors when the mean $\check{x}[0]$ and covariance $R$ are mismatched, as shown in Figure \ref{fig prior effect}.

Additionally, the MAP estimation is a batch estimation algorithm. 
As the robots continuously move and acquire new measurements over time, the size of the unknown state $x$ grows, making the computation of $x$ impractical for real-time operation.

In Sections \ref{sub sec tls}, \ref{sub sec nde}, \ref{sub sec marginalization}, we propose algorithms to address these challenges, respectively.

\subsection{Initial value estimation for optimizing MAP}
\label{sub sec tls}

In this subsection, we develop a real-time algorithm to estimate the relative positions and orientations of robots based on Algorithm \ref{al general}, whose estimation results can be used as the initial value when solving \eqref{eq logmap}.
For notation simplicity, we drop the time instant term $[k]$ in the following.

When noisy measurements $\tilde{\alpha}$ and $\tilde{\upsilon}$ are obtained, directly applying Algorithm \ref{al general} to estimate the relative positions and orientations of the robots by solving the linear equation $\tilde{A} x = \tilde{b}$ can result in significant localization errors.
Here, the linear equation in Theorem \ref{thm un lin sol} is denoted as $\tilde{A} x=\tilde{b}$ with $\tilde{A} = A + \Delta A$, $\tilde{b} = b + \Delta b$, where $A$, $b$ denote the true coefficient matrices in the linear equation and $\Delta A \in \mathbb{R}^{18 \times 15}$, $\Delta b \in \mathbb{R}^{15}$ represent the noise matrices due to measurement noise.

To reduce the negative effect of measurement noise, firstly we utilize the law that the sum of interior angles of a triangle is equal to $\pi$.
For example, measured angles $\tilde{\alpha}_{\triangle s' i j'} = [\tilde{\alpha}_{s' j' i}, \tilde{\alpha}_{j' s' i}, \tilde{\alpha}_{s' i j'}]\T \in \mathbb{R}^3$ utilized in Algorithm \ref{al general} belong to the triangle $\triangle i j' s'$.
To reduce the noise existing in these angles, a least squares problem with a constraint is formulated, i.e., 
\begin{equation}
	\label{eq lnp}
	\begin{aligned}
		\min_{\alpha_{\triangle s' i j'}} &~ \left\| \left| \alpha_{\triangle s' i j'} \right| - \left| \tilde{\alpha}_{\triangle s' i j'} \right| \right\|^2 \\
		\textrm{s.t.} &~ \textbf{1}\T \left| \alpha_{\triangle s' i j'} \right| = \pi,
	\end{aligned}
\end{equation}
where $\alpha_{\triangle s' i j'} = [\alpha_{s' j' i}, \alpha_{j' s' i}, \alpha_{s' i j'}]\T \in \mathbb{R}^3$ denotes the estimated angles, and $\left| \cdot \right|$ denotes the absolute value.
Problem \eqref{eq lnp} is known as a least norm problem \cite[Section 6.2]{boyd2004convex}, whose analytical solution is 
\begin{equation}
	\label{eq alpha}
	\alpha_{\triangle s' i j'} = 
	\textrm{sgn}(\tilde{\alpha}_{\triangle s' i j'}) 
	\odot \bigg( \left| \tilde{\alpha}_{\triangle s' i j'} \right|
	+ \frac{\pi - \textbf{1}\T \left| \tilde{\alpha}_{\triangle s' i j'} \right|}{3} \bigg),
\end{equation}
where $\textrm{sgn}(\tilde{\alpha}_{\triangle s' i j'})$ denotes the sign vector of $\tilde{\alpha}_{\triangle s' i j'}$ and $\odot$ denotes the Hadamard product.
For the other measured angles that exist in other triangles, a similar way can be utilized to reduce the measurement noise.

Based on the modified angles from \eqref{eq alpha}, new coefficient matrices $\tilde{A}$ and $\tilde{b}$ can be obtained.
To obtain the relative positions and relative orientations, an intuitive idea is to use the least square (LS) method to solve the linear equation $\tilde{A} x = \tilde{b}$, namely, solving the problem
\begin{equation*}
    \begin{aligned}
        \min_{x} ~ \big\| \tilde{A} x - \tilde{b} \big\|_2^2,
    \end{aligned}
\end{equation*}
where $x = [(p_{ji}^i)\T, (p_{mi}^i)\T, (p_{si}^i)\T, r_{ji}\T, r_{mi}\T, r_{si}\T]\T$.
However, it is unsuitable to use LS when both $\tilde{A}$ and $\tilde{b}$ are inaccurate \cite{van199310, markovsky2007overview}.
To achieve an estimation with better accuracy, we formulate the estimation problem as a total least square (TLS) problem, i.e., 
\begin{equation}
	\label{eq tls}
	\begin{aligned}
        \min_{x \in \mathbb{R}^{15}, \Delta C} ~ & \big\| \Delta C \big\|^2_2 \\
        \textrm{s.t.} ~~~~ & (\tilde{C} + \Delta C) \begin{bmatrix}
			x \\ -1
		\end{bmatrix} = 0,
    \end{aligned}
\end{equation}
where $\tilde{C} = [\tilde{A}, \tilde{b}] \in \mathbb{R}^{6d \times 16}$, 
$\Delta C \in \mathbb{R}^{6d \times 16}$ denotes the noise matrix to be estimated, and $d \in \mathbb{R}^+$ denotes the number of time instants utilized, as introduced in Theorem \ref{th loc cond}, whose effects will be introduced later.

The cost function in \eqref{eq tls} reaches its minimum value which equals the minimum eigenvalue of $\tilde{C}\T \tilde{C}$, if and only if $\frac{[x\T, -1]\T}{\left\| [x\T, -1]\T \right\|}$ equals the eigenvector corresponding to the minimum eigenvalue \cite{van199310}.
We apply singular value decomposition to express $\tilde{C}$ as $\tilde{C} = U S V\T$, where $S \in \mathbb{R}^{16 \times 16}$, $U \in \mathbb{R}^{6d \times 16}$ and $V \in \mathbb{R}^{16 \times 16}$.
The analytical formulation of the globally optimal solution of \eqref{eq tls} is given by
\begin{equation}
  \label{eq op2 xa}
  \check{x}_a = - \frac{1}{v_{16, 16}} [v_{1, 16}, \cdots, v_{15, 16}]\T,
\end{equation}
where $v_{i, 16} \in \mathbb{R}$ denotes the $i$th element of $V$'s 16th column.
A detailed derivation of \eqref{eq op2 xa} can be found in \cite{van199310}.

However, \eqref{eq op2 xa} is still not the best estimation due to the following two reasons.
Firstly, \eqref{eq op2 xa} is biased, since the noise in each element of $\tilde{C}$ does not necessarily follow a Gaussian distribution with zero mean, which is a fundamental assumption of TLS.
Secondly, by observing the composition of $x$, one can find that $r_{ji} = [\cos \theta_{j}^i, \sin \theta_{j}^i]\T$ belongs to the Stiefel manifold \cite[Section 7.2]{boumal2023introduction}, defined as $\mathcal{S}^1 = \left\{ x \in \mathbb{R}^{2} | \left\| x \right\| = 1 \right\}$.
Similarly, it can be inferred that $r_{mi} \in \mathcal{S}^1$ and $r_{si} \in \mathcal{S}^1$.
In \eqref{eq tls}, this information is not utilized.
Thus, \eqref{eq tls} can be reformulated as
\begin{equation}
	\label{eq tls mani}
    \begin{aligned}
        \min_{x \in \mathbb{R}^9 \times \mathcal{S}^3, \Delta C} ~ & \big\| \Delta C \big\|^2_{P^{-1}_{\Delta C}} \\
        \textrm{s.t.} ~~~~ & (\tilde{C} + \Delta C) \begin{bmatrix}
			x \\ -1
		\end{bmatrix} = 0,
    \end{aligned}
\end{equation}
where $P_{\Delta C} \in \mathbb{R}^{(6d \times 16) \times (6d \times 16)}$ denotes the covariance matrix of noise matrix $\Delta C$, which satisfies $\textrm{vec}(\Delta C) \sim \mathcal{N}(0, P_{\Delta C})$, and $\left\| \Delta C \right\|_{P^{-1}_{\Delta C}} = \sqrt{\textrm{vec}\T(\Delta C) P^{-1}_{\Delta C} \textrm{vec}(\Delta C)}$.
In \eqref{eq tls mani}, replacing the 2-norm with the matrix norm yields an unbiased estimation result.
Furthermore, \eqref{eq tls} is reformulated as an optimization problem on manifolds, ensuring that $r_{ji}$, $r_{mi}$ and $r_{si}$ always lie on the manifold $\mathcal{S}^1$.
The aim of introducing the manifold is to avoid solutions that may be unreasonable but optimal when solving \eqref{eq tls}.
To remove the constraint in \eqref{eq tls mani} for simplicity, we introduce a new variable $z = [x\T, -1]\T \in \mathbb{R}^9 \times \mathcal{S}^3 \times \mathbb{R}$.
By utilizing $z$, we reformulate \eqref{eq tls mani} as
\begin{equation}
	\label{eq tls hz}
	\begin{aligned}
		\min_{z \in \mathbb{R}^9 \times \mathcal{S}^3 \times \mathbb{R}} ~  \frac{1}{2} \| \tilde{C}z \|_{(P'_{\Delta C})^{-1}}^2, \\
	\end{aligned}
\end{equation}
where $P'_{\Delta C} = (z\T \otimes I) P_{\Delta C} (z\T \otimes I)\T \in \mathbb{R}^{18 \times 18}$.
The proof of this transformation can be found in \cite[Section 19]{pintelon2012system}.
The globally optimal solution of \eqref{eq tls hz} relates to the best estimation of the relative positions and orientations satisfying the MLE criterion with better accuracy.

The optimal solution of solving \eqref{eq tls hz} cannot be guaranteed to be globally optimal because the feasible set of the decision variable is non-convex. 
To mitigate the risk of convergence to local minima, an appropriate initial value is employed when solving \eqref{eq tls hz}, which is given by 
\begin{equation}
	\label{eq proj xa}
	\textrm{Proj}_x (\check{x}_a) = \left[ 
		\check{x}\T_{a, 1:9}, 
		\frac{\check{x}_{a, 10:11}\T}{\small\| \check{x}_{a, 10:11}\T \small\|}, 
		\frac{\check{x}_{a, 12:13}\T}{\small\| \check{x}_{a, 12:13}\T \small\|},
		\frac{\check{x}_{a, 14:15}\T}{\small\| \check{x}_{a, 14:15}\T \small\|}
	\right]\T,
\end{equation} 
where $\check{x}_{a, i:j}$ denotes the elements from the $i$th to the $j$th of $\check{x}_{a}$.
This initial value is chosen since it lies in the neighborhood of the globally optimal solution.
With this initialization, the solution can iteratively converge to the optimal solution \cite{ourtro, thomas2013geodesic}.

Many solvers have been developed to solve the optimization problem on manifolds, including the trust region method \cite{absil2007trust}, the splitting method \cite{lai2014splitting}, the Quasi-Newton method \cite{hu2019structured}, etc.
In this paper, we employ the Trust Region method, as it offers several advantages such as the quadratic convergence.
The trust region method constructs a local quadratic function that approximates the objective function of the optimization problem within a bounded domain, termed the trust region, where the discrepancy between the original and the quadratic functions remains small enough.
Within this trust region, a subproblem that involves minimizing the quadratic function is addressed using Newton’s method.
By iteratively adjusting the size of the trust region and solving the subproblem, the method ensures that the decision variable converges to the optimal solution.

We summarize this initial value estimation algorithm as Algorithm \ref{al wtls}.
The optimal solution $\check{x}$ of \eqref{eq tls hz} can then be used as the initial value when solving the MAP problem \eqref{eq logmap}.

\begin{algorithm}[tbhp]
    \caption{Robot $i$'s initial value estimation algorithm at time instant $t=k$}
    \label{al wtls}
    \begin{algorithmic}
		\REQUIRE The results of Algorithm \ref{al general}.
		\STATE Get angles $\alpha$ by \eqref{eq lnp}.
		\STATE Get coefficient matrices $\tilde{A}$, $\tilde{b}$ based on $\alpha$.
		\STATE Compute $\textrm{Proj}_x (\check{x}_a)$ following \eqref{eq proj xa}.
		\STATE Take $\textrm{Proj}_x (\check{x}_a)$ as the initial value, and seek the optimal solution $\check{x}[k]$ of \eqref{eq tls hz} by using the trust region method \cite{absil2007trust}.
	
        \RETURN The estimation $\check{x}[k]$.
    \end{algorithmic}
\end{algorithm}

Algorithm \ref{al wtls} has the potential to obtain the estimation results with higher accuracy.
According to \cite[Section 2.3]{markovsky2007overview}, as $6d \to \infty$, the estimation $\check{x}$ obtained by Algorithm \ref{al wtls} converges to the true value, indicating that increasing the number of independent linear equations improves the accuracy of $\check{x}$.
In Algorithm \ref{al wtls}, measurements from only three consecutive time instants ($d=3$) are used to form the linear equations, which is the minimum value required to satisfy the sufficient localizability condition in Theorem \ref{th loc cond}.
By increasing $d$ and incorporating more measurements to form the linear equations, the accuracy of $\check{x}$ improves, and thereby the accuracy of the final estimation result obtained by Algorithm \ref{al total} (which will be introduced later) also improves.

\begin{remark}
    \label{rem wtls}
    Algorithm \ref{al wtls} presents two fundamental differences compared to the widely used iteratively reweighted least squares (IRLS) algorithm \cite{holland1977robust,wolke1988iteratively}.
    First, Algorithm \ref{al wtls} considers a more realistic model $\tilde{A}x = \tilde{b}$, which accounts for noise in both $\tilde{A}$ and $\tilde{b}$, whereas IRLS assumes that noise only exists in $\tilde{b}$. 
    Second, IRLS algorithm addresses measurement noise/outliers through iterative reweighting of residuals.
    Algorithm \ref{al wtls} formulates the estimation problem as a WTLS optimization problem on manifolds, and ensures physically meaningful solutions even with noisy measurements, which offers a guarantee that IRLS cannot provide.
\end{remark}

\subsection{Density estimation}
\label{sub sec nde}

In this subsection, we propose an algorithm to estimate $p(x[0] | \check{x}[0])$ as introduced in \eqref{eq logmap}, where $\check{x}[0]$ denotes the solution obtained by using Algorithm \ref{al wtls}.
For simplicity in notation, we denote $x$ and $\check{x}_o$ to represent $x[0]$ and $\check{x}[0]$, respectively.
Let $q(x | \check{x}_o)$ represent the estimated conditional density of $p(x | \check{x}_o)$.

Let's consider the scenario, where once robots start to operate, the true relative positions and orientations $x$ at time instant $t=0$ are determined.
Then robots move, measure and use Algorithm \ref{al wtls} to obtain the estimation $\check{x}_o$ of $x$, implying that a relationship exists between $x$ and $\check{x}_o$.
Therefore, one can regard $\check{x}_o$ as an observation following a distribution determined by the parameter $x$. 
The objective is to estimate $p(x | \check{x}_o)$ given $\check{x}_o$, which can be considered a conditional density estimation problem \cite{west1993approximating, papamakarios2021normalizing}.
To address the conditional density estimation problem, we utilize the Neural Density Estimator (NDE) \cite{alsing2019fast, papamakarios2021normalizing}, which is a general tool for approximating the probability densities using flexible parameterized networks.
The intuition behind using the NDE mainly stems from its tractable computation and ease of implementation, especially compared to analytical-based algorithms such as \cite{west1993approximating, roeder1997practical, escobar1995bayesian, chen2005bayesian}, which often involve complicated formulations and high computational costs in high-dimensional density estimation problems.

\begin{figure*}
	\centering
	\includegraphics[width=1\textwidth]{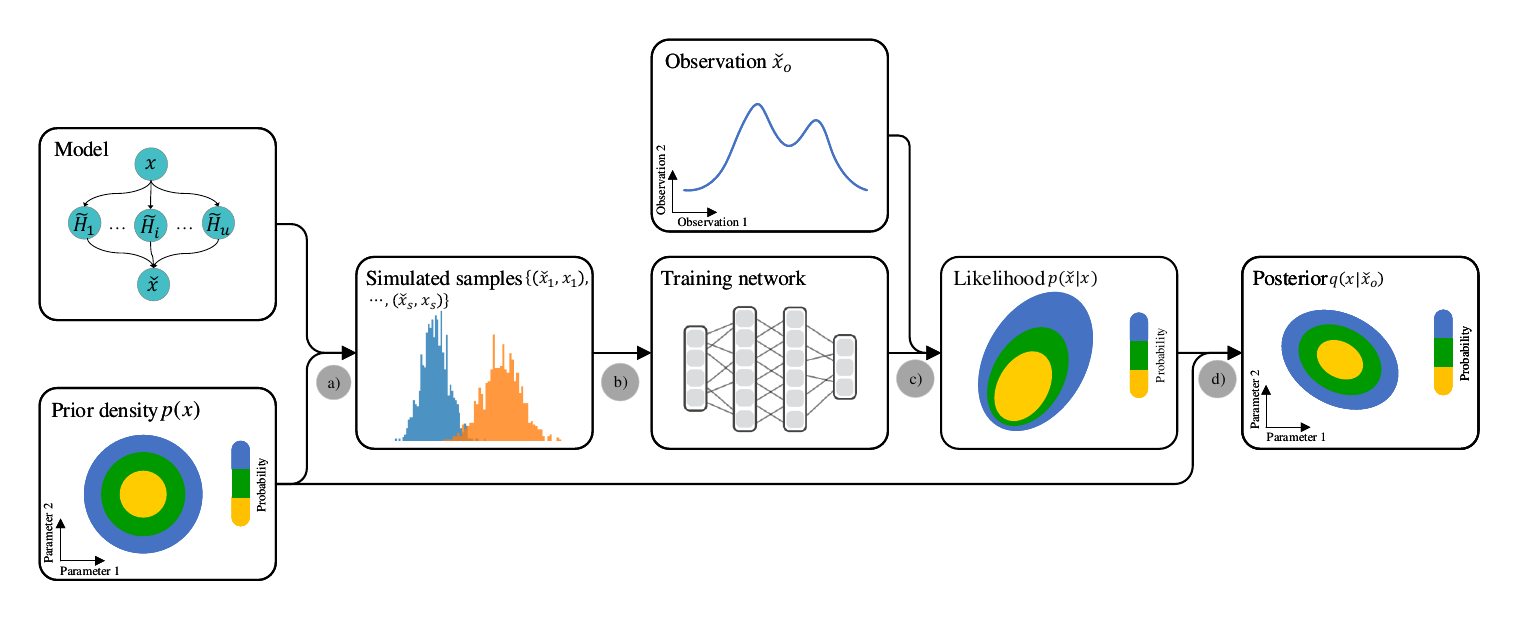}
	\caption{
		The NDE takes three inputs: a model that maps $x$ to $\check{x}$, prior density $p(x)$ and observation data $\check{x}_o$.
		NDE executes by a) simulating samples $(\check{x}_i, x_i)$ from prior density $p(x)$ and the model, 
		and b) training a neural network to fit the likelihood $p(\check{x} | x)$ based on the set of samples $\{ (x_1, \check{x}_1), \cdots, (x_s, \check{x}_s)\}$, where $s$ denotes the number of samples,
		and c) evaluating the likelihood $p(\check{x} | x)$ at the observed data $\check{x}_o$,
		and d) obtaining the estimated conditional posterior density via Bayes theorem $q(x | \check{x}_o) \varpropto p(\check{x}_o | x) p(x)$.
		}
	\label{fig fl cht}
\end{figure*}

To apply the NDE, as shown in Figure \ref{fig fl cht}, we first define a suitable prior density $p(x)$.
Secondly, we establish a model from $x$ to $\check{x}$, which enables us to simulate a set of samples $\left\{ (\check{x}_1, x_1), \cdots, (\check{x}_s, x_s)\right\}$.
Thirdly, we design the network structure and train it to fit the samples.
With these steps completed, the approximation for $p(x | \check{x}_o)$ can be obtained following Bayes theorem $q(x | \check{x}_o) \varpropto p(\check{x}_o | x) p(x)$.

\subsubsection{Prior density $p(x)$:}
For the prior density $p(x)$, since comprehensive information is lacking, a common strategy is to employ a uniform distribution to represent it \cite{van2021bayesian}.
However, it's important to note that the range of the uniform distribution cannot be infinite.
To determine an appropriate range, we establish a confidence bound for $x$ based on the Cramér-Rao lower bound (CRLB) of $\check{x}_o$, and utilize the confidence bound as the range of the uniform distribution.
According to \cite[Section 6]{boumal2014cramer}, the CRLB of the optimal solution on manifolds is given by
\begin{equation}
	\textrm{CRLB} \geq F^{-1}_z + \textrm{curvature terms},
\end{equation}
where $F^{-1}_z \in \mathbb{R}^{16 \times 16}$ denotes the inverse of the Fisher information matrix of \eqref{eq tls hz}, and the curvature terms will become negligible at large signal-to-noise ratios (SNR).
That enables us to approximate CRLB by $F^{-1}_z$.

\begin{lemma}
	\label{lemma fisher}
	\cite[Section 19.3]{pintelon2012system}
	Let $F_z$ be the Fisher information matrix of the parameter $z$ in \eqref{eq tls hz}.
	Then the expression of $F_z$ is given by
	\begin{equation}
		F_z = \left( 
			\frac{\partial (\tilde{C} z)}{\partial z} \right)\T (P'_{\Delta C})^{-1} \left( \frac{\partial (\tilde{C} z)}{\partial z} 
			\right).
	\end{equation}
\end{lemma}

Lemma \ref{lemma fisher} allows us to subsequently determine the CRLB of the estimation result from \eqref{eq tls hz}. 
Let $\check{\textrm{CRLB}}$ denote the approximation of CRLB based on $\check{x}_{o}$. 
Following \cite{ly2017tutorial}, one can compute the $95\%$ confidence interval for each element of $\check{x}_o$ by using
\begin{equation}
	\label{eq cl}
	\textrm{CI}_{i} = \left[ \check{x}_{o,i} - 1.96 \sqrt{\check{\textrm{CRLB}}_{i,i}}, ~ \check{x}_{o,i} + 1.96 \sqrt{\check{\textrm{CRLB}}_{i,i}} \right],
\end{equation}
where $\check{\textrm{CRLB}}_{i,i}$, $i \in \{ 1, \cdots, 15\}$, denotes the $i$th element of the $\check{\textrm{CRLB}}$ matrix's diagonal.
Using the confidence interval $\textrm{CI}_{i}$ enables us to define a finite interval for the uniform distribution $p(x)$.

\subsubsection{Likelihood density $p(\check{x} | x)$:}
To establish a model from $x$ to $\check{x}$, we start with the conditional density $p(\check{x} | x)$, which can be expressed as
\begin{equation}
	\label{eq p checkx x}
	\begin{aligned}
		p(\check{x} | x)
		=& \int_{\tilde{C}} p(\check{x}, \tilde{C} | x) d \tilde{C}
		= \int_{\tilde{C}} p(\check{x} | \tilde{C}, x) p (\tilde{C} | x) d \tilde{C} \\
		=& \int_{\tilde{C}} p(\check{x} | \tilde{C}) p (\tilde{C} | x) d \tilde{C}, \\
	\end{aligned}
\end{equation}
where we omit $x$ in $p(\check{x} | \tilde{C}, x)$ since it does not affect the result $\check{x}$ when $\tilde{C}$ is determined according to \eqref{eq tls hz}.
When $x$ is fixed, let $e = f_e(\tilde{C}) = \tilde{C} z$ where $e \sim \mathcal{N}(0, P'_{\Delta C})$ and $z = [x\T, -1]\T$.
This implies that the analytical form of $e$'s conditional density function $p_e(e | z)$ is known.
Notably, $\tilde{C}$ can be expressed as the inverse function of $e$, i.e., $\tilde{C} = f_e^{-1}(e)$. 
Applying the transformation of random variables \cite[Section 2]{papamakarios2021normalizing} one can deduce that 
\begin{equation}
	\label{eq pHx}
	p (\tilde{C} | x) = p(\tilde{C} | z) =  p_e( f_e(\tilde{C}) | z) \left| \det \frac{\partial f_e}{\partial \tilde{C}} \right|.
\end{equation}
Regarding $p(\check{x} | \tilde{C})$ in \eqref{eq p checkx x}, while an analytical form is not derivable, one can determine $\check{x}$ by solving \eqref{eq tls hz} given a specific $\tilde{C}$.
Let $\check{x} = f_{\check{x}}(\tilde{C})$ represent the relationship from $\tilde{C}$ to $\check{x}$, 
based on which and the discretization of \eqref{eq p checkx x}, one can establish a model that maps $x$ to $\check{x}$, i.e.,
\begin{equation}
	\label{eq checkx x}
	\check{x} = \sum_{i=1}^u w_i f_{\check{x}}(\tilde{C}_i),
\end{equation}
where $u \in \mathbb{R}$ denotes the sample number of $\tilde{C}_i$ and $w_i = \frac{p (\tilde{C}_i | x)}{\sum_{i=1}^u p (\tilde{C}_i | x)}$ denotes the weight assigned to the sample $\tilde{C}_i$.

\subsubsection{Network structure and training:}
\begin{figure}[tbhp]
	\color{blue}
	\centering\includegraphics[width=0.35\textwidth]{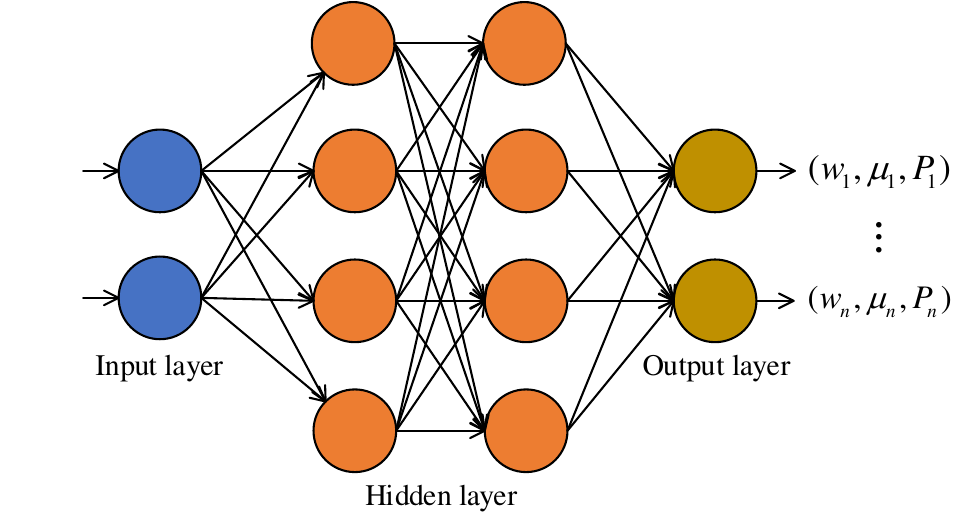}
	\caption{Network structure of the NDE}
	\label{fig structure nde}
\end{figure}

NDE takes a feedforward neural network comprising an input layer, multiple hidden layers and an output layer, as shown in Figure \ref{fig structure nde}.
The input layer consists of neurons that receive the input data. 
Hidden layers are responsible for learning the complex relationship between the data.
The output layer provides the final output of the network.
The density $q(\check{x}|x)$, which approximates $p(\check{x}|x)$, is formulated as a weighted sum of $n \in \mathbb{N}$ Gaussian distribution density functions, i.e.,
\begin{equation*}
	q(\check{x}|x) = \sum_{i=1}^n w_i(x) p(\check{x} | \mu_i(x), P_i(x)),
\end{equation*}
where $p(\check{x} | \mu_i(x), P_i(x))$ denotes the $i$th Gaussian distribution,
$\mu_i(\cdot)$, $P_i(\cdot)$ and $w_i(\cdot)$ denote the mean, covariance and the weight, respectively.
The weight, mean and covariance are all functions of $x$ and are produced as outputs of the neural network.
According to \cite{eldan2016power}, a deeper but small neural network (more hidden layers and less neurons of each hidden layer) is more accurate in approximating the target function.
Combining with other related references \cite{papamakarios2017masked, papamakarios2019neural} and our experience, we set the number of hidden layers to two and the number of neurons of each hidden layer is $30$.

\begin{figure}[tbhp]
\color{blue}
\centering\includegraphics[width=0.48\textwidth]{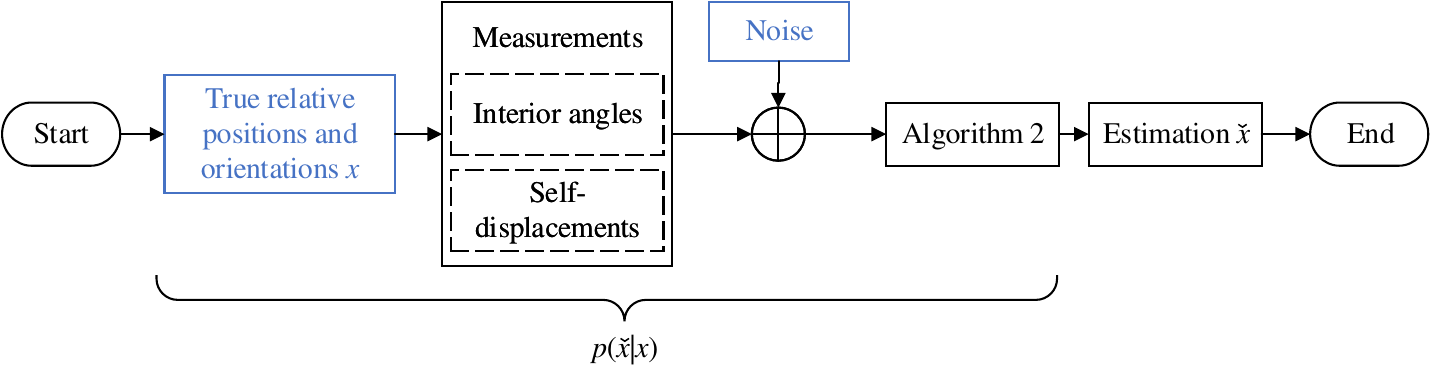}
\caption{Relationship between $x$ and $\check{x}$}
\label{fig x_x_o}
\end{figure}

The training data set $\left\{ (\check{x}_1, x_1), \cdots, (\check{x}_s, x_s)\right\}$ with sufficient size $s \in \mathbb{N}$ for NDE is generated by simulation based on operational data of the multi-robot system.
The density $p(\check{x} | x)$ can be seen as a block-box function to describe the probabilistic relationship between $x$ and $\check{x}$, as shown in Figure \ref{fig x_x_o}.
To achieve an accurate approximation for $p(\check{x} | x)$,  training data is generated after the current operational condition (i.e., $x$, which may vary across different environments) is established.
Although density $p(\check{x} | x)$ cannot be expressed analytically, one can simulate $\check{x}_i$ given $x_i$ by running the process shown in Figure \ref{fig x_x_o}, based on Algorithm \ref{al wtls} and noise characteristics.

The loss function is based on the Kullback-Leibler (KL) divergence \cite[Section 2.3]{papamakarios2021normalizing}, which quantifies the discrepancy between two probability densities.
Specifically, the KL divergence between the approximation $q(\check{x}|x)$ and the target density $p(\check{x}|x)$ is given by
  \begin{equation}
    \label{eq kl}
    d_{\textrm{KL}}(p || q) = \int p(\check{x}|x) \log \left( \frac{q(\check{x}|x)}{p(\check{x}|x)} \right) d \check{x}.
  \end{equation}
Since the target density $p(\check{x}|x)$ is unavailable, a Monte Carlo estimation method based on the sample set $\{ (\check{x}_1, x_1), \cdots, (\check{x}_s, x_s)\}$ can be utilized for approximating \eqref{eq kl}.

\begin{table}[tbhp]
    \centering
    \caption{Configurations of NDE}
    \label{tab nde}
    \begin{tabular}{m{0.24\textwidth}<{\centering}m{0.195\textwidth}<{\centering}}
      \toprule
      Name & Configuration \\
      \midrule
      Training data & Generated by simulation \\
      Network structure & Feedforward neural network \\
      Number of hidden layers & 2 \\
      Number of neurons in each hidden layer & 30 \\
      Active function &  tanh \\
      Prior assumption & The noise characteristics is known \\
      \bottomrule
    \end{tabular}
\end{table}

Table \ref{tab nde} summarizes the configuration of the NDE, while Algorithm \ref{al nde} outlines the complete procedure of the proposed algorithm.
\begin{algorithm}[tbhp]
    \caption{Robot $i$'s density estimation algorithm}
    \label{al nde}
    \begin{algorithmic}
		\REQUIRE The results of Algorithm \ref{al wtls}.
		\STATE Get initial estimation $\check{x}[0]$ (abbreviated as $\check{x}_o$) and coefficient matrix $\tilde{C}$ at time instant $t=0$ from Algorithm \ref{al wtls}.
		\STATE Compute the confidence interval Cl of $\check{x}_o$ by following \eqref{eq cl} based on $\tilde{C}$ and $\check{x}_o$, taken as the bound of the uniform distribution $p(x)$.
		\STATE Compute the conditional density $p (\tilde{C} | z)$ according to \eqref{eq pHx}.

        \FOR{$i = 1, \cdots, s$, where $s$ denotes the number of samples}

			\STATE Sample $x_i$ according to $p(x)$.

			\STATE Sample $\tilde{C}_1, \cdots, \tilde{C}_u$, where $u$ denotes the sample number of $\tilde{C}$ according to $p(\tilde{C} | z)$.

			\STATE Compute $\check{x}_i$ according to \eqref{eq checkx x}.
	 
        \ENDFOR

		\STATE Based on the sample set $\{ (x_1, \check{x}_1), \cdots, (x_s, \check{x}_s)\}$, train the NDE.

        \RETURN The density estimation $q(x | \check{x}_o)$.
    \end{algorithmic}
\end{algorithm}

\begin{remark}
	In this paper we employ the NDE to learn $p(x | \check{x})$ required by the MAP estimator.
	The key to using NDE is the linear relative localization algorithm's independence of prior information and the low cost computation.
	These advantages allow us to efficiently obtain the necessary inputs for NDE with minimal computation.
	A significant benefit of NDE is that it avoids intractable computation often associated with existing analytical-based algorithms.
	Moreover, rather than directly estimating $p(x | \check{x})$, we employ NDE to estimate the likelihood $p(\check{x} | x)$, and subsequently utilize the Bayes theorem to derive $p(x | \check{x})$.
	That's because estimating $p(x | \check{x})$ needs to determine the range of $\check{x}$ and the function from $\check{x}$ to $x$, which is intractable.
	Instead, estimating $p(\check{x} | x)$ requires the range of $x$ and the function from $x$ to $\check{x}$, which is easy to obtain.
\end{remark}

\subsection{MAP optimization on manifolds}
\label{sub sec map mani}

After getting $q(x[0] | \check{x}[0])$, we find the optimal solution of the MAP problem \eqref{eq logmap}.

As discussed in Section \ref{sub sec tls}, the state $x$ to be estimated in Algorithm \ref{al wtls} belongs to a product of manifolds, and therefore, \eqref{eq logmap} can also be reformulated as an optimization problem on manifolds.
Furthermore, by partitioning $p(x[k+1] | x[k], \tilde{\upsilon}[k])$, one can obtain
\begin{equation*}
\begin{aligned}
	& p(x[k+1] | x[k], \tilde{\upsilon}[k]) \\
	=& p(p^i[k+1], r[k+1] | p^i[k], r[k], \tilde{\upsilon}[k]) \\
	=& p(p^i[k+1] | p^i[k], r[k], \tilde{\upsilon}[k])
	p(r[k+1] | p^i[k], r[k], \tilde{\upsilon}[k]) \\
	=& p(p^i[k+1] | p^i[k], r[k], \tilde{\upsilon}[k]) 
	p(r[k+1] | r[k]),
\end{aligned}
\end{equation*}
where the second equation holds since $p^i[k+1]$ and $r[k+1]$ are independent variables according to the definitions, and the third equation holds since $p^i[k]$ and $\tilde{\upsilon}[k]$ have no effects on $r[k+1]$.
According to the transition model \eqref{eq state trans mod}, $p(r[k+1] | r[k]) = 1$ if $r[k+1] = r[k]$, and $p(r[k+1] | r[k]) = 0$ otherwise.
Thus, one can eliminate the term $p(r[k+1] | r[k])$ during optimization when $r[k+1] = r[k]$.
Consequently, we reformulate \eqref{eq logmap} as
\begin{equation}
	\label{eq logmap cons}
	\begin{aligned}
		&\min_{x[k] \in \mathbb{R}^9 \times \mathcal{M}^3} ~ 
		 \log q(x[0] | \check{x}[0]) 
		+ \frac{1}{2} \sum_{k=0}^{k_{w_1}} \left\| \tilde{\alpha}[k] - y(x[k]) \right\|^2_{Q} \\
		&+ \frac{1}{2} \sum_{k=0}^{k_{w_1}-1} \left\| p^i[k+1] - f_{p^i}(p^i[k], r, \tilde{\upsilon}[k]) \right\|^2_{P_{p^i}'[k]},
	\end{aligned}
\end{equation}
where $f_{p^i}(\cdot)$ denotes the transition model of $p^i$, and $P_{p^i}'$ denotes the submatrix of $P'$ corresponding to $p^i$.
Obtaining the analytical expression of $\frac{\partial \log q(x[0] | \check{x}[0]) }{\partial x}$ is infeasible due to the non-parameterized nature of the estimated $q(x[0] | \check{x}[0])$.
To address this, we leverage the automatic differentiation (AD) method \cite{baydin2018automatic} to approximate it.
Similarly, we utilize the trust region method \cite{absil2007trust} to solve \eqref{eq logmap cons}.

\begin{remark}
	\label{rem map scale}
	When the scale of the multi-robot system is large, i.e., more than one tetrahedron exists, Algorithms \ref{al general} and \ref{al wtls} are still executed in each tetrahedron.
    When it comes to the MAP optimization problem formulation, the relative positions and orientations of all neighbor robots can be integrated into a single unified problem.
    This is because restricting the MAP problem formulation within a tetrahedron would confine the estimation to the relative positions and orientations of only four robots.
    This limitation would require constructing and solving multiple MAP problems when multiple tetrahedra are present.
    Instead, once all neighbor robots' relative positions and orientations have been estimated initially using Algorithm \ref{al wtls}, formulating the relative localization problem of all neighbor robots in different tetrahedra as a single MAP problem is preferred.
\end{remark}

\subsection{Marginalization}
\label{sub sec marginalization}

\begin{figure}[hbtp]
	\centering\includegraphics[width=0.35\textwidth]{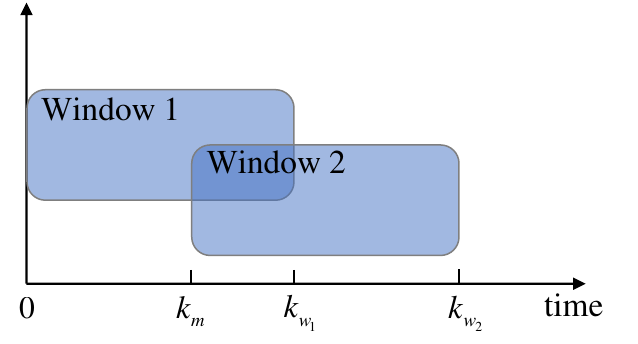}
	\caption{The sliding time window}
	\label{fig swf}
\end{figure}

We consider the scenario shown in Figure \ref{fig swf}. 
When the robots move during the time interval $[0, k_{w_1}]$, the MAP estimator introduced in Section \ref{sub sec map mani} is carried out at time instant $t = k_{w_1}$, resulting in the estimation of $x[0:k_{w_1}] = [x\T[0], \cdots, x\T[k_{w_1}]]\T$.
Then the robots continue to move and collect measurements until the time instant $t=k_{w_2}$.
The old states $x[0: k_{m}-1]$ and corresponding measurements are marginalized out. 
Only states $x[k_m:k_{w_1}]$ and their associated measurements remain active in the new window.
Now combining the remaining states $x[k_m:k_{w_1}]$ and new measurements obtained during $[k_{w_1}, k_{w_2}]$, we would like to compute the estimation for states $x[k_m : k_{w_2}]$ by using the MAP estimator.

To compute the MAP estimation at time instant $k_{w_2}$, one has to find the optimal solution to minimize the cost function
\begin{equation}
	\label{eq swf}
	\begin{aligned}
		&\min_{x[0:k_{w_2}]} f_1(x[0:k_{w_2}]) = \min_{x[k_{w_1}:k_{w_2}]} f_2(x[k_{w_1}:k_{w_2}]) \\
		&+ \min_{x[0:k_{m}-1], x[k_m:k_{w_1}]} f_3(x[0:k_{m}-1], x[k_m:k_{w_1}]),
	\end{aligned}
\end{equation}
where $f_1(\cdot)$ denotes the cost function of \eqref{eq logmap cons} for notation simplicity.
As introduced in Section \ref{sub sec map mani}, we employ the trust region method to solve the MAP optimization on manifolds. 
Generally, this method begins by forming a quadratic model of the cost function using the second-order Taylor approximation around the current iterate state. 
It then finds the optimal solution that minimizes the quadratic model in Euclidean space, and finally, maps the solution back onto the manifold.
In the second step, namely finding the optimal solution of the quadratic model, the old states $x[0:k_{m}-1]$ can be marginalized.
Our aim here is to approximate $f_3(x[0:k_{m}-1], x[k_m:k_{w_1}])$ in \eqref{eq swf}, where $x[0:k_{m}-1]$ and associated measurements no longer exist.
By performing a second-order Taylor expansion of $f_3(x[0:k_{m}-1], x[k_m:k_{w_1}])$ at $(\hat{x}[0:k_{m}-1], \hat{x}[k_m:k_{w_1}])$, one has
\begin{equation}
	\label{eq taylor}
	\begin{aligned}
	& f_3(x[0:k_{m}-1], x[k_m:k_{w_1}]) \\
	=& f_3(\hat{x}[0:k_{m-1}], \hat{x}[k_m:k_{w_1}])
	+ J\T \begin{bmatrix} \hat{x}[0:k_{m}-1] \\ \hat{x}[k_m:k_{w_1}] \end{bmatrix} \\
	&+ \begin{bmatrix} \hat{x}[0:k_{m}-1] \\ \hat{x}[k_m:k_{w_1}] \end{bmatrix}\T
	H \begin{bmatrix} \hat{x}[0:k_{m}-1] \\ \hat{x}[k_m:k_{w_1}] \end{bmatrix},
	\end{aligned}
\end{equation}
where $J$ and $H$ denote the Jacobian and Hessian matrices of $f_3(x[0:k_{m}-1], x[k_m:k_{w_1}])$ at the point $(\hat{x}[0:k_{m}-1], \hat{x}[k_m:k_{w_1}])$, respectively, $(\hat{x}[0:k_{m}-1], \hat{x}[k_m:k_{w_1}])$ denote the MAP estimation obtained at time instant $t=k_{w_1}$.

By utilizing Newton’s method \cite{boyd2004convex, absil2007trust}, which is widely used for solving such quadratic optimization problems, the optimal solution of \eqref{eq taylor} can be obtained.
The state update law of Newton’s method can be performed as
\begin{equation*}
	\begin{bmatrix} x_{new}[0:k_{m}-1] \\ x_{new}[k_m:k_{w_1}] \end{bmatrix}
	= \begin{bmatrix} \hat{x}[0:k_{m}-1] \\ \hat{x}[k_m:k_{w_1}] \end{bmatrix} + \begin{bmatrix} \Delta x[0:k_{m}-1] \\ \Delta x[k_m:k_{w_1}] \end{bmatrix},
\end{equation*}
where
\begin{equation*}
	\begin{bmatrix} \Delta x[0:k_{m}-1] \\ \Delta x[k_m:k_{w_1}] \end{bmatrix} = H^{-1} J.
\end{equation*}
Since our focus is on the changes in $x[k_m:k_{w_1}]$ without involving $x[0:k_{m}-1]$, one can partition $J = [J_m, J_r]$, $H = \begin{bmatrix} H_{mm} & H_{mr} \\ H_{rm} & H_{rr} \end{bmatrix}$, where the dimensions of the sub-matrices are corresponding to $x[0:k_{m}-1]$ and $x[k_m:k_{w_1}]$.
By employing the Schur complement \cite{dong2011motion, huang2011observability}, one can express the state update law as
\begin{equation}
	\label{eq hrjr}
	\Delta x[k_m:k_{w_1}] = H_s^{-1} J_s,
\end{equation}
where 
\begin{equation}
\begin{aligned}
	\label{eq js hs}
	& H_s = H_{rr} - H_{rm} H^{-1}_{mm} H_{mr}, \\
	& J_s = J_{r} - H_{rm} H^{-1}_{mm} J_{m}.
\end{aligned}	
\end{equation}
Focusing on \eqref{eq hrjr}, it is equivalent to using Newton's method to minimize the following cost function
\begin{equation}
	\begin{aligned}
		f_4(x[k_m:k_{w_1}]) =& \textrm{cons} + J_s\T (x[k_m:k_{w_1}] - \hat{x}[k_m:k_{w_1}]) \\
		&+ \frac{1}{2} \left\| x[k_m:k_{w_1}] - \hat{x}[k_m:k_{w_1}] \right\|^2_{H_s},
	\end{aligned}
\end{equation}
where $\textrm{cons}$ denotes a constant, which indicates that $f_3(x[0:k_{m-1}], x[k_m:k_{w_1}])$ in \eqref{eq swf} can be replaced by $f_4(x[k_m:k_{w_1}])$.
Then the new optimization problem at time instant $t = k_{w_2}$ can be reformulated as 
\begin{equation}
	\label{eq new swf}
	\begin{aligned}
		\min_{x[k_m:k_{w_2}]} f_5(x[k_m:k_{w_2}]) 
		=& \min_{x[k_m:k_{w_2}]} f_2(x[k_m:k_{w_2}]) \\
		&+ \min_{x[k_m:k_{w_1}]} f_4(x[k_m:k_{w_1}]),
	\end{aligned}
\end{equation}
and the specific formulation can be written as
\begin{equation}
	\label{eq logmap cons swf}
	\begin{aligned}
		&\min_{x[k] \in \mathbb{R}^9 \times \mathcal{M}^3} ~ 
		 J_s\T (x[k_m:k_{w_1}] - \hat{x}[k_m:k_{w_1}]) \\
		&+ \frac{1}{2} \left\| x[k_m:k_{w_1}] - \hat{x}[k_m:k_{w_1}] \right\|^2_{H_s} \\
		&+ \frac{1}{2} \sum_{k=k_m}^{k_{w_2}} \left\| \tilde{\alpha}[k] - y(x[k]) \right\|^2_{Q} \\
		&+ \frac{1}{2} \sum_{k=k_m}^{k_{w_2}-1} \left\| p^i[k+1] - f_{p^i}(p^i[k], r, \tilde{\upsilon}[k]) \right\|^2_{P_{p^i}'[k]},
	\end{aligned}
\end{equation}
where we drop the term cons since it has no effect on the solution.
Compared to $f_1(x[0:k_{w_2}])$ in \eqref{eq swf}, $f_5(x[k_m:k_{w_2}])$ no longer contains the marginalized state $x[0:k_{m}-1]$, thus maintaining the size of states to be optimized constant.
Besides, the information contained in marginalized states is captured in the matrix $H_{s}$.

Assuming all time windows have equal lengths, marginalization and MAP estimation are triggered at specific time instants $t = k_{w_i}, i \in \mathbb{N}$.
The proposed advanced MAP algorithm is summarized in the pseudo-code provided in Algorithm \ref{al total}.

\begin{algorithm}
    \caption{Robot $i$'s relative localization algorithm}
    \label{al total}
    \begin{algorithmic}
        \REQUIRE Measurements $\tilde{\alpha}$, $\tilde{\upsilon}$.

        \FOR{$k = 0, \cdots, k_{w_1} + 2$}
	 
			\IF{$k < 2$}
				\STATE Continue to get measurements.
			\ENDIF

			\STATE Get coefficient matrices $\tilde{A}$ and $\tilde{b}$ from Algorithm \ref{al general} based on measurements $\tilde{\alpha}$, $\tilde{\upsilon}$.

			\STATE Get the initial state estimation $\check{x}[k-2]$ and matrix $\tilde{C}$ from Algorithm \ref{al wtls} based on $\tilde{A}$ and $\tilde{b}$.

			\IF{$k = 2$}
				\STATE Get the density estimation $q(x[0] | \check{x}[0])$ from Algorithm \ref{al nde} based on $\check{x}[0]$ and the corresponding $\tilde{C}$.
			\ENDIF

			\IF{$k = k_{w_1} + 2$}
				\STATE Compute the optimal solution $\hat{x}[k]$ of \eqref{eq logmap cons} based on the initial estimation $\check{x}[k]$, and $q(x[0] | \check{x}[0])$, where $k=0, \cdots, k_{w_1}$.
			\ENDIF

        \ENDFOR

		\FOR{$k = k_{w_1} + 2, \cdots, k_{w_2} + 2$}

		\STATE Get coefficient matrices $\tilde{A}$ and $\tilde{b}$ from Algorithm \ref{al general} based on measurements $\tilde{\alpha}$, $\tilde{\upsilon}$.

		\STATE Get the initial state estimation $\check{x}[k-2]$ from Algorithm \ref{al wtls} based on $\tilde{A}$ and $\tilde{b}$.

		\IF{$k = k_{w_2} + 2$}
			\STATE Compute the Jacobian $J_s$ and Hessian $H_s$ according to \eqref{eq js hs}.
			\STATE Compute the optimal solution $\hat{x}[k]$ of \eqref{eq logmap cons swf} based on the initial estimation $\check{x}[k]$ and $J_s$, $H_s$, where $k=k_{m}, \cdots, k_{w_2}$. 
		\ENDIF

		\ENDFOR

        \RETURN The relative localization results $\hat{x}[k]$ of $i$'s neighbors.
    \end{algorithmic}
\end{algorithm}

For varying noise levels, we assume that the noise level varies slowly, meaning it remains approximately constant over a short sequence of consecutive time instants.
Algorithm \ref{al wtls} maintains the MLE criterion, while Algorithm \ref{al total} may no longer yield a MAP estimation.
This discrepancy arises because Algorithm \ref{al wtls}, a real-time estimation algorithm, utilizes fewer measurements over a short time period where the noise level remains constant.
In contrast, Algorithm \ref{al total}, a batch estimation algorithm, relies on measurements over a longer time period where the noise level varies.
Within Algorithm \ref{al total}, the noise level is set constant, which may fail to capture the noise level variability. 
Consequently, the estimation produced by Algorithm \ref{al total} under this setting may not align with the true MAP estimation.
To mitigate this issue, an additional algorithm such as \cite{bosse2016robust, patil2022robust} for approximating the noise level can be incorporated.

\begin{remark}
	\label{rem al4}
	The proposed relative localization framework is flexible and allows specific technological components to be replaced with alternative algorithms.
	Algorithm \ref{al nde} approximates the required prior probability density in \eqref{eq logmap cons}.
    Results from Algorithm \ref{al wtls} serve as the initial value for solving the MAP optimization problem in Algorithm \ref{al total}.
    Algorithm \ref{al general} serves as the foundation for the development of Algorithms \ref{al wtls} and \ref{al nde}. 
    Different algorithms with similar functions to Algorithms \ref{al general}, \ref{al wtls} and \ref{al nde} can be integrated into Algorithm \ref{al total}.
\end{remark}

\subsection{Computational complexity}
\label{sub sec computation}

We analyze the computational complexity of Algorithms \ref{al general}, \ref{al wtls} and \ref{al total}.
Let $x \in \mathbb{R}^n$ denote the relative positions and orientations at a single time instant.
As for Algorithm \ref{al nde}, since it is a neural density estimator, which is not a numerical optimization problem, we do not analyze its computational complexity.

The computational complexity of Algorithm \ref{al general} is $O(n^3)$.
In Algorithm \ref{al general}, the relative localization problem is formulated as a linear equation such as \eqref{basiccalculationr3D} and \eqref{eq th2}.
Solving such a linear equation requires $O(n^3)$ computational complexity, corresponding to the complexity of the matrix inversion operation \cite{krishnamoorthy2013matrix}.

The computational complexity of Algorithm \ref{al wtls} is $O(s_1(n^2 + n^3))$ and remains constant, where $s_1 \in \mathbb{N}$ denotes the total number of iterations required by the trust region method.
Algorithm \ref{al wtls} finds the optimal solution of \eqref{eq tls hz} iteratively by utilizing this method.
Each iteration involves solving a linear equation $\Delta x = H^{-1} J$ with $O(n^3)$ complexity and performing projection and retraction steps on the manifold with $O(n^2)$ complexity \cite[Section 5]{boumal2023introduction}, where $\Delta x \in \mathbb{R}^n$ denotes the update to $x$, and $H \in \mathbb{R}^{n \times n}$, $J \in \mathbb{R}^n$ are the corresponding Hessian and Jacobian matrices, respectively.
Therefore, the total computational complexity is $O(s_1(n^2 + n^3))$.

The complexity of Algorithm \ref{al total} is $O(s_2 ( (k_w n)^3 + (k_w n)^2) + k_w s_1(n^2 + n^3))$ and remains constant, where $k_w \in \mathbb{N}$ denotes the window size in the marginalization mechanism.
In Algorithm \ref{al total}, a marginalization mechanism is employed to ensure that the number of variables $x$ to be determined across different time instants remains constant, specifically limited to $k_w$.
Algorithm \ref{al total} first utilizes Algorithm \ref{al wtls} to compute the initial estimation of $x$, which incurs a computational complexity of $O(k_w s_1(n^2 + n^3))$.
Subsequently, the computational complexity of solving the MAP problem is $O(s_2 ( (k_w n)^3 + (k_w n)^2))$, where $s_2 \in \mathbb{N}$ denotes the total number of iterations for MAP.
Consequently, the total computational complexity of Algorithm \ref{al total} is the sum of these two components.

\section{Extensions}
\label{sec ext}

In this section, to show the universal significance of the proposed framework, we introduce the cases where the raw measurements are azimuth and elevation angles or inter-robot distances, other than the angles required by the framework.
Then we introduce the outlier detection and mitigation, and the robustness to sensor failures.

\subsection{The case of using azimuth and elevation angles}
\label{sub sec angle trans}

\begin{figure}[tbhp]
    \centering
    \subfigure[Definition of azimuth and elevation angles]
    {
        \includegraphics[width=0.21\textwidth]{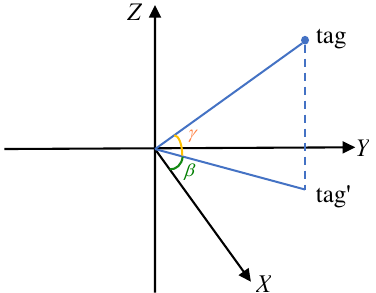}
    }
    \subfigure[Compution of $\alpha_{jmi}$]
    {
        \includegraphics[width=0.21\textwidth]{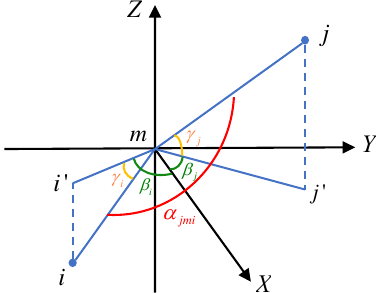}
    }
	\subfigure[Compution of angles in $\triangle m's'i$]
    {
        \includegraphics[width=0.21\textwidth]{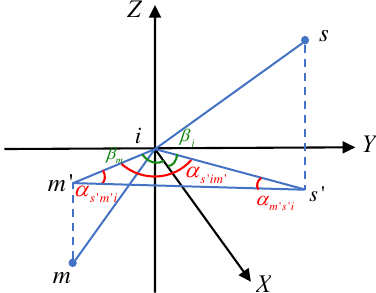}
    }
	\subfigure[Compution of $\alpha_{siz}$]
    {
        \includegraphics[width=0.21\textwidth]{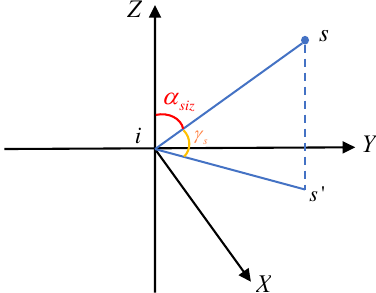}
    }
    \caption{The definition and computation of angles}
	\label{fig aoa sensor}
\end{figure}

Many commercial sensors for angle measurements are now designed to measure the azimuth and elevation angles, which can be transformed into the angles required by the proposed framework.
As illustrated in Figure \ref{fig aoa sensor}(a), the azimuth angle $\beta$ in a sensor's local frame is defined as the angle between the projection of the bearing from the tag to the origin onto the $XOY$ plane and the positive direction of the $X$-axis.
The elevation angle $\gamma$ is defined as the angle between the bearing from the tag to the origin and its projection on the $XOY$ plane.
The angles required by our algorithm are categorized into three types.

Firstly, we use $\alpha_{jmi}$ shown in Figure \ref{fig aoa sensor}(b) as an example of the first type of angles to illustrate the computation process.
The angle $\alpha_{jmi}$ can be computed based on $b^m_{mj} = [\cos\beta_j\cos\gamma_j,  \sin\beta_j\cos\gamma_j, \sin\gamma_j]\T$ and $b^m_{mi} = [\cos\beta_i\cos\gamma_i,  \sin\beta_i\cos\gamma_i,  \sin\gamma_i]\T$ following \eqref{anglecalcu}, where $\beta_j$ and $\gamma_j$ denote the azimuth and elevation angles of robot $j$ in robot $m$'s local frame, respectively, and the definitions of $\beta_i$ and $\gamma_i$ are similar.

Secondly, we take the angles in $\triangle m's'i$ as an example in Figure \ref{fig aoa sensor}(c) to explain how to calculate the second type of angles.
The angle $\alpha_{m'is'}$ can be computed based on the bearings $b^i_{im'} = [\cos\beta_m\cos\gamma_m,  \sin\beta_m\cos\gamma_m,  0]\T$ and $b^i_{is'} = [\cos\beta_s \cos\gamma_s,  \sin\beta_s \cos\gamma_s,  0]\T$ following \eqref{anglecalcu}.
Under the assumption that each robot has a common $Z$-axis direction, indicating that the $XOY$ planes of different robots' local frames are parallel, one can infer that $\alpha_{s'm'i} = \alpha_{s'mi'}$, which can be computed by robot $m$ in the same way introduced above.
The calculation of angle $\alpha_{m's'i}$ follows the same way.

Thirdly, the last type of angles is between the bearing from one robot to another and the $Z$-axis.
For instance, the angle $\alpha_{siz}$ in Figure \ref{fig aoa sensor}(d) can be computed based on $b^i_{is} = [\cos\beta_s\cos\gamma_s,  \sin\beta_s\cos\gamma_s,  \sin\gamma_s]\T$ and $b^i_{iz} = [0,  0,  1]\T$ following \eqref{anglecalcu}.

\subsection{The case of using distance measurements}

\label{sub sec dis trans}

\begin{figure}[bhtp]
	\centering\includegraphics[width=0.25\textwidth]{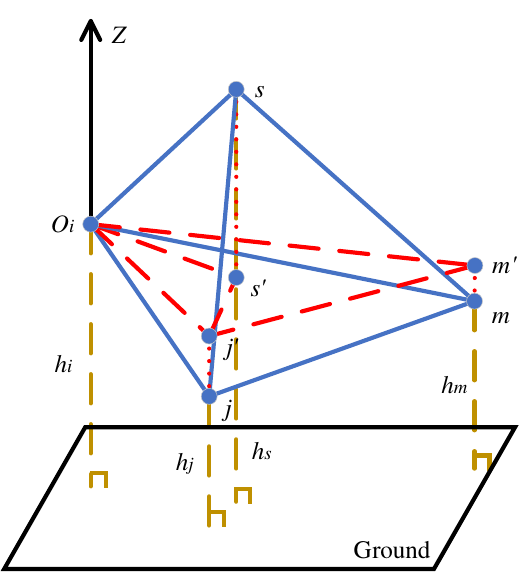}
	\caption{Determination of angles from inter-robot distances and heights with respect to a flat ground}
	\label{fig dis2angle}
\end{figure}

The proposed relative localization framework can also be effectively utilized based on inter-robot distance measurements.
It is achieved by converting the measured distances into interior angles provided each robot has the capability to measure its height relative to a flat ground, which most aerial robots have.
The interior angles $\alpha_{mji},\alpha_{sji},\alpha_{jmi},\alpha_{jsi}$ can be determined from distance measurements using the law of cosines.
The determination of angles in $\triangle ij's'$, $\triangle im's'$ and those related to $Z$-axis direction should combine the distance measurements and the heights of robots.
According to Figure \ref{fig dis2angle}, one directly has $\alpha_{jij'} = \arcsin \left( \frac{\left\| h_i - h_j \right\|}{l_{ij}} \right)$, $\alpha_{j'iZ} = \frac{\pi}{2} + \alpha_{jij'} \textrm{sgn}(h_i - h_j)$, and $l_{ij'} = l_{ij} \cos \alpha_{jij'}$, where $l_{ij}$ denotes the distance between robots $i$ and $j$.
Similarly, one can also obtain $l_{j's'}$ and $l_{is'}$.
Then, in $\triangle ij's'$, one has $\alpha_{j'is'} = \arccos \left( \frac{l_{is'}^2 + l_{ij'}^2 - l_{j's'}^2}{2 l_{is'} l_{ij'}} \right)$.
Using similar equations in other triangles, one can determine all the angles required by the proposed framework.

\subsection{Robustness to sensor failures}
\label{sub sec sensor failure}

If the sensor for measuring a robot's self-displacements fails, the proposed relative localization framework becomes inoperable, since no alternative measurements are available to estimate the self-displacement associated with the failed sensor.
If only one sensor measuring angles or distances fails, the localization framework can continue to operate normally. 
However, if multiple such sensors fail simultaneously, the proposed framework will no longer function.

\begin{figure}[tbhp]
  \centering
  \subfigure[Sensors for angle measurements]
  {
      \includegraphics[width=0.2\textwidth]{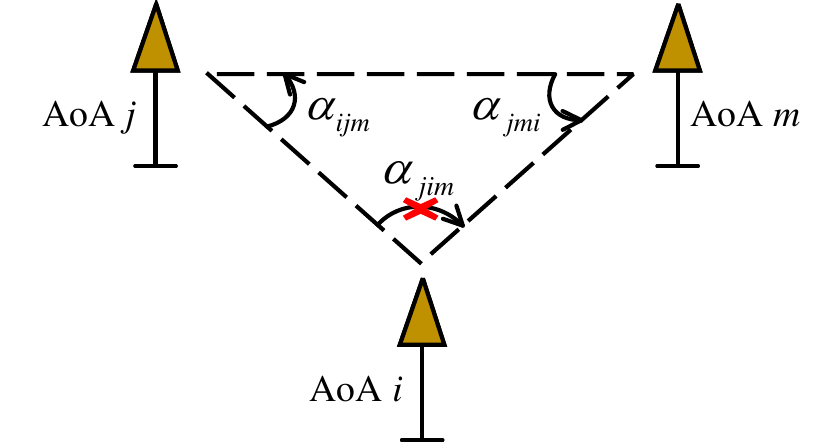}
  }
  \subfigure[Sensors for distance measurements]
  {
      \includegraphics[width=0.215\textwidth]{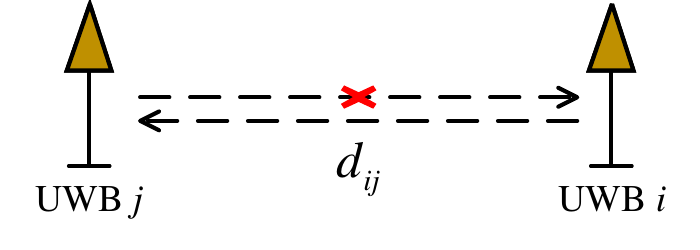}
  }
  \caption{Robustness to sensor failures}
  \label{fig sensor failure2}
\end{figure}

Here we mainly focus on permanent sensor failures. 
In contrast, if a sensor fails intermittently or momentarily, only a few measurements at specific time instants are abnormal, which can be treated as outliers as introduced in Section \ref{sub sec outlier}.
As shown in Figure \ref{fig sensor failure2}(a), if the onboard AoA sensor of robot $i$ fails, the angle $\alpha_{jim}$ cannot be directly measured.
By using the fact that the sum of interior angles in a triangle equals $\pi$, $\alpha_{jim}$ can be estimated by $\alpha_{jim} = \pi - \alpha_{ijm} - \alpha_{jmi}$.
In Figure \ref{fig sensor failure2}(b), if the onboard UWB sensor of robot $i$ fails, it cannot measure the distance but can still emit electromagnetic signals. 
In this case, the distance measurement $d_{ij}$ can still be obtained by the UWB sensor onboard robot $j$.
Then robot $i$ can obtain necessary measurements through inter-robot communication.
However, if the AoA or UWB sensors on more than two robots in a same tetrahedron fail simultaneously, no other sensor can provide information about the required measurements.

Sensor failures can result in persistent outlier measurements. 
Based on the outlier detection method proposed in Section \ref{sub sec outlier}, if a measurement is consistently detected as an outlier over consecutive time instants, the robot can infer that a sensor failure has occurred.

\subsection{Outlier detection and mitigation}
\label{sub sec outlier}

Non-line-of-sight (NLOS), such as occlusions between two UWB sensors, and multipath effects are two common issues when using UWB or AoA sensors.
According to statistical analysis in \cite{alavi2006using, barral2019nlos}, the two issues can introduce outliers into the measurements.
Thus, we model NLOS and multipath negative effects as measurement outliers and propose a detection and mitigation method to address them.

We first present the definition of an outlier.
If the measurement noise follows a Gaussian distribution $\mathcal{N}(0, \sigma)$, $99.7\%$ of the noise lies in the interval $[-3\sigma, 3\sigma]$.
Any measurement whose noise exceeds this interval, which occurs with negligible probability, is regarded as an outlier.
  
\begin{definition}
Assume that the noise in measurements follows a Gaussian distribution $\mathcal{N}(0, \sigma)$.
A measurement is an outlier if the noise within it exceeds the interval $[-3\sigma, 3\sigma]$.
\end{definition}

To detect outliers in distance measurements, we define the residual as
\begin{equation}
\label{req e}
\begin{aligned}
e_{ij}[k] 
=& \Big| \tilde{d}_{ij}[k] - \| p^i_{ji}[k-1] + \Delta \tilde{p}_i^i[k-1] \\
&- (\Delta\tilde{p}_j^j[k-1])^{\times} \cdot r_{ji}[k-1] \| \Big|,
\end{aligned}
\end{equation}
where $\tilde{d}_{ij}[k] \in \mathbb{R}$ denotes the distance measurement between robots $i$ and $j$ at time instant $t=k$,
$\Delta \tilde{p}_i^i \in \mathbb{R}^3$, $\Delta \tilde{p}_j^j \in \mathbb{R}^3$ are the self-displacement measurements of robots $i$ and $j$, respectively,
and $p^i_{ji} \in \mathbb{R}^3$, $r_{ji} \in \mathcal{S}^1$ denote the estimation of relative position and orientation, respectively.
If the measurement $\tilde{d}_{ij}[k]$ is ordinary, the residual $e_{ij}[k]$ is less than a predefined threshold.
In contrast, if $\tilde{d}_{ij}[k]$ is an outlier, $e_{ij}[k]$ will exceed the threshold.
However, checking $e_{ij}[k]$ alone is not sufficient since a large $e_{ij}[k]$ may result from either an outlier $\tilde{d}_{ij}[k]$ or inaccuracies in $(p_{ji}^i[k-1], r_{ji}[k-1])$.

\begin{figure}[tbhp]
	\centering
	\subfigure[Non-rigid graph]
	{
		\includegraphics[width=0.13\textwidth]{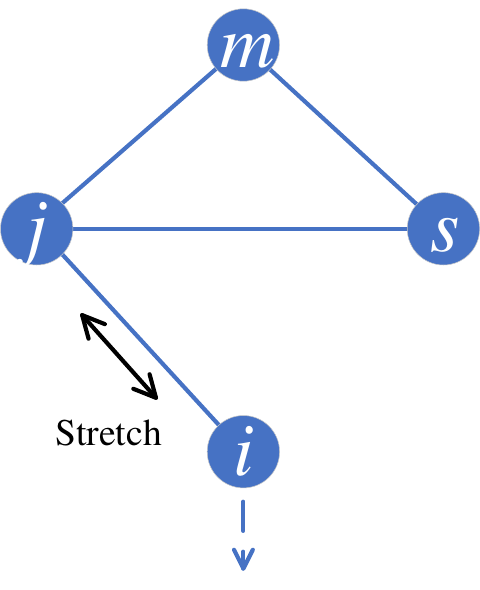}
	}
	\subfigure[Non-rigid graph]
	{
		\includegraphics[width=0.13\textwidth]{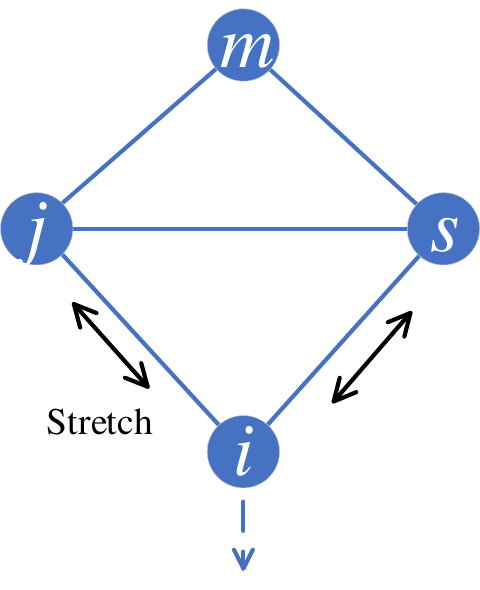}
	}
	\subfigure[Rigid graph]
	{
		\includegraphics[width=0.13\textwidth]{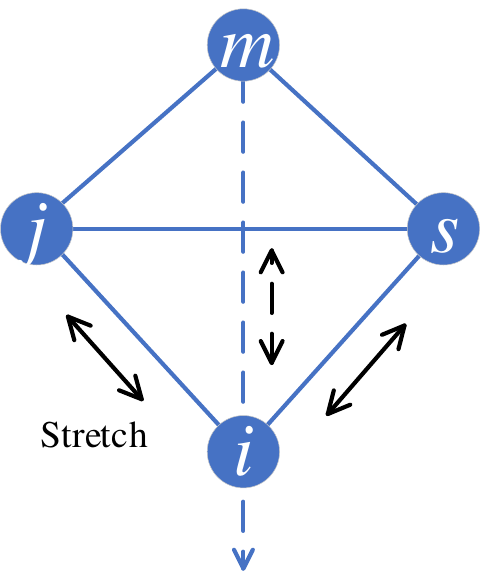}
	}
	\caption{Outlier detectability among robots $i$, $j$, $m$, and $s$}
	\label{fig outliers}
\end{figure}

To address this ambiguity, the triangle inequality can be utilized.
As shown in Figure \ref{fig outliers}(c), if robot $i$  undergoes a significant displacement from $t=k-1$ to $t=k$, all related distances $\tilde{d}_{ij}[k]$, $\tilde{d}_{im}[k]$ and $\tilde{d}_{is}[k]$ should stretch.
Conversely, if $\tilde{d}_{ij}[k]$ is an outlier, only $\tilde{d}_{ij}[k]$ will exhibit an anomalous change.
This case can be verified by checking triangle inequalities in triangles $\triangle ijm$ and $\triangle ijs$, namely,
\begin{equation}
\label{req tri ine}
\begin{aligned}
	& \tilde{d}_{ij}[k] < \tilde{d}_{im}[k] + \tilde{d}_{jm}[k], \\
	& \tilde{d}_{ij}[k] < \tilde{d}_{is}[k] + \tilde{d}_{js}[k], \\
\end{aligned}
\end{equation}
where $\tilde{d}_{jm}[k]$, $\tilde{d}_{js}[k]$ can be obtained by robot $i$ through inter-robot communication.
If both inequalities in \eqref{req tri ine} are violated,  $\tilde{d}_{ij}[k]$ can be reliably identified as an outlier.

A necessary condition for the proposed outlier detection method is that the multi-robot system is tetrahedrally angle rigid.
As illustrated in Figure \ref{fig outliers}(a), if only the edge $(i, j)$ exists, the triangle inequalities in \eqref{req tri ine} cannot be established, rendering outlier detection infeasible.
In Figure \ref{fig outliers}(b), although one triangle inequality can be formed in $\triangle ijs$, it is insufficient to reliably identify outliers.
Specifically, when the inequality is violated, it is ambiguous whether $\tilde{d}_{ij}$ is the outlier or one of $\tilde{d}_{is}$ or $\tilde{d}_{js}$ is the outlier.

After detecting outliers, we mitigate their negative impact on the localization results.
Assume that $\tilde{d}_{ij}[k]$ is identified as an outlier.
Since $\tilde{d}_{ij}[k]$ cannot be utilized, we substitute it with the predicted distance $\| p^i_{ji}[k-1] + \Delta \tilde{p}_i^i[k-1] - (\Delta \tilde{p}_j^j[k-1])^{\times} \cdot r_{ji}[k-1] \|$.
As introduced in Section \ref{sub sec dis trans}, the distance measurements are subsequently transformed into interior angles, which are employed in the proposed relative localization framework.
The resulting angles can be further refined to suppress noise.
For example, in triangle $\triangle ijs$ shown in Figure \ref{fig outliers}(c), the geometric constraint that the sum of interior angles equals $\pi$ can be exploited.
Then a constrained least squares problem \eqref{eq lnp} can be formulated to mitigate the effects of outliers.

\section{Simulations and experiments}
\label{sec sim}

We conduct simulations and experiments to investigate the relative localization among four robots sharing unaligned frames introduced in Section \ref{fourfollowerscase2}.
The video summarizing the experimental results is uploaded to https://www.bilibili.com/video/BV1pkGgzmE47.

\subsection{Improved localization accuracy across Algorithms \ref{al general}-\ref{al total}}
\label{sub sec sim ouralg}

The aim of this simulation is to demonstrate that the relative localization accuracy improves progressively across Algorithms \ref{al general}, \ref{al wtls} and \ref{al total} in the presence of measurement noise.

There are four robots moving in a $10\textrm{m} \times 10\textrm{m} \times 10\textrm{m}$ space.
We conduct a Monte Carlo simulation containing $50$ simulations.
We assume that both angle and self-displacement measurement noise follow a Gaussian distribution with zero mean and a noise scale $\sigma \in \left\{ 0.005, 0.01, 0.05, 0.08 \right\}$.
We utilize the root mean square error (RMSE), defined as $\sqrt{\frac{1}{15} \sum_{i=1}^{15} (\hat{x}_i - x_i)^2}$, where $\hat{x}$ and $x$ denote the estimation and the true value, respectively, to quantify the localization accuracy, where a lower RMSE indicates better accuracy.

\begin{figure}[tbhp]
    \centering
    \subfigure[RMSE of the proposed algorithms' localization results]
    {
        \includegraphics[width=0.35\textwidth]{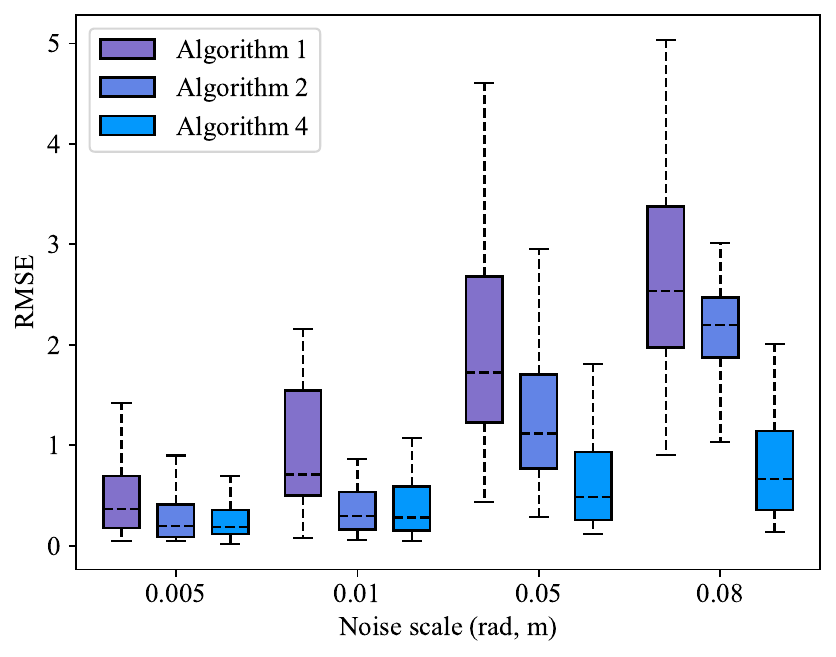}
    }
    \subfigure[Probability density estimation of $p^i_{ji}$ and $r_{ji}$. The black dash lines represent the true values, the blue solid curves demonstrate the probability densities of different values, the dark and light blue shaded regions correspond to the 68$\%$ and 95$\%$ credible intervals, respectively.]
    {
        \includegraphics[width=0.35\textwidth]{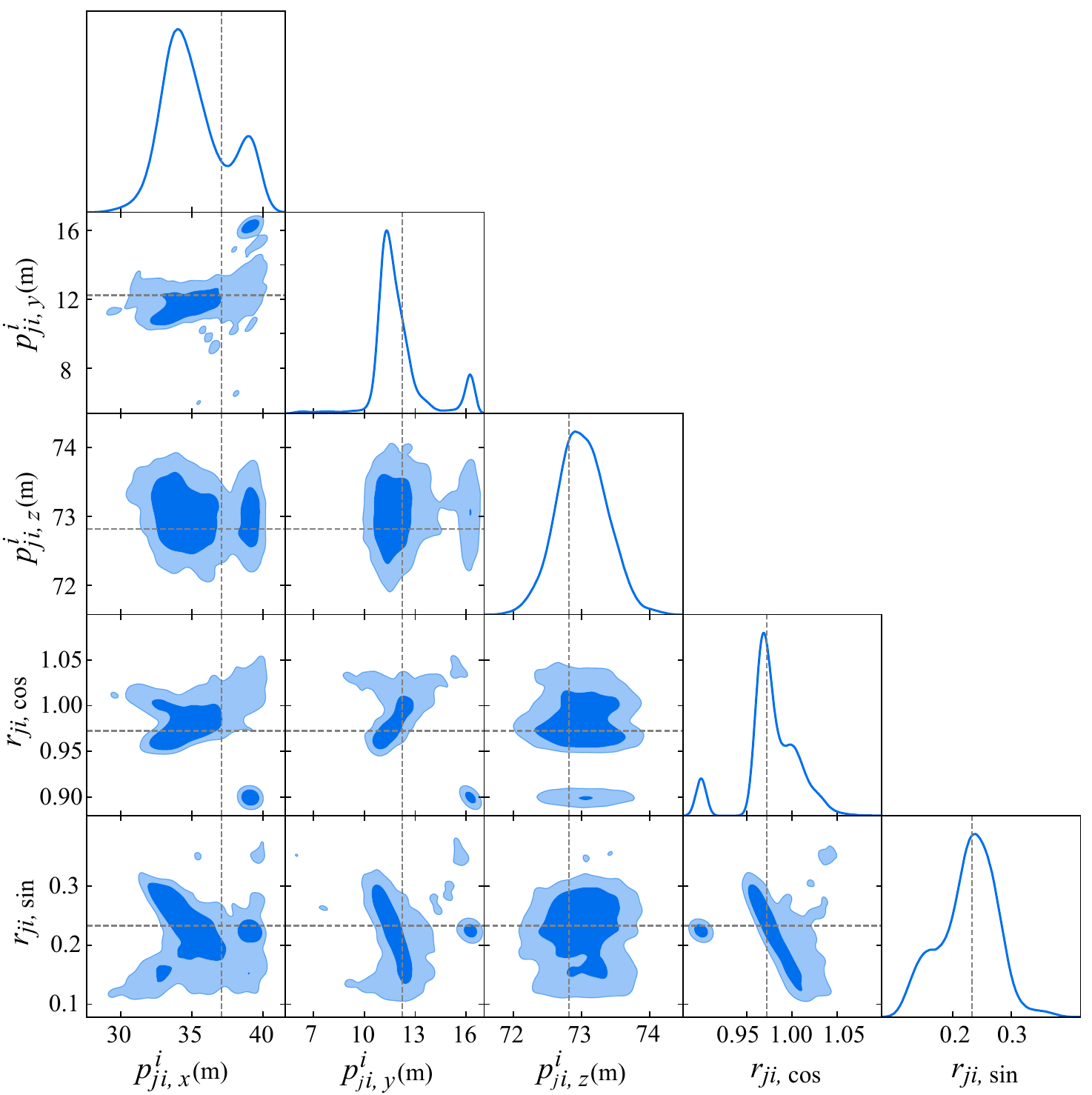}
    }
    \caption{Localization results of the proposed algorithms}
	\label{fig sim ouralg}
\end{figure}

As shown in Figure \ref{fig sim ouralg}(a) and Table \ref{tab sim ouralg}, the localization errors of the results from Algorithm \ref{al general} to Algorithm \ref{al total} gradually decrease, which indicates that the proposed localization framework shown in Figure \ref{fig str} can gradually improve the localization accuracy.
From Figure \ref{fig sim ouralg}(b) one can see that the highest probability density is nearly aligned with the true value, indicating the effectiveness of the density estimation $q(x[0] | \check{x}[0])$ from Algorithm \ref{al nde}.
Additionally, compared to the curve of the probability density, it is evident that the density $p(x[0] | \check{x}[0])$ cannot be simplified to a Gaussian distribution.

\begin{table}[tbhp]
	\centering
	\caption{Mean and variance of RMSE shown in Figure \ref{fig sim ouralg}(a)}
	\label{tab sim ouralg}
	\setlength\tabcolsep{4pt}
	\begin{tabular}{m{0.06\textwidth}<{\centering}m{0.08\textwidth}<{\centering}m{0.06\textwidth}<{\centering}m{0.06\textwidth}<{\centering}m{0.06\textwidth}<{\centering}m{0.06\textwidth}<{\centering}}
		\toprule
		& & $\sigma=0.005$ & $\sigma=0.01$ & $\sigma=0.05$ & $\sigma=0.08$ \\
		\midrule
		\multirowcell{2}{Algo-\\rithm \ref{al general}} &
		Mean  & 0.6888 & 1.1441 & 2.3897 & 2.9730  \\
		& Variance & 1.4541 & 1.1655 & 4.0540 & 6.5170 \\[1ex]
		\multirowcell{2}{Algo-\\rithm \ref{al wtls}} &
		Mean & 0.4202 & 0.5492 & 1.4391 & 2.2578  \\
		& Variance & 0.3592 & 0.8765 & 1.1242 & 0.6563 \\[1ex]
		\multirowcell{2}{Algo-\\rithm \ref{al total}} &
		Mean  & 0.3008 & 0.4920 & 0.6893 & 0.8342  \\
		& Variance & 0.1404 & 0.2980 & 0.3525 & 0.4096 \\
		\bottomrule
	\end{tabular}
\end{table}

\subsection{Effectiveness of Algorithm \ref{al wtls}}

We assess the effectiveness of Algorithm \ref{al wtls} by comparing the localization results from Algorithm \ref{al total} under different initial values when starting to optimize \eqref{eq logmap cons}.

\begin{figure}[tbhp]
    \centering
    \includegraphics[width=0.35\textwidth]{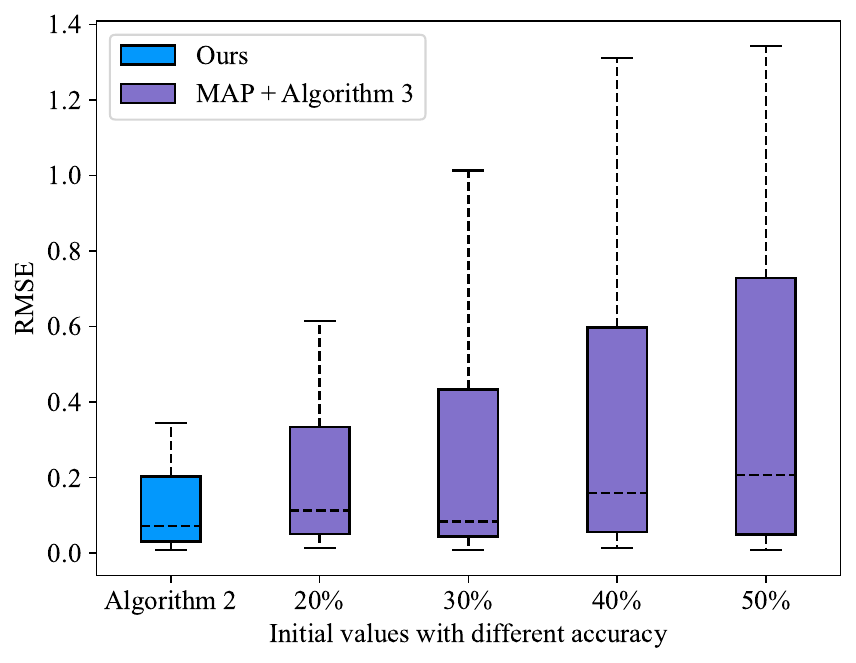}
    \caption{Localization results of the proposed algorithms with different initial values. The percentage on the $x$-axis represents the accuracy level of the initial values and the smaller the percentage, the more accurate the initial values.}
	\label{fig wtls nde}
\end{figure}

Specifically, we evaluate the results when the initial values are replaced with different initial values exhibiting different levels of accuracy.
Both the evaluated and compared algorithms utilize the same NDE, with the only difference being the initial values.
As shown in Figure \ref{fig wtls nde} and Table \ref{tab fig nde}, the accuracy of the localization results derived from Algorithm \ref{al total} decreases as the accuracy of the initial values decreases. 
This is because when solving a complex non-convex optimization problem such as \eqref{eq logmap cons}, the accuracy of the initial values significantly impacts the accuracy of the final results. 
Furthermore, using the results of Algorithm \ref{al wtls} as initial values when optimizing \eqref{eq logmap cons} can significantly enhance accuracy.

\begin{table}[tbhp]
	\centering
	\caption{Mean and variance of RMSE shown in Figure \ref{fig wtls nde}}
	\label{tab fig nde}
	\setlength\tabcolsep{5pt}
	\begin{tabular}{cccccc}
		\toprule
		& Ours & $20\%$ & $30\%$ & $40\%$ & $50\%$ \\
		\midrule
		Mean & \textbf{0.2070} & 0.2363 & 0.2647 & 0.3573 & 0.4185 \\
		Variance & \textbf{0.0870} & 0.1147 & 0.0949 & 0.1486 & 0.2174 \\
		\bottomrule
	\end{tabular}
\end{table}

\subsection{Effectiveness of Algorithm \ref{al nde}}
\label{sub sec sim nde}

We first conduct a simulation to verify that the NDE in Algorithm \ref{al nde} can accurately approximate the probability density.
Subsequently, we apply it to the relative localization problem addressed in this paper to demonstrate its effectiveness in improving localization accuracy.

In the first simulation, we focus on $x = [p_{ji}\T, ~\theta_j^i]\T \in \mathbb{R}^4$ for clarity, where $\theta_j^i \in \mathbb{R}$ denotes the relative rotation angle associated with $r_{ji}$.
Here we compress $r_{ji}$ into $\theta_j^i$ since $r_{ji} =[\cos \theta_j^i,~ \sin \theta_j^i]\T \in \mathcal{S}^1$ contains only one degree of freedom.
The prior density of $x$ is defined as a truncated Gaussian distribution, namely $x \sim \mathcal{N}(0, P_1)$ constrained to a $10\textrm{m} \times 10\textrm{m} \times 10\textrm{m}$ space.
The conditional density is set as a Gaussian mixture, i.e., $p(\check{x} | x) = 0.5 \cdot \mathcal{N}(x, P_2) + 0.5 \cdot \mathcal{N}(x, P_3)$, where $P_1$, $P_2$ and $P_3$ are symmetric positive definite matrices randomly generated to serve as covariance matrices.
Note that $\check{x}_o$ is set as zero.
Since the posterior density in this simulation is analytically tractable as $p(x | \check{x}_o) \varpropto p(\check{x}_o | x) p(x)$, Markov Chain Monte Carlo (MCMC) sampling method \cite{van2018simple} can be employed to depict the distribution of $p(x | \check{x}_o)$, which serves as the ground truth.
The network structure and training process are configured as described in Section \ref{sub sec nde}.

\begin{figure}[tbhp]
\centering
\subfigure[Loss of the NDE in the training process]
{
	\includegraphics[width=0.4\textwidth]{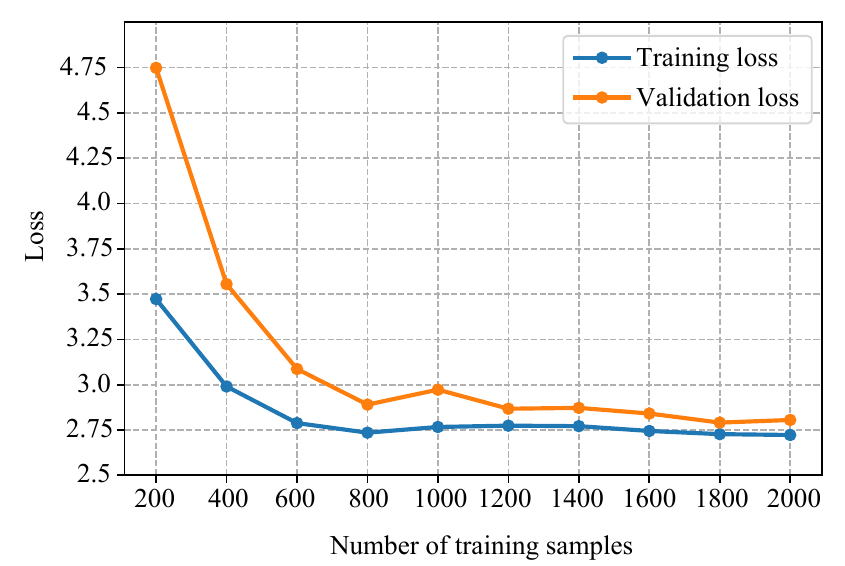}
}
\subfigure[Comparison between the NDE and the ground truth. The dark and light blue shaded regions correspond to the 68$\%$ and 95$\%$ credible intervals, respectively.]
{
	\includegraphics[width=0.35\textwidth]{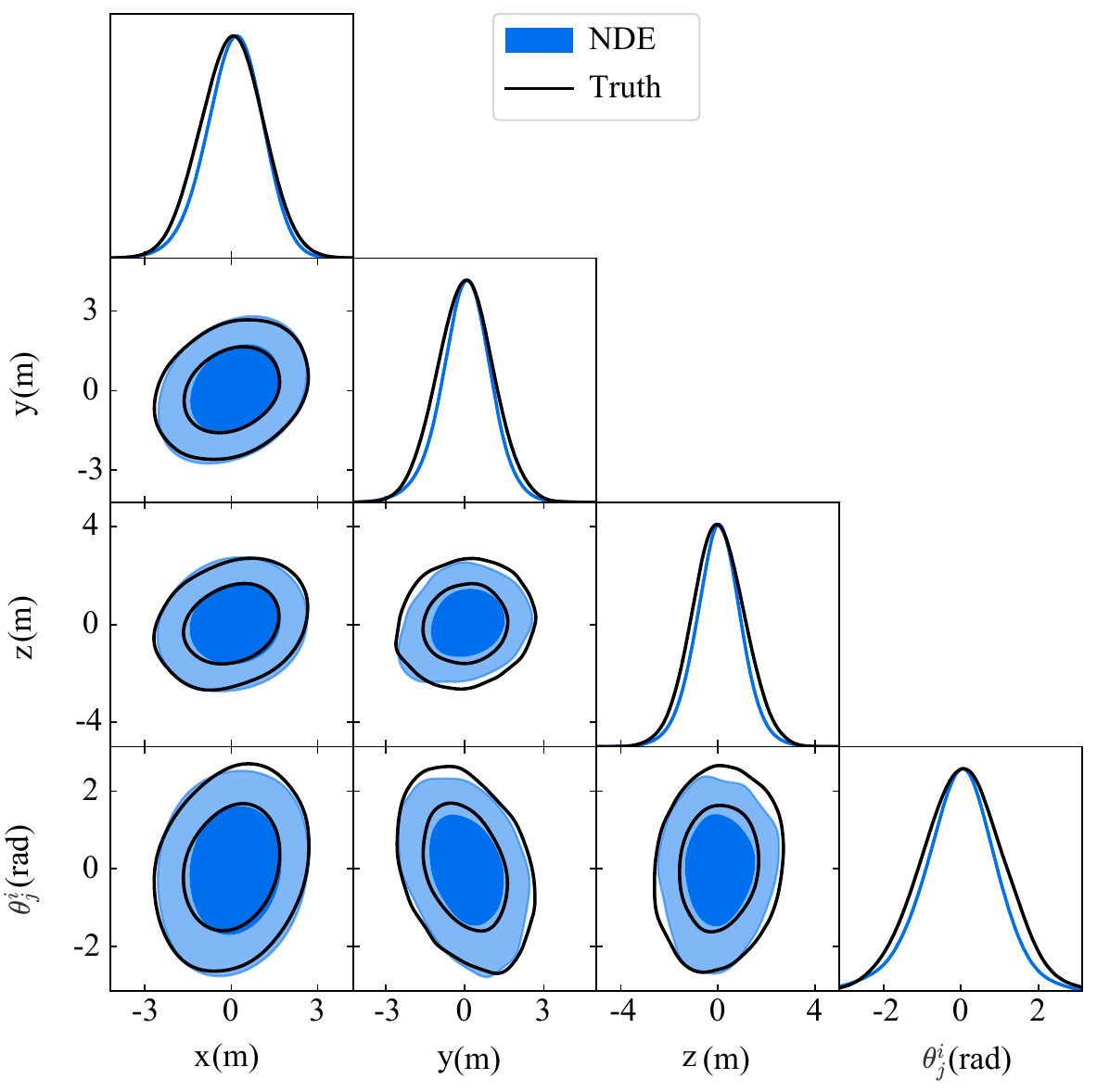}
}
\caption{Training process and approximation accuracy of the NDE}
\label{fig training nde}
\end{figure}

Figure \ref{fig training nde}(a) shows the evolution of the loss function during training.
A subset of the training dataset $\left\{ (\check{x}_1, x_1), \cdots, (\check{x}_s, x_s)\right\}$ is reserved as a validation set, and the validation loss indicates the generalization performance of the NDE.
As shown in the figure, the loss converges rapidly and stabilizes after approximately 1000 training samples.
Figure \ref{fig training nde}(b) shows the comparison of the output of NDE and the ground truth.
The diagonal plots show the marginal probability densities of each component in $x$, while the off-diagonal plots depict the joint probability densities between pairs of components.
The results demonstrate that the NDE approximation aligns closely with the ground truth, indicating that the NDE effectively captures the target density.

In the second simulation, we apply the NDE to the relative localization problem addressed in this paper.
Since the true probability density $p(x | \check{x}_o)$ is intractable, we evaluate the final localization accuracy to verify the effectiveness of it.

Many existing algorithms such as \cite{vezzani2017memory} and \cite{cossette2021relative} assume a Gaussian prior distribution for the MAP estimator.
Here we compare the localization accuracy under NDE and several Gaussian distributions with different parameters.
There are mainly four types of Gaussian distributions based on the combinations of different means and covariances.
The first case is that the mean of the Gaussian distribution equals the truth and the covariance's eigenvalues are small.
This is the ideal case where the mean is highly accurate, and a large weight is assigned to it. 
The second case is that the mean is accurate but the covariance's eigenvalues are large indicating a small weight, which reduces the prior estimation's influence on the MAP estimator's final estimation.
The third and fourth cases are the inaccurate mean with a small and large covariances, respectively.
The key difference between these comparisons lies in the use of different prior probability densities, with identical initial values derived from Algorithm \ref{al wtls} being utilized.

\begin{figure}[tbhp]
    \centering
    \includegraphics[width=0.45\textwidth]{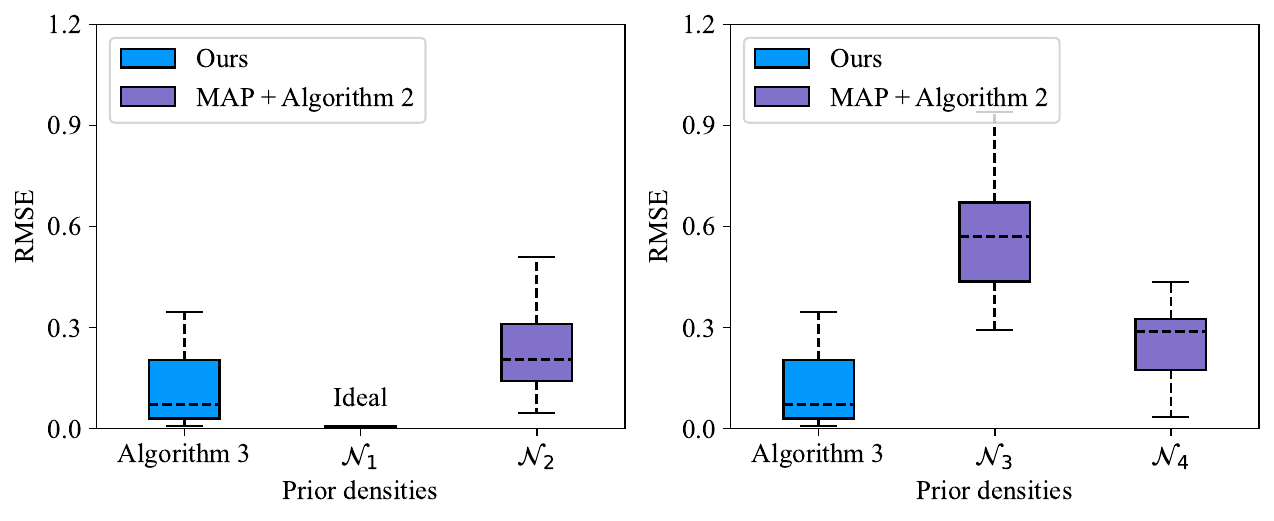}
    \caption{Localization results of different prior densities. Specifically, $\mathcal{N}_1 = \mathcal{N}(\mu_0, (0.1 \sigma)^2 \cdot I)$, $\mathcal{N}_2 = \mathcal{N}(\mu_{0}, (100 \sigma)^2 \cdot I)$, $\mathcal{N}_3 = \mathcal{N}(\mu_{0.3}, (10 \sigma)^2 \cdot I)$, $\mathcal{N}_4 = \mathcal{N}(\mu_{0.3}, (10 \sigma)^2 \cdot I)$, where $\mu_{z}$ denotes a mean offset by $z$ from the true value $\mu$.}
	\label{fig nde}
\end{figure}

As shown in Figure \ref{fig nde} and Table \ref{tab sim nde},
the localization accuracy level of Algorithm \ref{al total} integrating Algorithm \ref{al nde} follows that with the smallest localization error obtained by using $\mathcal{N}_1$, and is better than those obtained by using $\mathcal{N}_2$, $\mathcal{N}_3$ and $\mathcal{N}_4$.
This indicates that Algorithm \ref{al nde} is effective in providing a relatively accurate prior estimation.
The superior localization accuracy associated with $\mathcal{N}_1$ is expected, given that these results rely on highly precise priors, which indicates that the relative positions and orientations of the robots at the initial time instant are precisely known.
However, this condition is idealized, as it is often challenging to obtain exact relative positions and orientations at the initial time instant in practical scenarios.

\begin{table}[tbhp]
	\centering
	\caption{Mean and variance of RMSE shown in Figure \ref{fig nde}}
	\label{tab sim nde}
	\setlength\tabcolsep{5pt}
	\begin{tabular}{cccccc}
		\toprule
		& Ours & $\mathcal{N}_1$ & $\mathcal{N}_2$ & $\mathcal{N}_3$ & $\mathcal{N}_4$ \\
		\midrule
		Mean & 0.2070 & 0.0176 & 0.2556 & 0.5801 & 0.2529 \\
		Variance & 0.0870 & 0.0021 & 0.0268 & 0.0306 & 0.0108 \\
		\bottomrule
	\end{tabular}
\end{table}

\subsection{Comparison with existing algorithms}
\label{sub sec comp}

We compare the proposed Algorithm \ref{al total} with existing relative localization algorithms.
To the best of our knowledge, there is no published literature on using angle measurements for relative localization. 
Most existing works rely on inter-robot distance measurements. 
Thus in this simulation we utilize distance and self-displacement measurements to realize relative localization.
SDP \cite{jiang20193}, PF \cite[Section 4.2.8]{barfoot2024state}, EKF \cite[Section 4.2.3]{barfoot2024state} and nonlinear least square (NLS) \cite{ziegler2021distributed} serve as the baseline algorithms.

We conduct Monte Carlo simulations under different levels of measurement noise and different robot feasible spaces.
The measurement noise scale $\sigma$ belongs to $\{0.005, 0.01, 0.05, 0.08\}$, while the feasible space dimensions are set to $5\textrm{m} \times 5\textrm{m} \times 5\textrm{m}$, $10\textrm{m} \times 10\textrm{m} \times 10\textrm{m}$, and $15\textrm{m} \times 15\textrm{m} \times 15\textrm{m}$.
The number of particles in PF is set to 3000 because the localization result of PF has a certain degree of randomness due to the random generation of particles. 
We find in our simulations that this setting can lead to stable estimation results when using the same measurements.

\begin{figure*}[tbhp]
    \centering
    \subfigure[RMSE within a $5\textrm{m} \times 5\textrm{m} \times 5\textrm{m}$ space]
    {
        \includegraphics[width=0.32\textwidth]{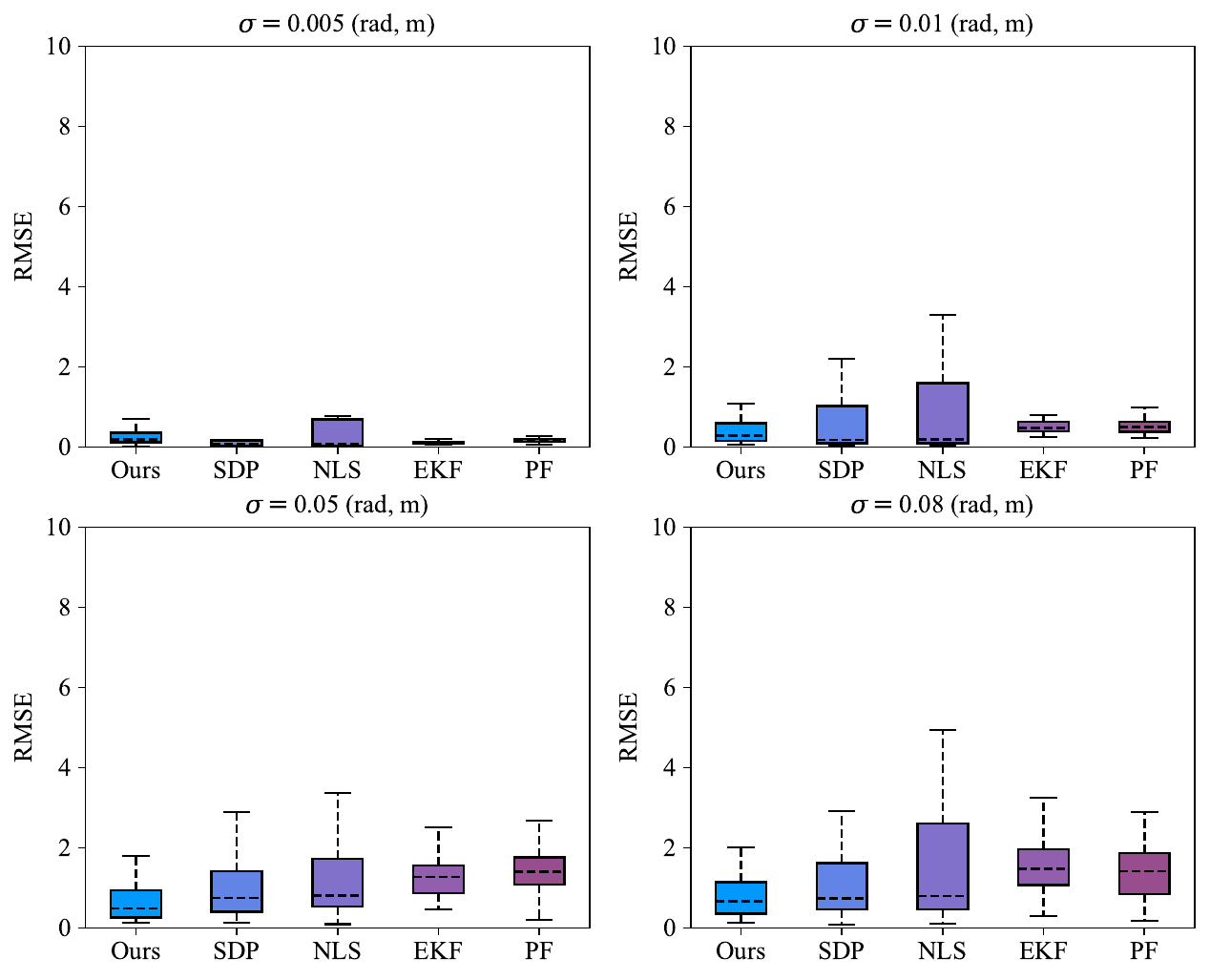}
    }
    \subfigure[RMSE within a $10\textrm{m} \times 10\textrm{m} \times 10\textrm{m}$ space]
    {
        \includegraphics[width=0.32\textwidth]{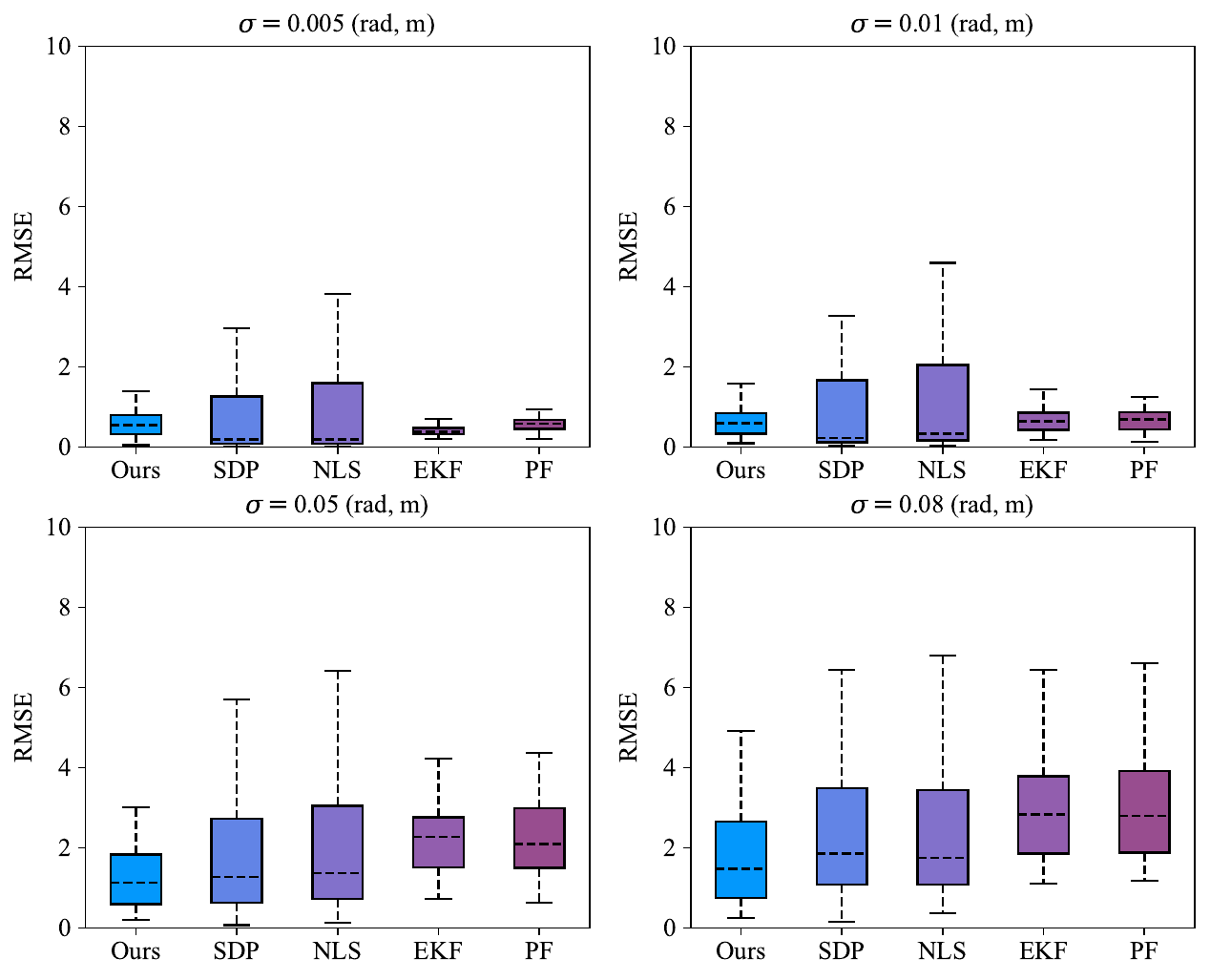}
    }
	\subfigure[RMSE within a $15\textrm{m} \times 15\textrm{m} \times 15\textrm{m}$ space]
    {
        \includegraphics[width=0.32\textwidth]{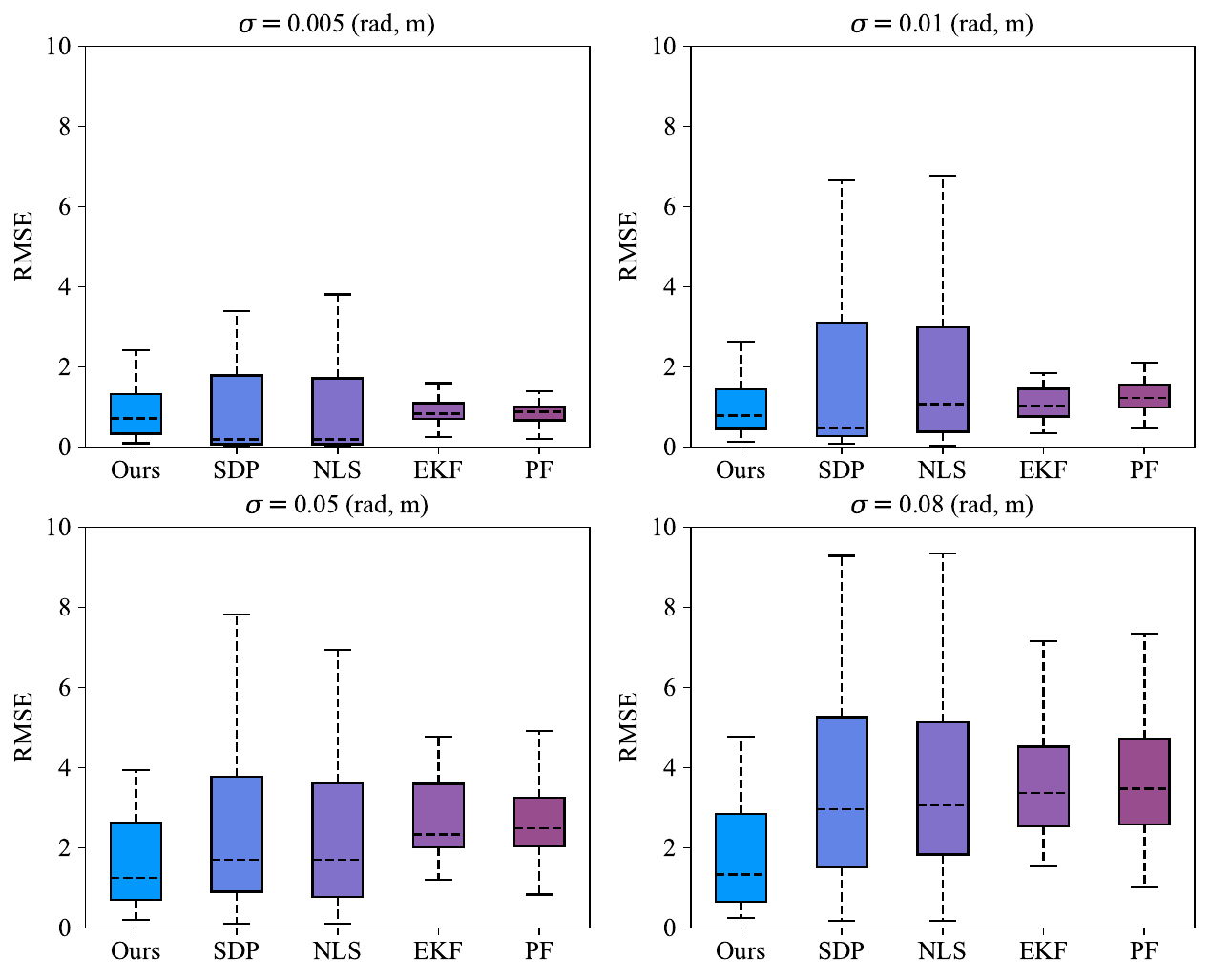}
    }
    \caption{Localization results comparison of different algorithms under different robot feasible spaces and noise scales}
	\label{fig comp algs}
\end{figure*}

Specifically, as shown in Figure \ref{fig comp algs}, the proposed Algorithm \ref{al total} demonstrates high accuracy and small RMSE variance across different measurement noise scales and feasible space dimensions compared to other algorithms.
Additionally, Figure \ref{fig comp algs} shows that the RMSE variance of NLS's localization results is larger than those of other algorithms under the same simulation conditions. 
This is because the NLS formulates the localization estimation problem as a non-convex optimization problem and does not consider how to choose the initial values when optimizing this problem. 
In contrast, SDP proposed in \cite{jiang20193} first transforms the original non-convex optimization problem into an SDP problem and then uses the solution of SDP as the initial values to solve the original non-convex optimization problem. 
Since the algorithm proposed in \cite{jiang20193} additionally considers the choice of initial values, the final localization results of SDP have higher accuracy and smaller RMSE variance compared to NLS. 
However, under high noise, the SDP solution may deviate from the true value due to the difficulty in satisfying all constraints, degrading performance.
Our proposed framework introduces Algorithm \ref{al wtls} to obtain a reliable initialization.
By optimizing only at a single time instant and reformulating the problem on manifolds, Algorithm \ref{al wtls} avoids the instability caused by noise and ensures more robust localization.

It is important to note that when the measurement noise scale $\sigma$ is small, such as $\sigma=0.005$ or $\sigma=0.01$, the localization results of EKF and PF appear better than those of the other three algorithms in Figure \ref{fig comp algs}. 
However, this does not imply that EKF and PF perform better than those three algorithms in this case. 
The superior results of EKF and PF rely upon a highly accurate prior estimation, which must be provided manually before running EKF and PF.
If the prior estimation is not sufficiently accurate, the localization results will not be as favorable as those obtained in this simulation.
In contrast, SDP and NLS do not require any prior estimation, and the proposed Algorithm \ref{al total} can obtain the prior estimation through Algorithm \ref{al wtls} automatically, indicating that the proposed algorithm, along with SDP and NLS, requires less manual intervention.

\begin{table*}[tbhp]
	\centering
	\caption{Mean and variance of RMSE shown in Figure \ref{fig comp algs}}
	\label{tab comp}
	\setlength\tabcolsep{0pt}
	\setlength\extrarowheight{2pt}
	\begin{tabular*}{\textwidth}{@{\extracolsep{\fill}}*{12}{c}}
		\toprule
		\multirow{2}{*}{Space size} & \multirow{2}{*}{Noise scale} & \multicolumn{5}{c}{Mean} & \multicolumn{5}{c}{Variance} \\
		\cline{3-12}
		& & Ours & SDP & NLS & EKF & PF & Ours & SDP & NLS & EKF & PF \\
		\midrule
		\multirowcell{4}{$5\textrm{m} \times 5\textrm{m}$\\$\times 5\textrm{m}$} 
		& $\sigma$ = 0.005 & 0.3007 & 0.4678 & 0.7940 & \textbf{0.1329} & 0.1674 
		& 0.1404 & 0.6575 & 1.8542 & 0.0098 & \textbf{0.0029} \\
		& $\sigma$ = 0.01 & \textbf{0.4919} & 0.5920 & 0.8852 & 0.5530 & 0.5830 
		& 0.2980 & 0.5631 & 1.6760 & \textbf{0.1139} & 0.1538 \\
		& $\sigma$ = 0.05 & \textbf{0.6893} & 0.9869 & 1.3602 & 1.3608 & 1.4939 
		& 0.3525 & 0.6642 & 1.4636 & 0.3733 & \textbf{0.3411} \\
		& $\sigma$ = 0.08 & \textbf{0.8341} & 1.1999 & 1.4985 & 1.4756 & 1.4134 
		& \textbf{0.4096} & 1.2903 & 1.7717 & 0.4269 & 0.6573 \\ [1ex]
		\multirowcell{4}{$10\textrm{m} \times 10\textrm{m}$\\$\times 10\textrm{m}$}  
		& $\sigma$ = 0.005 & 0.7046 & 1.1853 & 1.0296 & \textbf{0.4220} & 0.5836 
		& 0.3623 & 3.3074 & 2.5749 & 0.0252 & \textbf{0.0247} \\
		& $\sigma$ = 0.01 & 0.6945 & 1.2457 & 1.3228 & \textbf{0.6691} & 0.6511 
		& 0.3649 & 3.4654 & 2.8329 & \textbf{0.0958} & 0.1096 \\
		& $\sigma$ = 0.05 & \textbf{1.3986} & 1.8023 & 2.0022 & 2.2584 & 2.2457 
		& 1.1988 & 8.2078 & 2.6422 & 0.9335 & \textbf{1.0377} \\
		& $\sigma$ = 0.08 & \textbf{1.8034} & 2.3996 & 2.5172 & 2.9632 & 2.9481 
		& 2.1651 & 3.7895 & 4.1693 & \textbf{1.7484} &  1.9307 \\ [1ex]
		\multirowcell{4}{$15\textrm{m} \times 15\textrm{m}$\\$\times 15\textrm{m}$}  
		& $\sigma$ = 0.005 & 1.0297 & 1.4550 & 1.8628 & 0.9735 & \textbf{0.8053} 
		& 0.9743 & 8.2078 & 8.8584 & 0.2566 & \textbf{0.0866} \\
		& $\sigma$ = 0.01 & \textbf{1.2769} & 1.9655 & 2.4457 & 1.2908 & 1.3319 
		& 2.4924 & 10.6520 & 10.1230 & \textbf{1.1002} & 1.2489 \\
		& $\sigma$ = 0.05 & \textbf{1.8287} & 2.8812 & 2.7378 & 3.5586 & 2.9978 
		& \textbf{2.6317} & 9.4242 & 7.6467 & 3.3291 & 3.2806 \\
		& $\sigma$ = 0.08 & \textbf{1.9841} & 3.3544 & 3.8136 & 3.6713 & 3.7829 
		& 3.5792 & 6.8213 & 8.2346 & \textbf{2.2610} & 2.4673 \\
		\bottomrule
	\end{tabular*}
\end{table*}

\subsection{Computational time}
\label{sub sec ctime}

In the last simulation, we evaluate the computational time of algorithms. 
The focus of this simulation is primarily on scenarios where the measurement noise scale is relatively large ($\sigma=0.05$), since when noise scale is less, batch estimation algorithms, including Algorithms \ref{al wtls}-\ref{al total}, SDP, and NLS, may not be strictly necessary.
The computational time of different algorithms is recorded in Monte Carlo simulations.
For all optimization problems formulated by different algorithms, the trust region method \cite{absil2007trust} offered by the Pymanopt library\endnote{https://pymanopt.org} is utilized to serve as the solver, and the solver for the SDP problem is offered by the CVXPY library\endnote{https://www.cvxpy.org}.
Default configurations are applied to all solvers.

\begin{figure}[tbhp]
	\centering
	\includegraphics[width=0.35\textwidth]{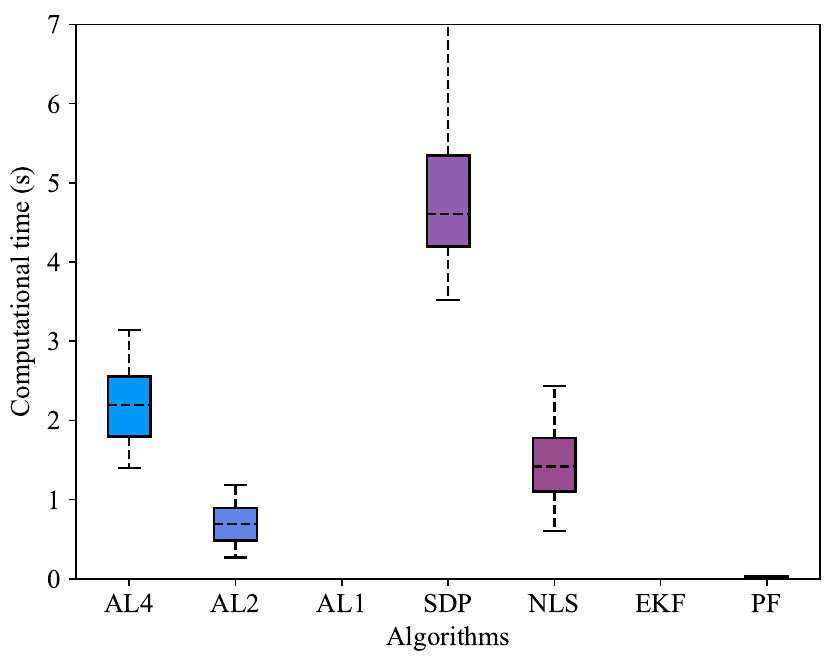}
	\caption{Computational time of different algorithms. The term 'AL' denotes the proposed algorithm.}
	\label{fig ctime}
	\centering
\end{figure}

As shown in Figure \ref{fig ctime} and Table \ref{tab ctime}, the proposed Algorithm \ref{al general} exhibits the lowest computational time, as it only involves solving a linear equation.
In contrast, Algorithm \ref{al total} demands greater computational time, primarily due to its joint estimation of relative positions and orientations across multiple time instants, leveraging nearly all available information to improve localization accuracy.
The simulation results align with the theoretical analysis of the algorithms' computational complexity introduced in Section \ref{sub sec computation}, which primarily determine their computational time.

The computational time of Algorithm \ref{al general}, EKF, and PF is relatively small because they estimate the decision variables (relative position and orientation) at a single time instant without iterative procedures.
In contrast, Algorithm \ref{al total}, NLS and SDP simultaneously estimate relative positions and orientations across multiple time instants, which indicates that the decision variables have a higher dimension.
Notably, the computational complexity of NLS is $O((k_w n)^3)$, which is smaller than that of Algorithm \ref{al total}, explaining the lower computational time required by Algorithm \ref{al total}.
In SDP, by introducing constraints, the non-convex optimization problem is transformed into a convex one.
These additional constraints substantially increase the computational burden, accounting for much of the computational time.

\begin{table}[tbhp]
	\centering
	\caption{Mean and variance of computational time shown in Figure \ref{fig ctime}. The term 'AL' denotes the proposed algorithm.}
	\label{tab ctime}
	\begin{tabular}{ccc}
		\toprule
		& Mean & Variance \\
		\midrule
		AL4 & 3.2168 & 6.5041 \\
		AL2 & 0.6907 & 0.0587 \\
		AL1 & $4.6822 \times 10^{-5}$ & $2.0485 \times 10^{-11}$ \\
		SDP & 4.9373 & 1.2842 \\
		NLS & 1.5188 & 0.3792 \\
		EKF & 0.0001 & $8.6725 \times 10^{-12}$ \\
		PF & 0.0311 & $4.1254 \times 10^{-8}$ \\
		\bottomrule
	\end{tabular}
\end{table}

\subsection{Indoor experiment}
\label{sub sec indoor}
\begin{figure}[tbhp]
	\centering
	\includegraphics[width=0.32\textwidth]{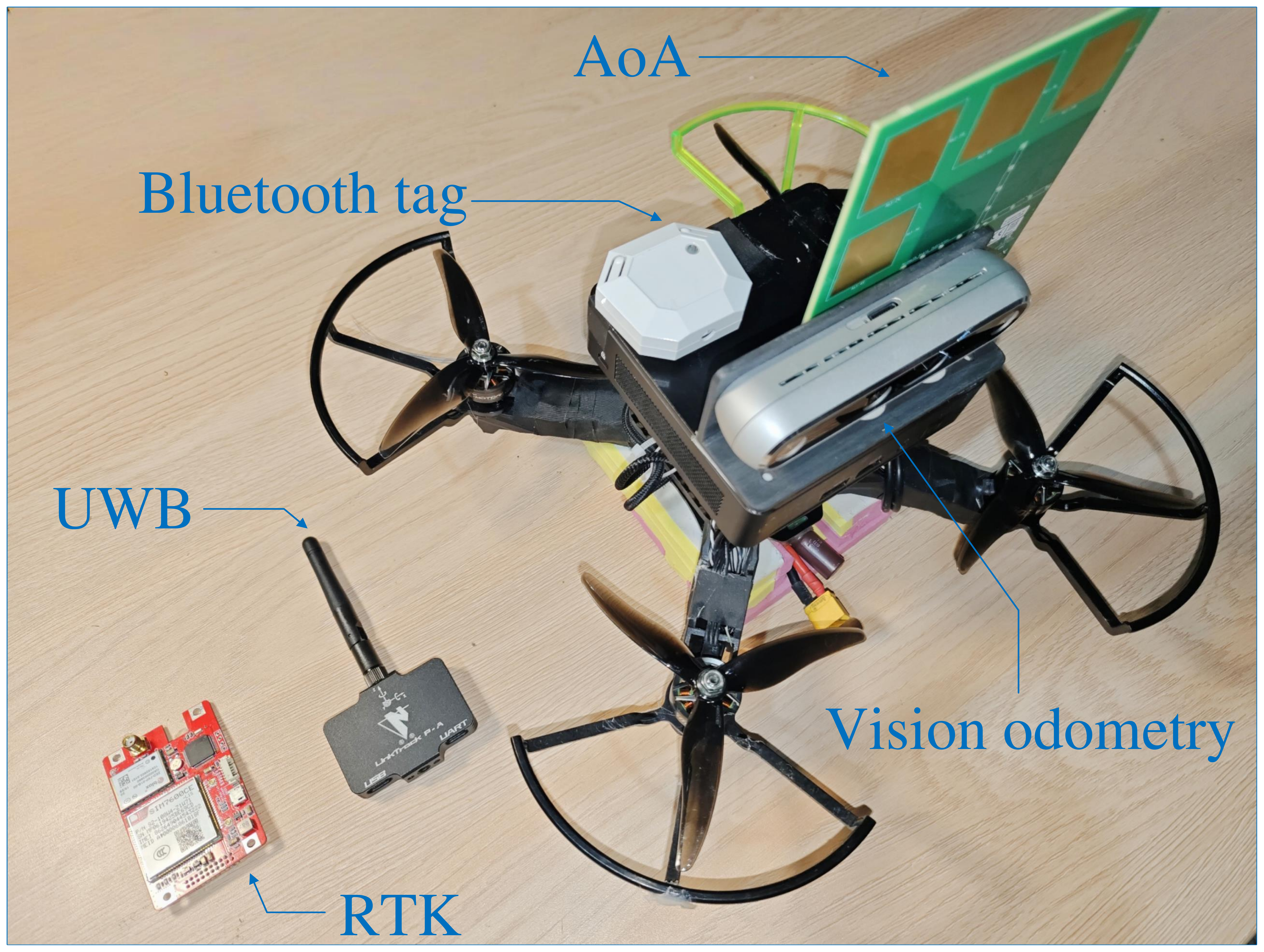}
	\caption{Drone and sensors utilized in experiments}
	\label{fig uav structure}
	\centering
\end{figure}

In the first real-world experiment, we utilize angle and self-displacement measurements to achieve relative localization for multi-robot systems.
To the best of our knowledge, no prior work has been proposed for relative localization based on angle measurements.
To demonstrate the superiority of the proposed relative localization framework, we compare it with EKF and PF, which serve as baselines, since they are universal estimation algorithms regardless of specific types of measurements.
The SDP and NLS algorithms are not included in the comparison since they rely on distance measurements, which are not available in the indoor experiment.

The experiment involves a multi-robot system consisting of four drones, each depicted in Figure \ref{fig uav structure}.
Each drone is equipped with a XPLR-AOA\endnote{https://www.u-blox.com} kit, consisting of an angle-of-arrival (AoA) sensor and a Bluetooth tag. 
The AoA sensor measures the azimuth and elevation angles of electromagnetic waves from Bluetooth tags on other drones. 
The transformation introduced in Section \ref{sub sec angle trans} is applied to convert the raw measurements into the angles required by our algorithms.
Angle measurements are sampled at 50 Hz.
VINS \cite{qin2018vins} captures the self-displacements of the drone at 200 Hz, using a RealSense D455\endnote{https://www.intelrealsense.com} camera and the integrated IMU of the CUAV Nora\endnote{https://www.cuav.net} flight controller.
An Intel NUC 12 Pro\endnote{https://www.intel.sg} with an i7-1260P processor serves as the onboard computer of each drone.
The ground truth of relative positions and orientations is provided by a motion capture system, Vicon\endnote{https://www.vicon.com}, sampled at 100 Hz.
To simplify the experiment, one of the drones (Drone 3) is kept stationary while its sensors remain active, which will not affect the validation of the algorithm.
It is worth mentioning that the accuracy of the measured self-displacements and angles directly determines the accuracy of the proposed relative localization algorithms.
Therefore, on the one hand, by following \cite{nguyen2019persistently}, sensor fusion is embedded in the robots’ self-displacement and angle measurements by using the robots’ raw measured data and the motion capture system's measurements.

\begin{figure}[tbhp]
  	\centering
	\includegraphics[width=0.25\textwidth]{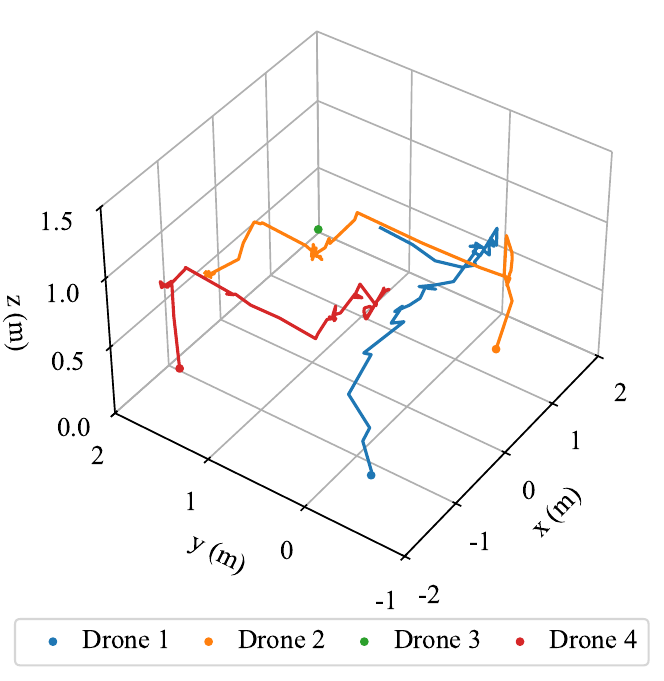}
	\caption{Traces of drones in the indoor experiment}
	\label{fig ang exp trace}
\end{figure}

\begin{figure}[tbhp]
	\centering
  	\includegraphics[width=0.3\textwidth, height=7cm]{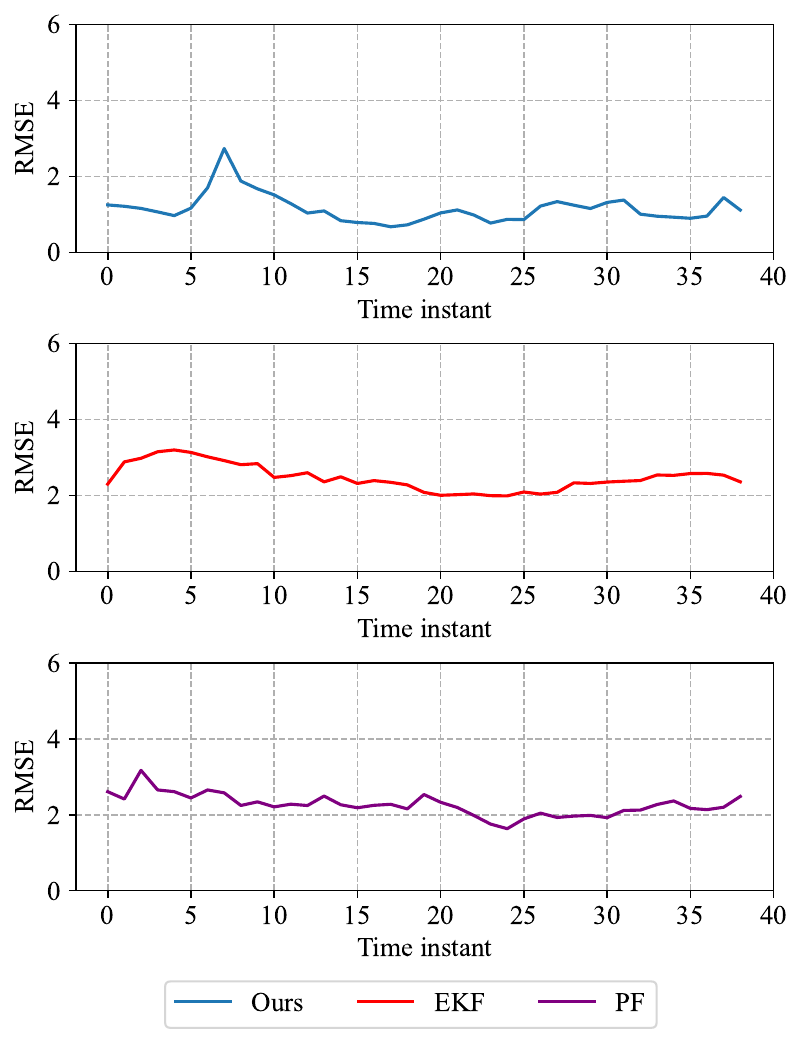}
  	\caption{Localization results comparison of different algorithms in the indoor experiment with angle measurements}
  	\label{fig ang exp comp2}
\end{figure}

The traces of drones are shown in Figure \ref{fig ang exp trace}.
The estimation results are shown in Figure \ref{fig ang exp comp2} and Table \ref{tab ang comparison2}, which show that the proposed framework achieves more accurate localization than both EKF and PF.
Additionally, the angle measurement noise is about $0$-$0.087$rad, and the self-displacement noise is about $0$-$0.2$m, computed based on the ground truth provided by the motion capture system.

\begin{table}[tbhp]
	\centering
	\caption{Mean and variance of RMSE shown in Figure \ref{fig ang exp comp2}}
	\label{tab ang comparison2}
	\begin{tabular}{cccc}
		\toprule
		&  Ours & EKF & PF\\
		\midrule
	 	Mean & \textbf{1.1524} & 2.4690 & 2.2556 \\
		Variance & 0.1403 & 0.1172 & \textbf{0.0801} \\
		\bottomrule
	\end{tabular}
\end{table} 

\subsection{Outdoor experiment}
\label{sub sec outdoor}

In the second experiment, we replace angle measurements with distance measurements to compare the localization accuracy of the proposed framework with several existing algorithms in an outdoor experiment.
In addition to EKF and PF, SDP and NLS algorithms are also included in the comparison.

A UWB sensor, LinkTrack\endnote{https://www.nooploop.com}, provides distance measurements at 10 Hz, and the ground truth is obtained using a real time kinematic (RTK) sensor, OEM-F9P-4G\endnote{https://www.qxwz.com}, sampled at 5 Hz.
The transformation introduced in Section \ref{sub sec dis trans} is used to obtain the angles required by our algorithm from the distance measurements.

\begin{figure}[tbhp]
	\centering
	\includegraphics[width=0.45\textwidth]{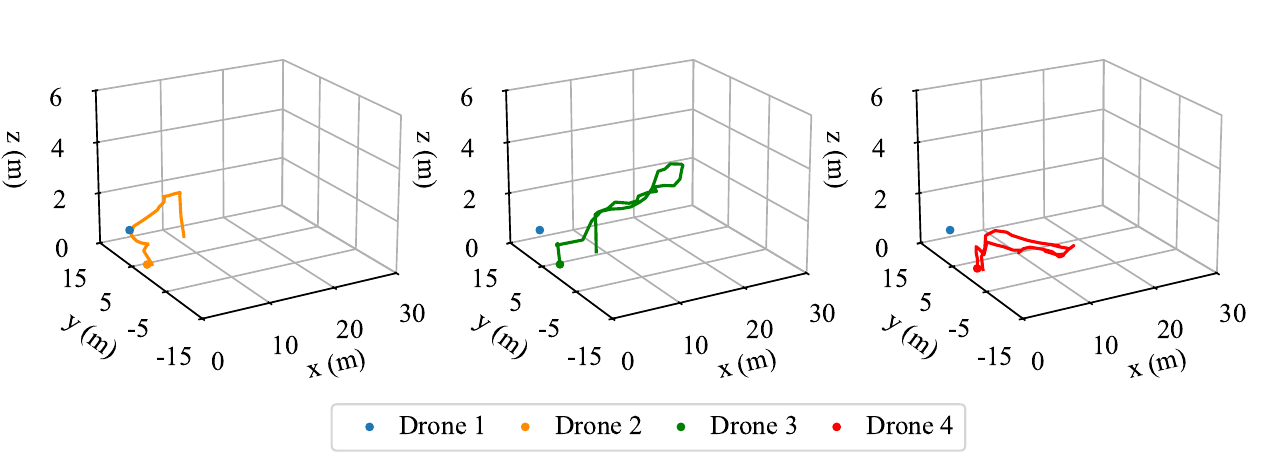}
	\caption{Traces of drones in the outdoor experiment}
	\label{fig trace outdoor}
	\centering
\end{figure}

\begin{figure}[tbhp]
	\centering
	\includegraphics[width=0.4\textwidth]{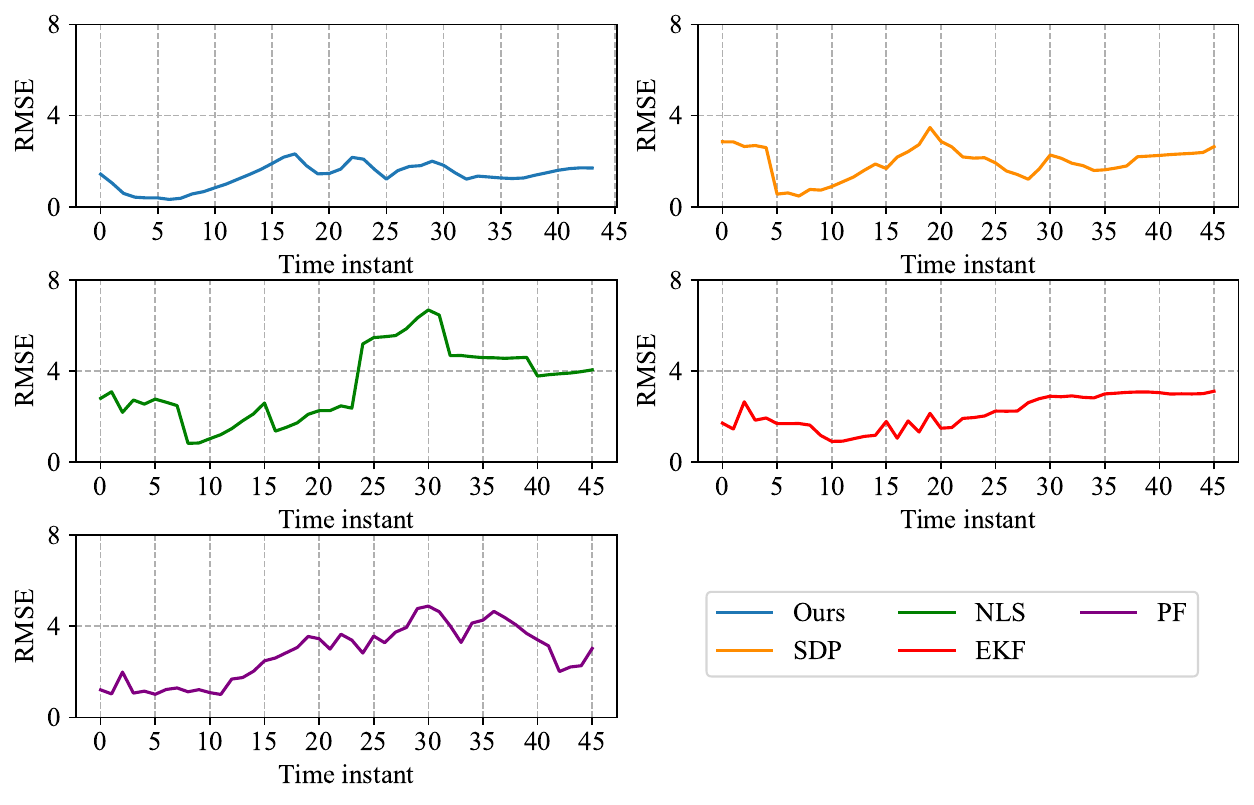}
	\caption{Localization results comparison in the outdoor experiment with distance measurements}
	\label{fig distance exp res}
	\centering
\end{figure}

\begin{figure}[tbhp]
	\centering
	\includegraphics[width=0.4\textwidth]{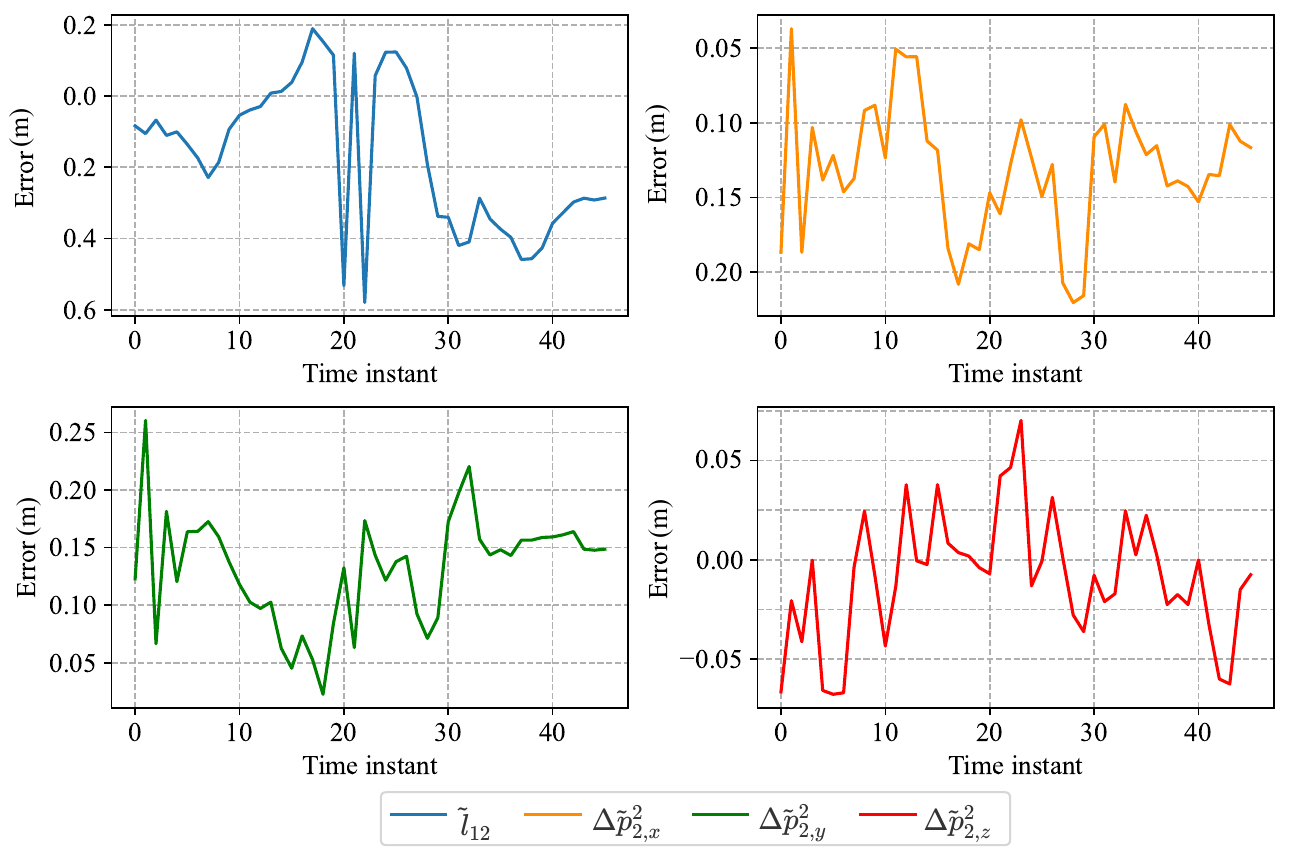}
	\caption{Measurement error in the outdoor experiment. The terms $\tilde{l}_{12}$, $\Delta \tilde{p}_{2,x}^2$ denote the error in the distance measurement between drones $1$ and $2$, and the error in the drone $2$'s self-displacement along the $X$-axis, respectively.}
	\label{fig mea err}
	\centering
\end{figure}

The traces of drones are shown in Figure \ref{fig trace outdoor} and the localization results are shown in Figure \ref{fig distance exp res} and Table \ref{tab comparison}. 
From both the figure and table, it is evident that our proposed algorithm achieves more accurate localization results.
Besides, starting from time instant $t=10$, localization errors of all algorithms increase.
This is due to the drones' expanded motion range, which results in measurements with higher noise levels, as illustrated in Figure \ref{fig mea err}.
Additionally, the distance measurement noise is about $0$-$0.4$m, computed based on the ground truth provided by the RTK sensor.

\begin{table}[tbhp]
	\centering
	\caption{Mean and variance of RMSE shown in Figure \ref{fig distance exp res}}
	\label{tab comparison}
	\begin{tabular}{cccccc}
		\toprule
		&  Ours & SDP & NLS & EKF & PF\\
		\midrule
	 	Mean & \textbf{1.3584} & 1.9391 & 3.3990 & 2.1595 & 2.7848 \\
		Variance & \textbf{0.2754} & 0.4749 & 2.5693 & 0.5245 & 1.4583 \\
		\bottomrule
	\end{tabular}
\end{table} 

\theendnotes

\section{Conclusion}
This paper has proposed a systematic and practical 3-D relative localization framework for multi-robot systems. 
Firstly, we have developed a linear relative localization theory, including the linear localization algorithm and the sufficient conditions for localizability.
Secondly, by utilizing the linear relative localization algorithm, we have addressed two primary challenges existing in the MAP estimator, i.e., the lack of knowledge of the prior density and the choice of initial values when solving the MAP optimization problem. 
Additionally, we also have proposed a marginalization mechanism to maintain a constant size of the state to be estimated.
Simulations and experiments have verified the effectiveness of the framework.


\section*{Appendix I: {Proof of Lemma \ref{lemma1}}}
\label{sec pro lemma1}

By elementary operations, $[A_j, A_m]$ can be transformed as
\begin{equation}
	\label{eq lem1 proof}
	\begin{aligned}
		&[A_j, A_m] \\
		&= \begin{bsmallmatrix}
		s (\alpha_{s'j'i}) I_2 & 0_{2\times 1} & 0_{2\times 2} & 0_{2\times 1} \\
		0_{1\times 2} & c(\alpha_{miZ}) s(\alpha_{mji}) & 0_{1\times 2} & -c(\alpha_{jiZ})s(\alpha_{jmi}) \\
		0_{2\times 2} & 0_{2\times 1} & s(\alpha_{s'm'i}) I_2 & 0_{2\times 1} \\
		0_{1\times 2} & c(\alpha_{siZ})s(\alpha_{sji}) & 0_{1\times 2}& 0_{1\times 1} \\
		\end{bsmallmatrix} \\
		&\stackrel{c_3 \leftrightarrow c_6}{\longrightarrow} \begin{bsmallmatrix}
		s(\alpha_{s'j'i}) I_2 & 0_{2\times 1} & 0_{2\times 2} & 0_{2\times 1} \\
		0_{1\times 2} & -c(\alpha_{jiZ})s(\alpha_{jmi}) & 0_{1\times 2} & c(\alpha_{miZ}) s(\alpha_{mji}) \\
		0_{2\times 2} & 0_{2\times 1} & s(\alpha_{s'm'i})I_2 & 0_{2\times 1} \\
		0_{1\times 2} & 0_{1\times 1} & 0_{1\times 2}&  c(\alpha_{siZ})s(\alpha_{sji}) \\
		\end{bsmallmatrix}  \\
		&\stackrel{r_3 - k r_6}{\longrightarrow} 
		\begin{bsmallmatrix}
		s(\alpha_{s'j'i}) I_2 & 0_{2\times 1} & 0_{2\times 2} & 0_{2\times 1} \\
		0_{1\times 2} & -c(\alpha_{jiZ})s(\alpha_{jmi}) & 0_{1\times 2} & 0_{1\times 1} \\
		0_{2\times 2} & 0_{2\times 1} & s(\alpha_{s'm'i}) I_2 & 0_{2\times 1} \\
		0_{1\times 2} & 0_{1\times 1} & 0_{1\times 2}&  c(\alpha_{siZ}) s(\alpha_{sji}) \\
	\end{bsmallmatrix},
	\end{aligned}
\end{equation}
where $c(\cdot)$, $s(\cdot)$ denote the cosine and sine functions, respectively,
$c_i, r_j$ denote the $i$th column and $j$th row of $[A_j, A_m]$, respectively, and $k = \frac{\cos \alpha_{miZ} \sin \alpha_{mji}}{\cos\alpha_{siZ} \sin \alpha_{sji}}$.
Given Assumption \ref{ass not collinear}, the transformation in \eqref{eq lem1 proof} holds.
Specifically,
condition a) in Assumption \ref{ass not collinear} ensures that $\sin \alpha_{jmi} \not = 0$, $\sin \alpha_{sji} \not = 0$, $\cos \alpha_{jiZ} \not = 0$, $\cos \alpha_{miZ} \not = 0$, $\cos \alpha_{siZ} \not = 0$, 
and
condition b) ensures that $\sin \alpha_{s'j'i} \not = 0$, and 
condition c) ensures $\sin \alpha_{s'm'i} \not = 0$.
As a result, all diagonal elements of the resultant matrix in \eqref{eq lem1 proof} are non-zero.
By the rank invariance property of elementary matrix operations \cite[Section 0.4]{horn2012matrix}, it follows $\textrm{rank}([A_j, A_m]) = 6$.
Similarly, $\textrm{rank}([A_j, A_s]) = 6$ and $\textrm{rank}([A_m, A_s]) = 6$ can be derived.

From the angle-induced linear equation $A_j p^i_{ji} + A_m  p^i_{mi} + A_s p^i_{si} = 0$ related to the tetrahedron $\tetrahedron ijms$, one can derive 
\begin{equation*}
	\begin{aligned}
	&\begin{bmatrix}
	A_j & A_m
	\end{bmatrix}
	\begin{bmatrix}
	p^i_{ji} \\
	p^i_{mi}
	\end{bmatrix}
	+ A_s p^i_{si} = 0 \\
	\Rightarrow &
	\begin{bmatrix}
	p^i_{ji} \\
	p^i_{mi}
	\end{bmatrix}
	=-\begin{bmatrix}
	A_j & A_m
	\end{bmatrix}^{-1} A_s p^i_{si}.
	\end{aligned}
\end{equation*}

\section*{Appendix II: {Proof of Theorem \ref{theorem8}}}
\label{pro theorem8}

The proof consists of two steps.
The first is to prove the linear equation \eqref{eq aligned Q1} is consistent.
It is evident that there must exist solutions that satisfy each equation in the linear equation \eqref{eq aligned Q1}, since it is derived from \eqref{eq aligned pji psi}, \eqref{eq linear eq kp} and \eqref{eq aligned rela}.
Consequently, \eqref{eq aligned Q1} is consistent.

Subsequently, we prove that when the algebraic condition is satisfied, namely $\textrm{rank}(Q_1(k,k+1)) = 3$.
The expression of $Q_1(k,k+1)$ is given by \eqref{eq Q1},
\begin{figure*}
\begin{equation}
	\label{eq Q1}
	\begin{bsmallmatrix}
		\begin{smallmatrix}
			\frac{1}{s(\alpha_{s'j'i}[k])} 
			\big(\big.
				-s(\alpha_{s'j'i}[k]) s(\alpha_{j's'i}[k+1]) R(\alpha_{s'ij'}[k+1]) \\
				+ s(\alpha_{s'j'i}[k+1]) s(\alpha_{j's'i}[k]) R(\alpha_{s'ij'}[k])
			\big.\big)
		\end{smallmatrix} 
		& 0_{2 \times 1}
		\\ 0_{1 \times 2} 
		& \begin{smallmatrix}
			\frac{s(\alpha_{jsi}[k])}{s(\alpha_{jmi}[k]) s(\alpha_{sji}[k]) c(\alpha_{siz}[k])}  
			\big(\big. 
				s(\alpha_{mji}[k+1]) c(\alpha_{miz}[k+1]) s(\alpha_{jmi}[k]) c(\alpha_{jiz}[k]) \\ 
				- s(\alpha_{mji}[k]) c(\alpha_{miz}[k]) s(\alpha_{jmi}[k+1]) c(\alpha_{jiz}[k+1]) 
			\big.\big)
		\end{smallmatrix}
		\\ \begin{smallmatrix}
				\frac{1}{s(\alpha_{s'm'i}[k])} 
				\big(\big.
					-s(\alpha_{s'm'i}[k]) s(\alpha_{m's'i}[k+1]) R(\alpha_{s'im'}[k+1]) \\
					+ s(\alpha_{s'm'i}[k+1]) s(\alpha_{m's'i}[k]) R(\alpha_{s'im'}[k])
				\big.\big) 
		\end{smallmatrix} 
		& 0_{2 \times 1}
		\\ 0_{1 \times 2} 
		& \begin{smallmatrix}
			\frac{1}{s(\alpha_{sji}[k]) c(\alpha_{siz}[k])}
			\big(\big.
				s(\alpha_{jsi}[k]) s(\alpha_{sji}[k+1]) c(\alpha_{jiz}[k]) c(\alpha_{siz}[k+1]) \\
				- s(\alpha_{jsi}[k+1]) s(\alpha_{sji}[k]) c(\alpha_{jiz}[k+1]) c(\alpha_{siz}[k])
			\big.\big)
		\end{smallmatrix}
	\end{bsmallmatrix}.
\end{equation}
\end{figure*}
where $c(\cdot)$, $s(\cdot)$ denote the cosine and sine functions, respectively.
If $\textrm{rank}(Q_1(k,k+1)) = 3$, one $3 \times 3$ submatrix's rank of $Q_1(k,k+1)$ equals three.
We take the first two rows and the last row of $Q_1(k,k+1)$ to form the submatrix, denoted as $Q_1'(k,k+1)$.
To confirm $\textrm{rank}(Q_1'(k,k+1)) = 3$, we need to verify $\left| Q_1'(k,k+1) \right| \not = 0$.
We first determine the condition for $\left| Q_1'(k,k+1) \right| = 0$, and then establish the condition for $\left| Q_1'(k,k+1) \right| \not = 0$ that is the inverse of the former condition.

Note that $Q'_1(k,k+1)$ is a diag block matrix, implying $\left| Q'_1(k,k+1) \right| = \left| Q_{1,1}'(k,k+1) \right| \cdot \left| Q_{1,2}'(k,k+1) \right|$, where $Q_{1,1}'(k,k+1)$ comprises the elements of the first two rows and first two columns of $Q_1(k,k+1)$, and $Q_{1,2}'(k,k+1)$ is the last row and last column element of $Q_1(k,k+1)$.
We can derive $\left| Q'_{1,1}(k,k+1) \right| = \left( \sin \alpha_{j's'i}[k] \sin \alpha_{s'j'i}[k+1] - \sin \alpha_{j's'i}[k+1] \sin \alpha_{s'j'i}[k] \right)^2 \\+ 2\sin \alpha_{j's'i}[k] \sin \alpha_{s'j'i}[k+1] \sin \alpha_{j's'i}[k+1] \sin \alpha_{s'j'i}[k] (1 - \cos \Delta \alpha_{s'ij'}[k])$.
Since none of the angles $\sin \alpha_{j's'i}[k]$, $\sin \alpha_{s'j'i}[k+1]$, $\sin \alpha_{j's'i}[k+1]$, $\sin \alpha_{s'j'i}[k]$ are equal to zero, $\left| Q'_{1,1}(k,k+1) \right| = 0$ if $\Delta \alpha_{s'ij'}[k] = 0$ and $\frac{l_{j'i}[k]}{l_{j'i}[k+1]} = \frac{l_{s'i}[k]}{l_{s'i}[k+1]}$, which is derived from $\sin \alpha_{j's'i}[k] \sin \alpha_{s'j'i}[k+1] - \sin \alpha_{j's'i}[k+1] \sin \alpha_{s'j'i}[k] = 0$ according to the law of sines.
That means $\left| Q'_{1,1}(k,k+1) \right| = 0$ if $\triangle j's'i[k] \sim \triangle j's'i[k+1]$, denoted as $C_1$.

For $Q_{1,2}'(k,k+1)$, we can derive that $\left| Q_{1,2}'(k,k+1) \right| = 0$ if $\frac{\sin \alpha_{jsi}[k] \cos\alpha_{jiz}[k]}{\sin \alpha_{jsi}[k+1] \cos\alpha_{jiz}[k+1]} - \frac{\sin\alpha_{sji}[k] \cos \alpha_{siz}[k]}{\sin\alpha_{sji}[k+1] \cos \alpha_{siz}[k+1]} = 0$, denoted as $C_2$.
Note that $\cos \alpha_{siz}$ and $\cos \alpha_{jiz}$ are guaranteed to be nonzero since Assumption \ref{ass not collinear} holds.
In conclusion, the condition for $\left| Q_1'(k,k+1) \right| = 0$ is $C_1 \lor C_2$, and then the condition for $\left| Q_1'(k,k+1) \right| \not = 0$ is $\lnot C_1 \land \lnot C_2$, i.e., one of $\Delta \alpha_{j's'i}[k]$, $\Delta \alpha_{s'j'i}[k]$ and $\Delta \alpha_{s'ij'}[k]$ is not equal to zero, and $\frac{\sin \alpha_{jsi}[k] \cos\alpha_{jiz}[k]}{\sin \alpha_{jsi}[k+1] \cos\alpha_{jiz}[k+1]} - \frac{\sin\alpha_{sji}[k] \cos \alpha_{siz}[k]}{\sin\alpha_{sji}[k+1] \cos \alpha_{siz}[k+1]} \not = 0$.

Similarly, when we take the third, fourth, and fifth rows of $Q_1(k,k+1)$ to form a submatrix, we can get the similar sufficient condition that guarantees $\textrm{rank}(Q_1(k,k+1)) = 3$.

\section*{Appendix III: {Proof of Theorem \ref{thm  un lin sol}}}
\label{sec pro th2}

As shown in Figure \ref{fig th2} where robots share unaligned frames, the relative localization problem considered in this paper involves the determination of three unknown vectors, namely, $p^i_{ji}$, $p^i_{mi}$, $p^i_{si}$ at time instant $t=k$, and three constant vectors $t=k$, $r_{ji}$, $r_{mi}$ and $r_{si}$, which collectively contain 15 scalar unknowns.
To uniquely determine all unknown scalars, at least 15  independent and relevant linear equations are required.
As introduced in Section \ref{fourfollowerscase2}, these equations can be constructed based on measurements collected from time instant $t=k$ to $t=k+2$.
In what follows, we provide an algebraic proof to demonstrate that all derived equations are linearly independent.

The proof proceeds in two steps.
We first demonstrate that the linear equation \eqref{eq th2} is consistent, i.e., there exists at least one solution of \eqref{eq th2}.
Since all linear equations in \eqref{eq th2} are derived from the fundamental relationships \eqref{eq aligned pji psi}, \eqref{eq linear eq kp}, \eqref{eq aligned rela}, \eqref{eq una times}, \eqref{eq k plus plus} and \eqref{eq una angle kpp}, it is evident that the true relative positions and orientations satisfy all the fundamental relationships if no measurement noise exists.
Thus, \eqref{eq th2} is consistent.

Secondly, we prove that the true relative positions and orientations can be obtained by \eqref{eq th2}.
For \eqref{eq th2} having a unique solution, $\textrm{rank}([Q_2, ~ q_2]) = 9$ is required.
From the consistency result, $\textrm{rank}(Q_2) = \textrm{rank}([Q_2, ~ q_2]) = 9$.
Under Assumptions \ref{ass not collinear}-\ref{ass sim} and the condition that the nonsingularity of $Q\T_2 Q_2$ holds in Theorem \ref{thm  un lin sol}, one can derive that $\textrm{rank}(Q_2) = \textrm{rank}(Q\T_2 Q_2) = 9$.
Consequently, a unique solution can be obtained by solving \eqref{eq th2}.

\section*{Appendix IV: {Proof of Theorem \ref{th loc cond}}}
\label{sec pro loc cond}

It can be inferred that the set $\mathcal{T}_{\tetrahedron i_1 j_1 m_1 s} \subset \mathcal{T}$ is a tetrahedrally angle rigid set since $\mathcal{T}$ is a tetrahedrally angle rigid set.
Therefore, the relative positions and orientations of $j_1$, $m_1$ and $s$ in robot $i_1$'s local frame can be obtained by using angle and self-displacement measurements according to Theorem \ref{theorem8} or Theorem \ref{thm  un lin sol}.
Moreover, there always exists another tetrahedron $\mathcal{T}_{\tetrahedron i_2 j_2 m_2 s} \subset \mathcal{T}$ such that it shares a common vertex $s$ with $\mathcal{T}_{\tetrahedron i_1 j_1 m_1 s}$ according to the definition of $\mathcal{T}$ in Section \ref{sec msd}.
The relative positions and orientations of $i_2$, $j_2$, $m_2$ defined in $s$'s local frame can be determined by the same way.
By transforming the relative positions and orientations of $i_2$, $j_2$, $m_2$ defined in $s$'s local frame into those defined in $i_1$'s local frame, the relative positions and orientations of $i_2$, $j_2$, $m_2$ can be obtained.
Using the same way sequentially for the remaining tetrahedra, relative positions and orientations of all the robots can be determined in robot $i$'s local frame. 

\begin{dci}
  The author(s) declared no potential conflicts of interest with respect to the research, authorship, and/or publication of this paper.
\end{dci}

\begin{acks}
	We appreciate Senior Editor Dr. Dongjun Lee and Associate Editor Prof. Changjoo Nam for the review process of this paper.
	We also appreciate Jianbin Ma, Yao Fang and Aoqi Li for their assistance in the real-world experiments.
\end{acks}

\begin{funding}
	The work of Chenyang Liang, Baoyi Cui and Jie Mei was supported in part by the Shenzhen Science and Technology Program under Grant JCYJ20241202124010014, and in part by the Guangdong Basic and Applied Basic Research Foundation under Grant 2023B1515120018 and Grant 2024B1515040008.
	The work of Liangming Chen was supported in part by the National Natural Science Foundation of China under Grant 62473190 and in part by Guangdong Provincial Natural Science Foundation under Grant 2024A1515011523 and in part by Guangdong Provincial Key Laboratory of Fully Actuated System Control Theory and Technology under Grant 2024B1212010002.
\end{funding}

\bibliographystyle{SageH}
\bibliography{../Refs}

\end{document}